\documentclass{article} 
\usepackage{iclr2025_conference,times}

\usepackage{amsmath,amsfonts,bm}

\def\eqref#1{equation~\ref{#1}}

\def\1{\bm{1}}

\DeclareMathAlphabet{\mathsfit}{\encodingdefault}{\sfdefault}{m}{sl}
\SetMathAlphabet{\mathsfit}{bold}{\encodingdefault}{\sfdefault}{bx}{n}

\usepackage{amsmath}
\usepackage{amssymb}
\usepackage{graphicx}
\usepackage{multirow}
\usepackage[utf8]{inputenc} 
\usepackage[T1]{fontenc}    
\usepackage{hyperref}
\usepackage{url}            
\usepackage{booktabs}       
\usepackage{amsfonts}       
\usepackage{nicefrac}       
\usepackage{microtype}      
\usepackage{xcolor}         
\usepackage{algorithm}
\usepackage{algpseudocode}
\usepackage{mathtools}
\usepackage{bbm}
\usepackage{wrapfig}
\usepackage{subcaption}
\usepackage{caption}
\usepackage{multirow}
\usepackage{makecell}
\usepackage{pifont}

\title{U-shaped and Inverted-U Scaling behind \\Emergent Abilities of Large Language Models}

\author{
    Tung-Yu Wu,
    Melody Lo \\
    National Taiwan University \\
    \texttt{\small \{b08901133, peiyulo\}@ntu.edu.tw}\\
}

\iclrfinalcopy 
\begin{document}

\maketitle

\addtocontents{toc}{\protect\setcounter{tocdepth}{0}}
\begin{abstract}

Large language models (LLMs) have been shown to exhibit \textit{emergent abilities} in some downstream tasks, where model performance stagnates at first and then improves sharply and unpredictably with scale beyond a threshold. In this work, we investigate the phenomenon by grouping questions based on difficulty level and provide a possible explanation for emergent abilities. Specifically, we observe U-shaped scaling for hard questions and inverted-U scaling followed by steady improvement for easy questions. The two scaling patterns initially offset each other, causing stagnant overall performance. The performance starts to soar when the scaling pattern of easy questions reverts from inverse to standard scaling, leading to emergent abilities. Based on this finding, we propose a simple yet effective pipeline, called \textit{Slice-and-Sandwich}, to predict the emergence threshold and model performance beyond the threshold. Our code is publicly available at \url{https://github.com/tony10101105/ExpEmergence}.
\end{abstract}
\section{Introduction}
\label{sec: intro}
Large language models (LLMs)~\citep{team2023gemini, achiam2023gpt, brown2020language, touvron2023llama, touvron2023llama2, workshop2022bloom, li2023textbooks, jiang2024mixtral} have shown strong potential in various downstream applications~\citep{jumper2021highly, fawzi2022discovering, naveed2023comprehensive, kaddour2023challenges}. Though the training-loss scaling law has been well established~\citep{kaplan2020scaling, NEURIPS2022_c1e2faff}, the literature is inconclusive regarding how performance on downstream tasks scales. In particular, for certain downstream tasks~\citep{srivastava2023beyond, lin2021truthfulqa, pilehvar-camacho-collados-2019-wic}, LLMs display \textit{emergent abilities}: performance stagnates even when model training compute scales up hundredfold, and then improves sharply at an unpredictable critical threshold~\citep{wei2022emergent, schaeffer2024emergent}.

Some prior work~\citep{schaeffer2024emergent, schaeffer2024has, lu2023emergent} link emergent abilities to crude performance metrics that fail to capture small model improvements. \citet{hu2023predicting} introduces the \textit{PASSUNTIL} metric, showing gradual model improvement with scale. \citet{schaeffer2024emergent} finds that emergent abilities mainly happen on string-match and multiple-choice tasks~\citep{schaeffer2024emergent}, for which traditional performance measures exhibit strong discontinuity. They propose continuous metrics such as Brier Score~\citep{brier1950verification} and linear metrics such as token edit distance (TED)~\citep{schaeffer2024emergent} to better predict LLM scaling law on downstream tasks. \citet{schaeffer2024has} further ranks several performance metrics in correlation with the model scale. On the other hand, \citet{michaud2024quantization} establish the quantization model of neural scaling to explain the emergent drop of cross-entropy loss from the aspect of next-token prediction.

Another focus of prior literature is the predictability of emergent abilities measured in traditional metrics like accuracy, which is crucial for monitoring and forecasting LLMs' potentially harmful task-wise capabilities, such as writing certain malicious code~\citep{charan2023text}. Though \citet{wei2022emergent} characterizes emergent abilities as \textit{unpredictable} performance soar, some studies~\citep{ruan2024observational, gadre2024language, hu2023predicting, owen2024predictable, ye-etal-2023-predictable} have proposed pipelines to estimate task-specific scaling law. However, they usually incorporate models past the emergence threshold into the training set to fit a Sigmoid function and do not predict emergence abilities.

\begin{figure*}[tb]
  \centering
  \begin{subfigure}{0.45\linewidth}
    \includegraphics[width=\textwidth]{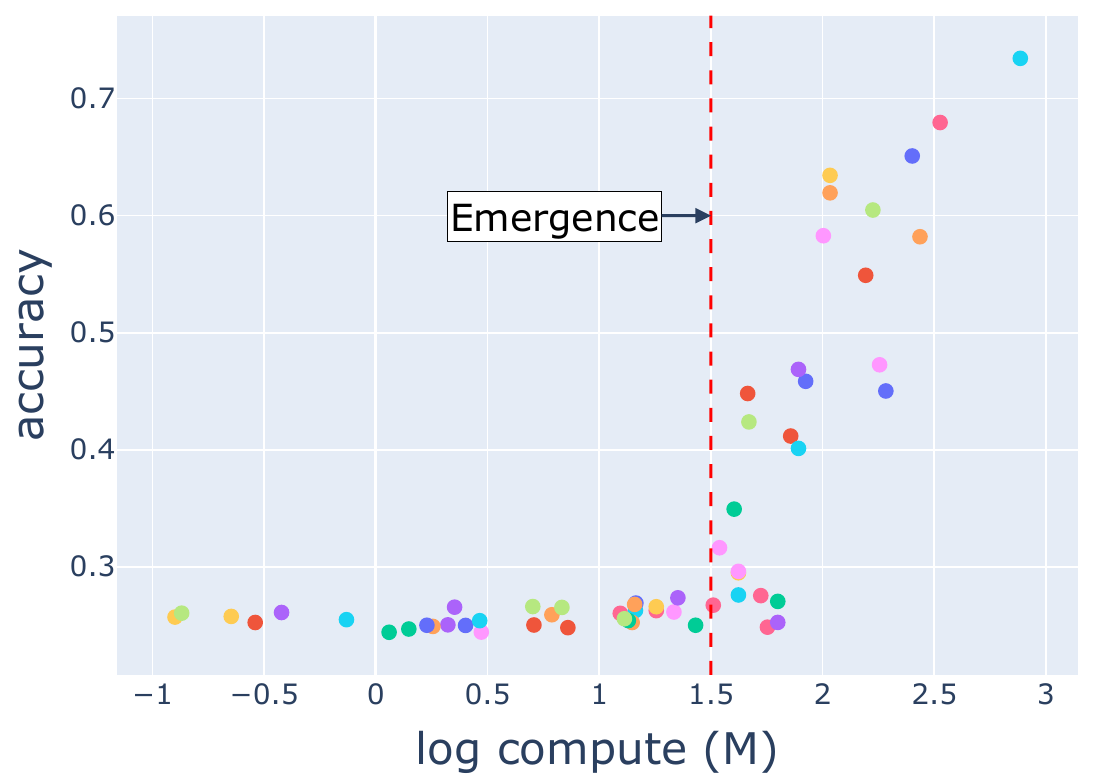}
    \caption{Accuracy vs. log compute (M).}
    \label{subfig: mmlu accuracy}
  \end{subfigure}
  \hfill
  \begin{subfigure}{0.45\linewidth}
    \includegraphics[width=\textwidth]{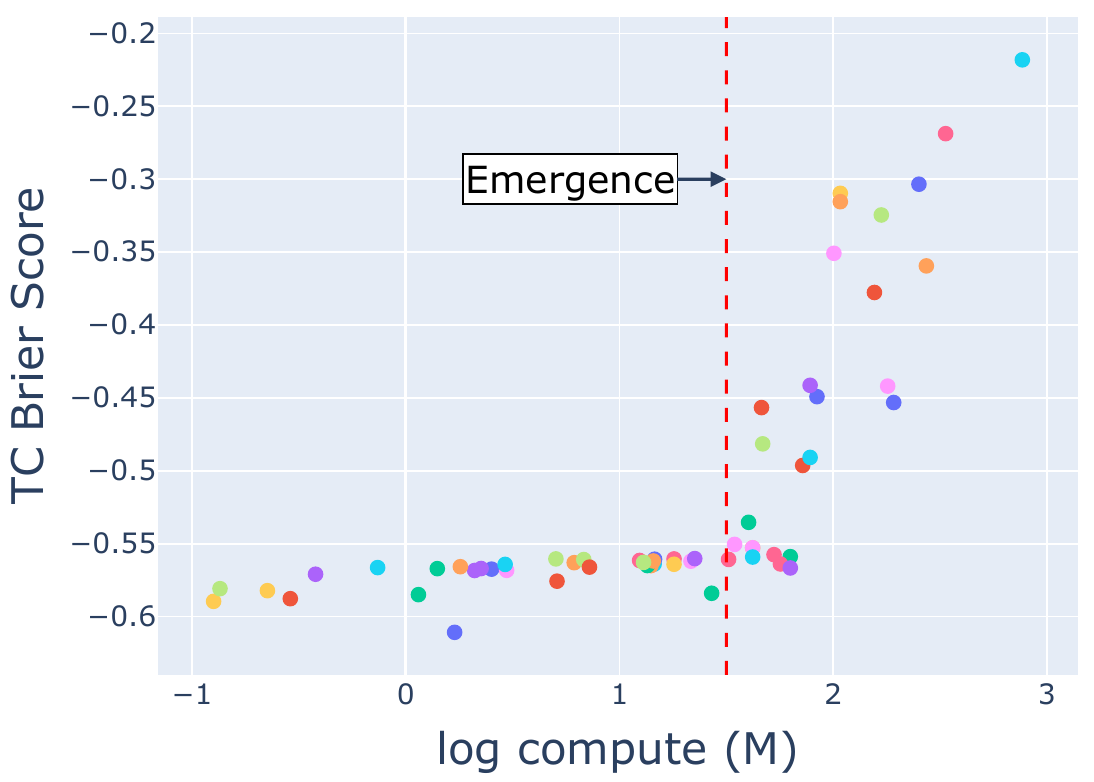}
    \caption{TC Brier Score vs. log compute (M).}
    \label{subfig: mmlu brier}
  \end{subfigure}
  \hfill
  \begin{subfigure}{0.85\linewidth}
    \includegraphics[width=\textwidth]{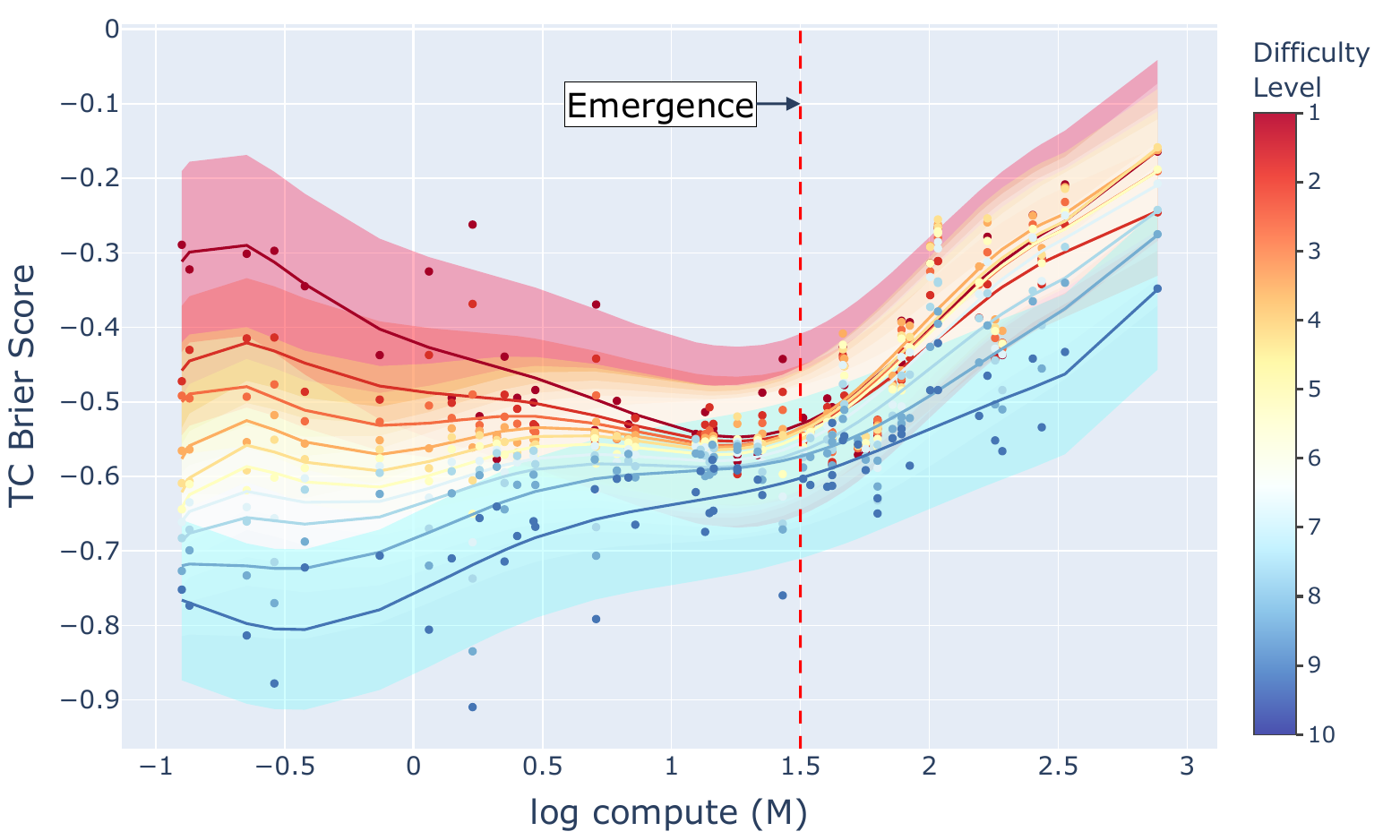}
    \caption{U-Shaped and inverted-U scaling with MMLU's questions clustered into 10 groups. Higher levels are harder questions.}
    \label{subfig: mmlu phenomenon}
  \end{subfigure}
  \caption{The accuracy,  Target-Conditioned (TC) Brier Score, U-shaped and inverted-U scaling on the MMLU benchmark~\citep{hendryckstest2021} evaluated using 56 LLMs. Sec.~\ref{sec2-subsub: continuous metric} provides details on the TC Brier Score, which captures granular changes in model performance. App.~\ref{sup: implementation details} provides details of the 56 LLMs.}
  \label{fig: mmlu phenomenon}
  \vskip -0.25cm
\end{figure*}
This paper contributes to both fronts of discussion on emergent abilities, especially for multiple-choice tasks. First, we propose a novel procedure to evaluate the performance of LLMs on questions grouped by different difficulty levels. Fig.~\ref{fig: mmlu phenomenon} shows the evaluation result of 56 LLMs with diverse training compute on the MMLU benchmark, whose 14042 questions are clustered into 10 groups based on their difficulty levels, with higher levels denoting harder questions. Both model performance and difficulty of each question are measured and calculated by the \textit{Target-Conditioned (TC) Brier Score} (see Sec.~\ref{sec2-subsub: continuous metric} for details), which is our proposed continuous metric that is highly correlated with accuracy but can capture more granular changes in model predictions. We observe that performance on hard questions exhibits U-shaped scaling~\citep{wei-etal-2023-inverse, mckenzie2023inverse}, where it worsens with scale at first and then reverses to improve with scale. In contrast, performance on easy questions exhibits an inverted U-shape followed by steady improvement with scale, consistent with the previously reported deep double descent of testing loss~\citep{nakkiran2021deep}. Moreover, the point at which performance reverts from inverse to standard scaling roughly coincides with the emergence threshold beyond which model performance begins to soar. Our observation could explain why LLM's performance on some multiple-choice tasks stagnates for models below the emergence threshold: the scaling trend on easy questions offsets that on hard questions.

This observation of U-shaped and inverted-U scaling provides a basis to predict the forthcoming sharp increase in model performance, a defining feature of emergent abilities. We propose \textit{Slice-and-Sandwich} pipeline, where we first group questions on a given downstream task by difficulty levels, use data before the emergence threshold to fit the performance on easy and hard questions separately, then forecast performance on easy and hard questions separately beyond the emergence threshold. We show that \textit{Slice-and-Sandwich} captures the performance soar well.

We summarize our contributions as follows:
\begin{itemize}
\item We demonstrate that, for some downstream tasks previously shown to display emergent abilities, under a proper continuous metric, LLM's performance exhibits opposing scaling trends: inverted-U vs. U-shape, on easy vs. hard questions below the emergence threshold, and steadily improves beyond the emergence threshold. 

\item Based on the observation of inverted-U vs. U-shape on easy vs. hard questions, we propose a simple yet effective pipeline, \textit{Slice-and-Sandwich}, to forecast model performance past the emergence threshold. Experimental results on three iconic datasets show its effectiveness.
\end{itemize}
\section{Scaling Trend by Difficulty Level: U-shape vs. Inverted-U}
\label{sec: phenomenon}
This section documents LLM's scaling trend by question difficulty level. Sec.~\ref{sec2-sub: term} defines terminologies such as log compute, emergence threshold, and our performance metrics. Sec.~\ref{sec2-sub: measure diff} describes how we group questions by difficulty level. Sec.~\ref{sec2-sub: ppp present} presents and discusses the results of 6 iconic multiple-choice tasks with emergent abilities.

\subsection{Terminology}
\label{sec2-sub: term}
\subsubsection{Log Compute and Emergence Threshold}
For clearer visualization, in this paper, we refer to an LLM's log compute $M$ as:
\begin{equation}
  M=\log_{10}(\frac{C}{10^{21}}), 
  \label{eq: effective model size}
\end{equation}
where $C\approx6ND$~\citep{kaplan2020scaling} is the total training compute (FLOPs) of an LLM, $N$ is the number of model parameters, and $D$ is the number of training tokens. The emergence threshold $T$ is identified manually as the log compute where the model accuracy exhibits a sharp improvement, as illustrated in Fig.~\ref{subfig: mmlu accuracy}.

\subsubsection{Continuous Performance Metrics}
\label{sec2-subsub: continuous metric}
Prior work~\citep{schaeffer2024emergent, schaeffer2024has, lu2023emergent} has advocated for performance metrics that distinguish finer differences. One candidate metric is the Brier Score~\citep{brier1950verification}:
\begin{equation}
Brier=\frac{1}{K} \sum_{t=1}^K \sum_{i=1}^C\left(\hat{p}_{t,i}-p_{t,i}\right)^2,
  \label{eq: original brier score}
\end{equation}
where $K$ is the number of samples and $C$ is the number of classes. $p_{t,i}$ is 1 if the $t$-th sample belongs to class $i$, and 0 otherwise. $\hat{p}_{t,i}$ is the model's predicted probability of the $t$-th sample being of class $i$.

However, the Brier Score depends not only on the model's predicted probability of the target class (choice) but also on the predicted probability distribution of all classes. Since there is no a priori reason which type of distribution on non-target classes signify better ability \footnote{\citet{schaeffer2024has} found that $\hat{p}^{con}_{t,c}$ (denoted as $p_{\theta}^{Choices}(\text{correct choice})$ in \citet{schaeffer2024has})) has higher correlation with log compute than Brier Score does.}, we propose the \textit{Target-Conditioned (TC) Brier Score} that lumps all non-target available classes into one class, conditioned on available classes, and has an opposite sign to Eq.~\ref{eq: original brier score} so that higher score means higher performance:
\begin{equation}
  TC\_Brier=-\frac{1}{K} \sum_{t=1}^K \left(\hat{p}^{con}_{t,c}-1\right)^2,
  \label{eq: binary brier score}
\end{equation}
where $\hat{p}^{con}_{t,c}$ is the model's predicted probability of the $t$-th sample being of target class $c$ conditional on available classes, i.e.,
\begin{equation}
\hat{p}^{con}_{t,c}=\frac{\hat{p}_{t,c}}{\displaystyle\sum_{i \in \text{available classes}}\hat{p}_{t,i}},
    \label{eq: conditional prob}
\end{equation}
where c is the target class, $\hat{p}_{t,c}$ is the model's predicted probability of the $t$-th sample being of the target class.  We discuss the effect of conditioning in App.~\ref{sup: brier score comparison}.

\begin{figure*}[tb]
  \centering
  \begin{subfigure}{0.45\linewidth}
    \includegraphics[width=\textwidth]{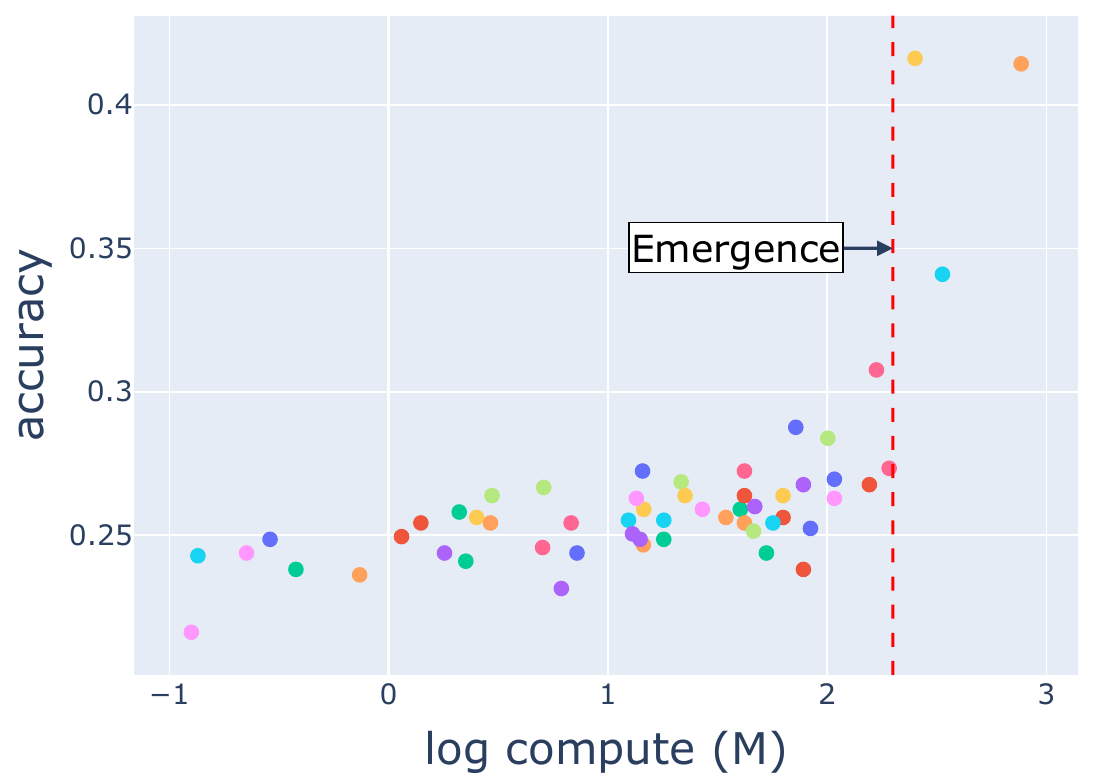}
    \caption{Accuracy vs. log compute (M).}
    \label{subfig: persian-qa accuracy}
  \end{subfigure}
  \hfill
  \begin{subfigure}{0.45\linewidth}
    \includegraphics[width=\textwidth]{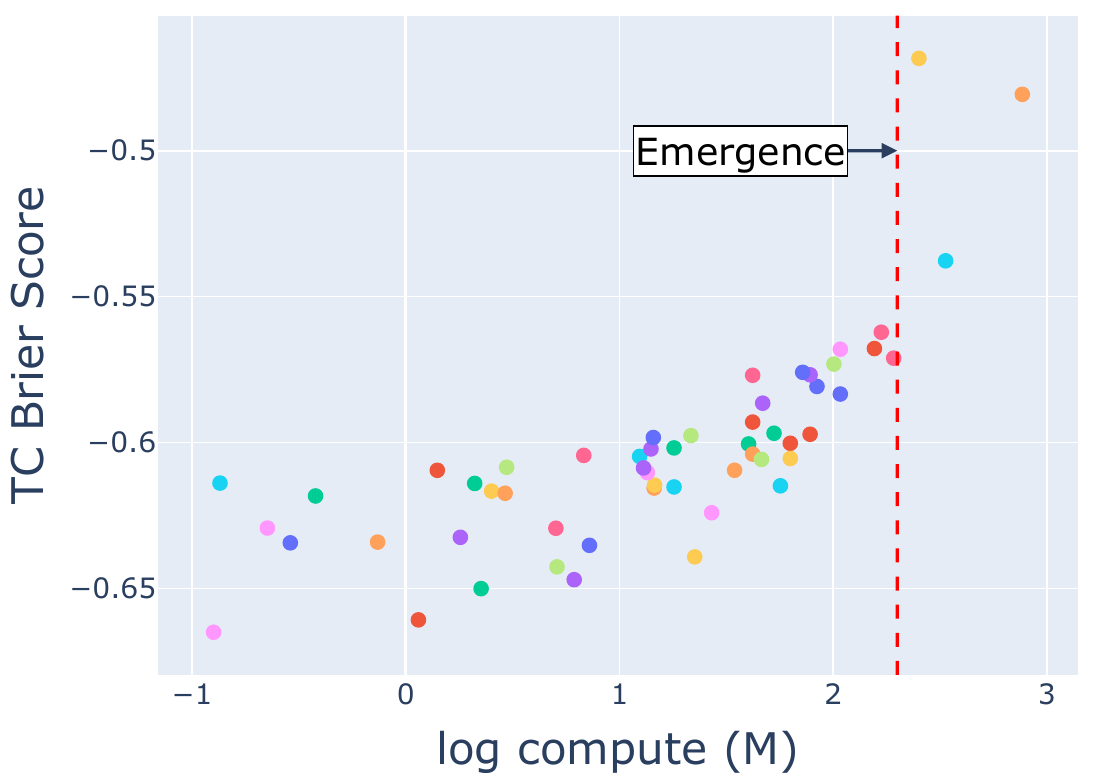}
    \caption{TC Brier Score vs. log compute (M).}
    \label{subfig: persian-qa brier}
  \end{subfigure}
  \hfill
  \begin{subfigure}{0.85\linewidth}
    \includegraphics[width=\textwidth]{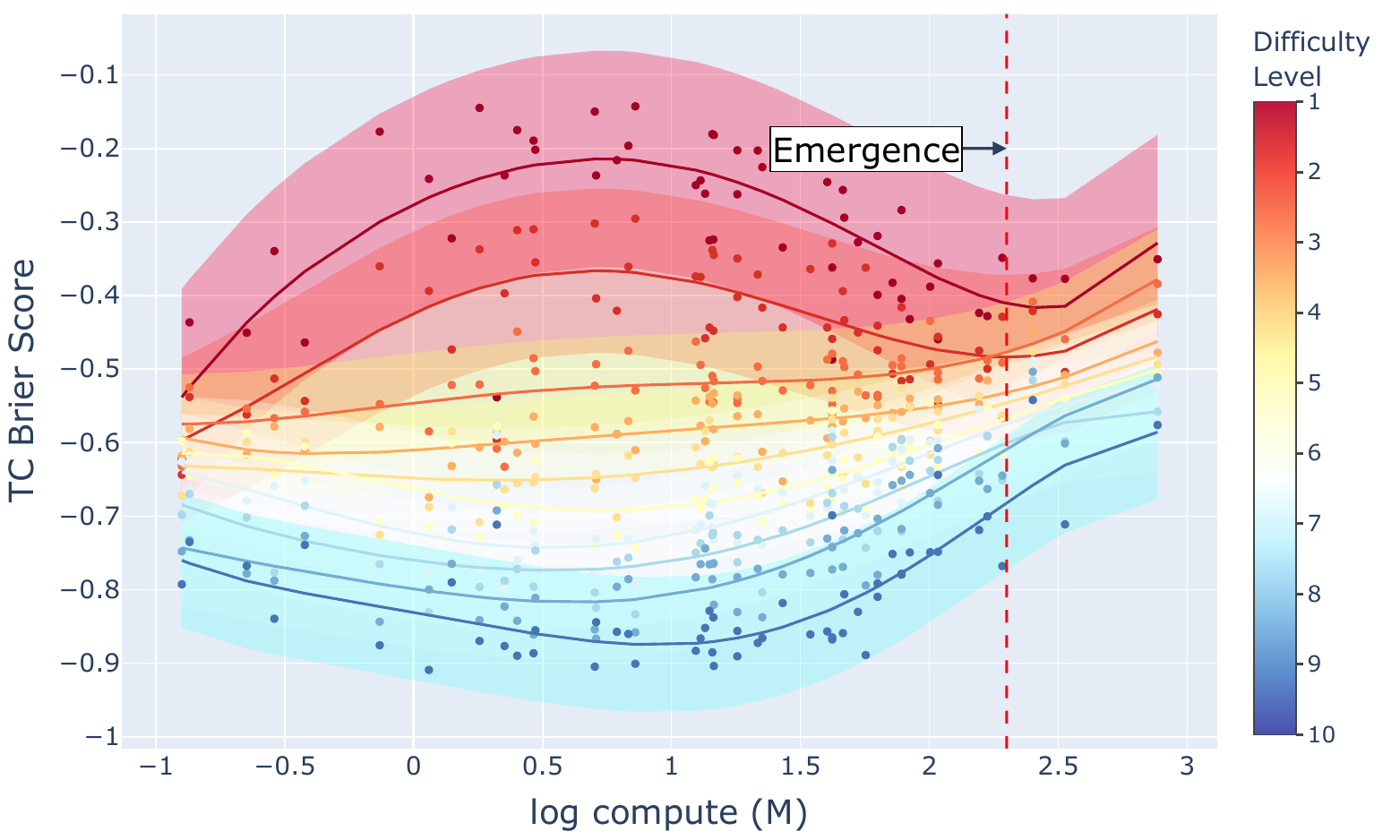}
    \caption{U-Shaped and inverted-U scaling.}
    \label{subfig: persian-qa phenomenon}
  \end{subfigure}
  \caption{The accuracy, TC Brier Score, U-Shaped and inverted-U scaling on the Persian-QA dataset in BIG-bench~\citep{srivastava2023beyond}.}
  \label{fig: persian-qa phenomenon}
  \vskip -0.25cm
\end{figure*}
\subsection{Grouping Questions by Difficulty Levels}
\label{sec2-sub: measure diff}
\subsubsection{Measuring Question Difficulty Level}

For a question $q$ of a downstream task with emergence threshold $T$, we define its difficulty level $D_q$ to be the question-level TC Brier score that takes as samples the outputs on question $q$ from all $L$ LLMs smaller than the emergence threshold $T$. More specifically, we define
\[
D_q=-\frac{1}{L} \sum_{t=1}^L \left(\hat{p}^{con}_{t,c}-1\right)^2.
\] 

\subsubsection{Question Sorting and Grouping}
Because model performance on individual questions is quite noisy, we group questions by difficulty levels. First, we sort questions by ascending difficulty level. Then, we evenly divide the sorted questions into $G$ groups. Thus, each group has a different difficulty level.

\subsection{U-Shaped and Inverted-U Scaling}
\label{sec2-sub: ppp present}
\begin{figure*}[tb]
  \centering
  \begin{subfigure}{0.45\linewidth}
    \includegraphics[width=\textwidth]{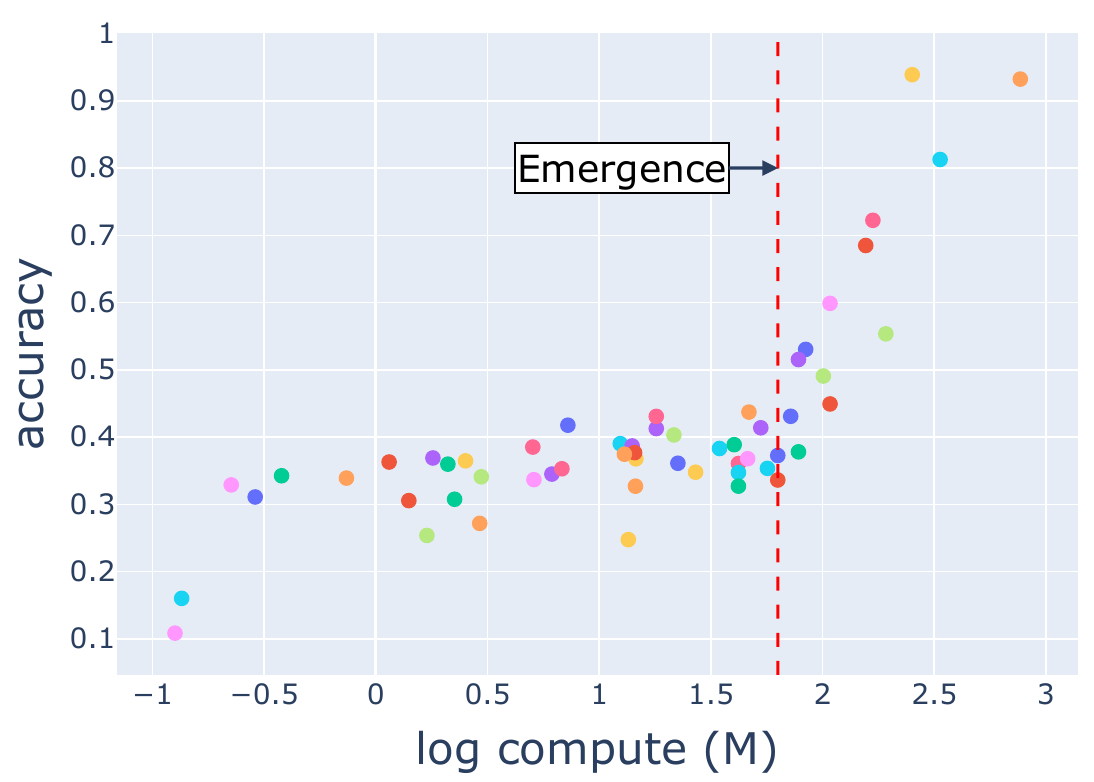}
    \caption{Accuracy vs. log compute (M).}
    \label{subfig: arithmetic accuracy}
  \end{subfigure}
  \hfill
  \begin{subfigure}{0.45\linewidth}
    \includegraphics[width=\textwidth]{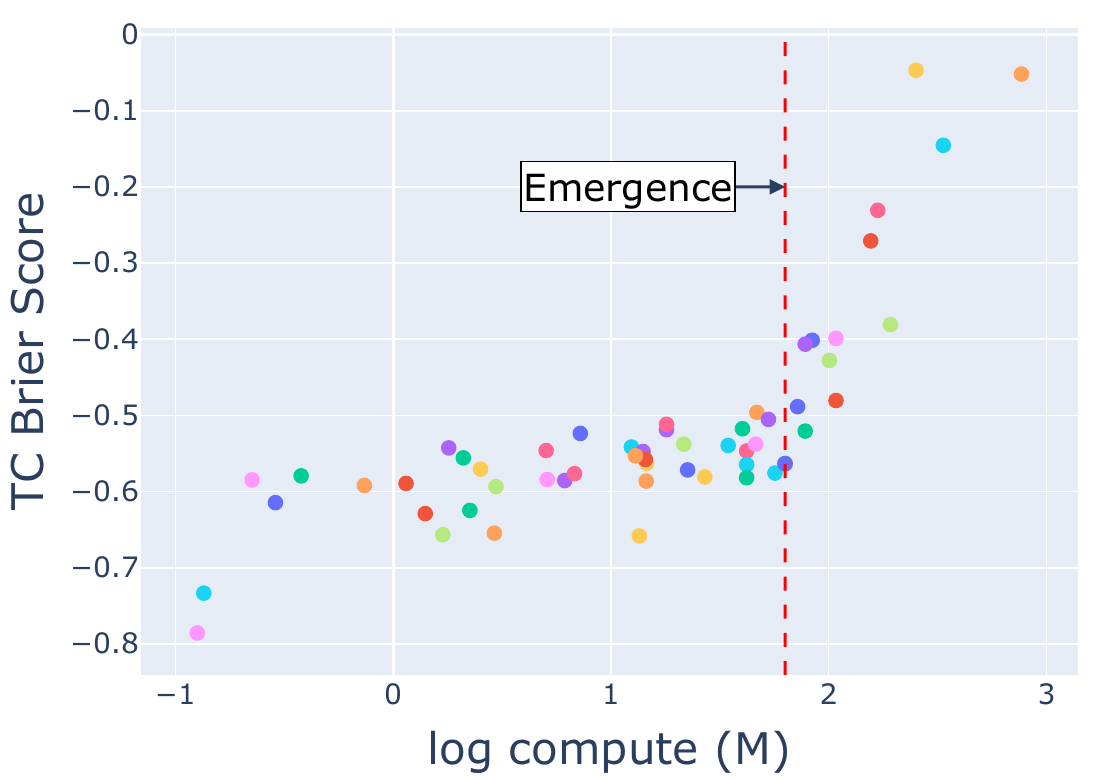}
    \caption{TC Brier Score vs. log compute (M).}
    \label{subfig: arithmetic brier}
  \end{subfigure}
  \hfill
  \begin{subfigure}{0.85\linewidth}
    \includegraphics[width=\textwidth]{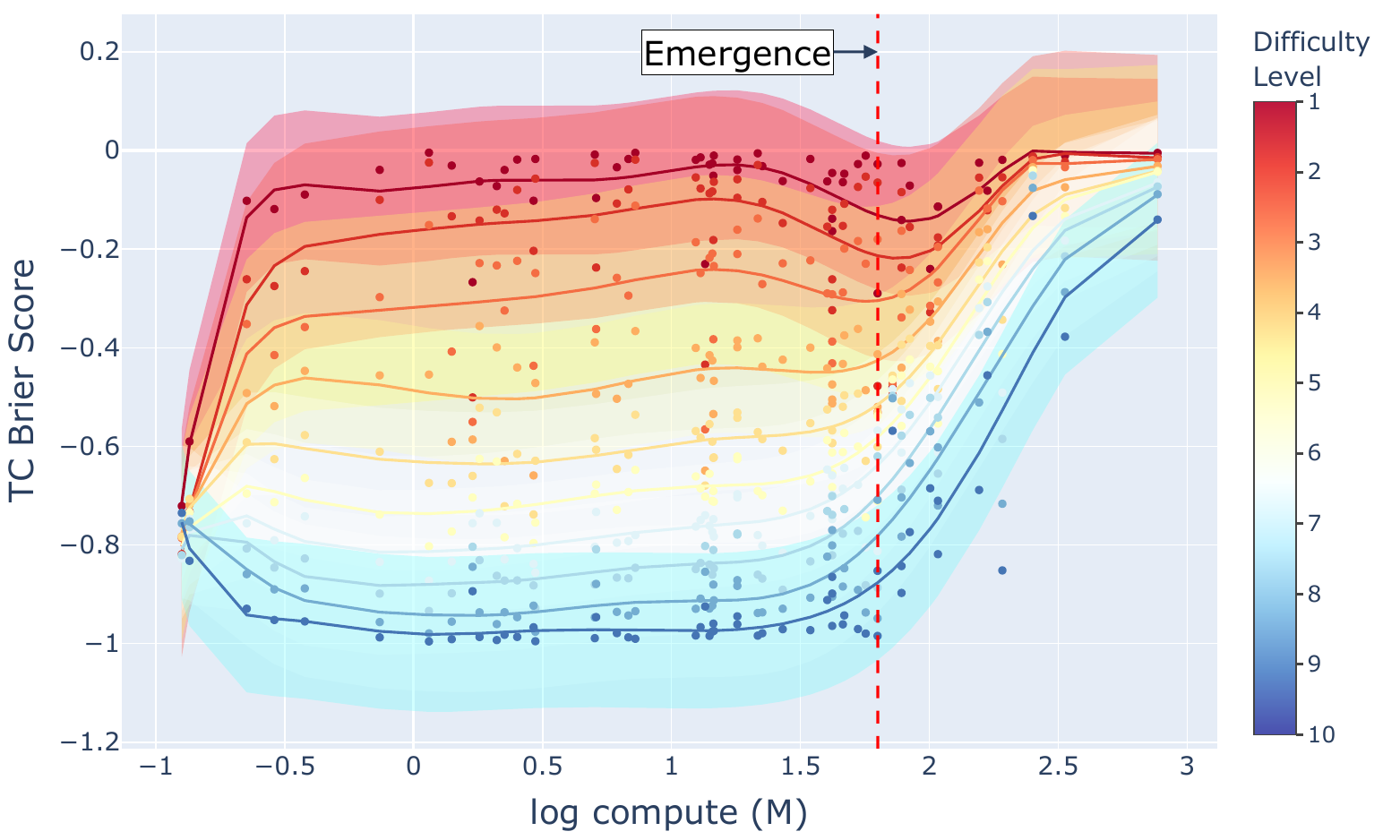}
    \caption{U-shaped and inverted-U scaling.}
    \label{subfig: arithmetic phenomenon}
  \end{subfigure}
  \caption{The accuracy, TC Brier Score, U-shaped and inverted-U scaling on the arithmetic dataset in BIG-bench~\citep{srivastava2023beyond}.}
  \label{fig: arithmetic phenomenon}
\end{figure*}
\begin{figure*}[tb]
  \centering
  \begin{subfigure}{0.32\linewidth}
    \includegraphics[width=\textwidth]{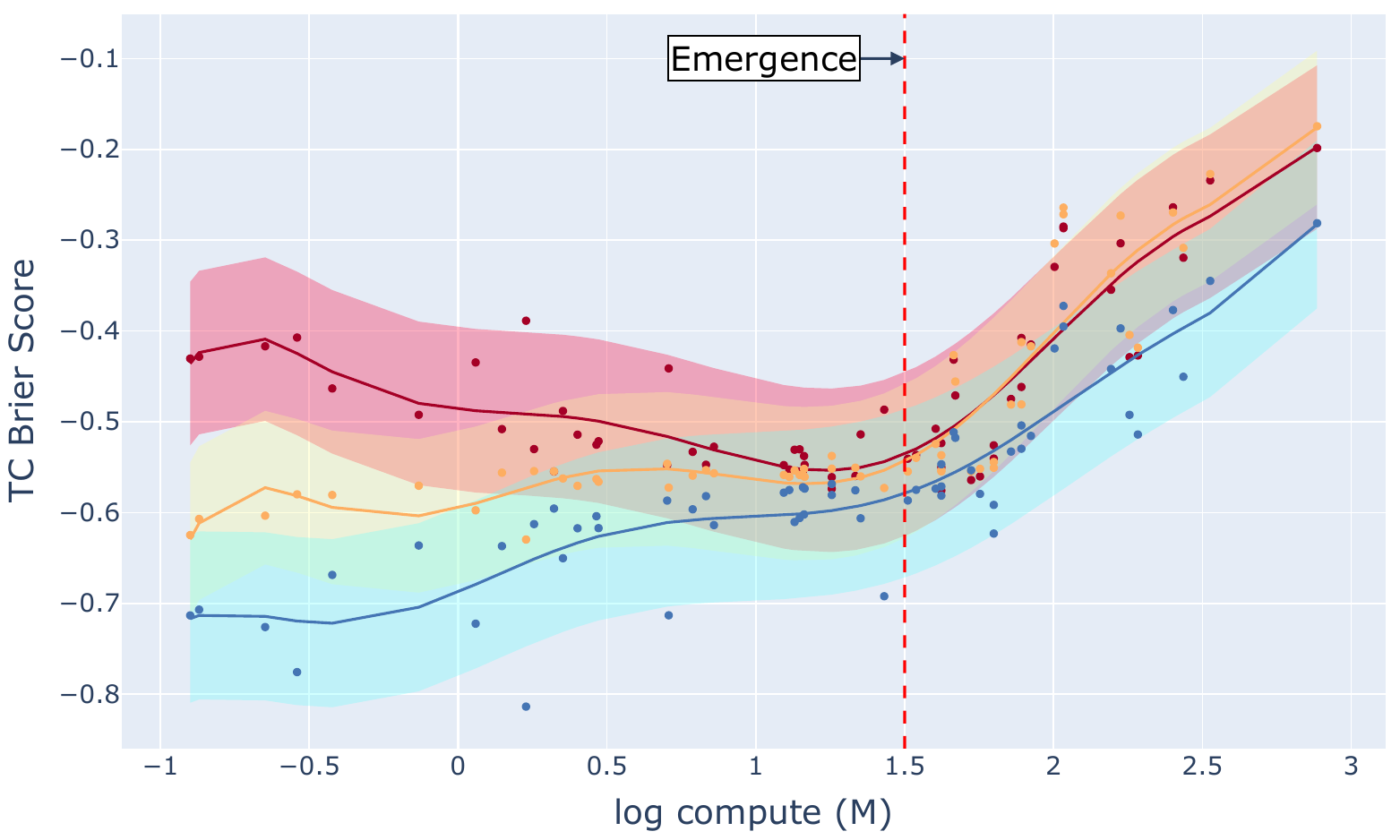}
    \caption{MMLU.}
    \label{subfig: mmlu-n3-6datasets}
  \end{subfigure}
  \hfill
  \begin{subfigure}{0.32\linewidth}
    \includegraphics[width=\textwidth]{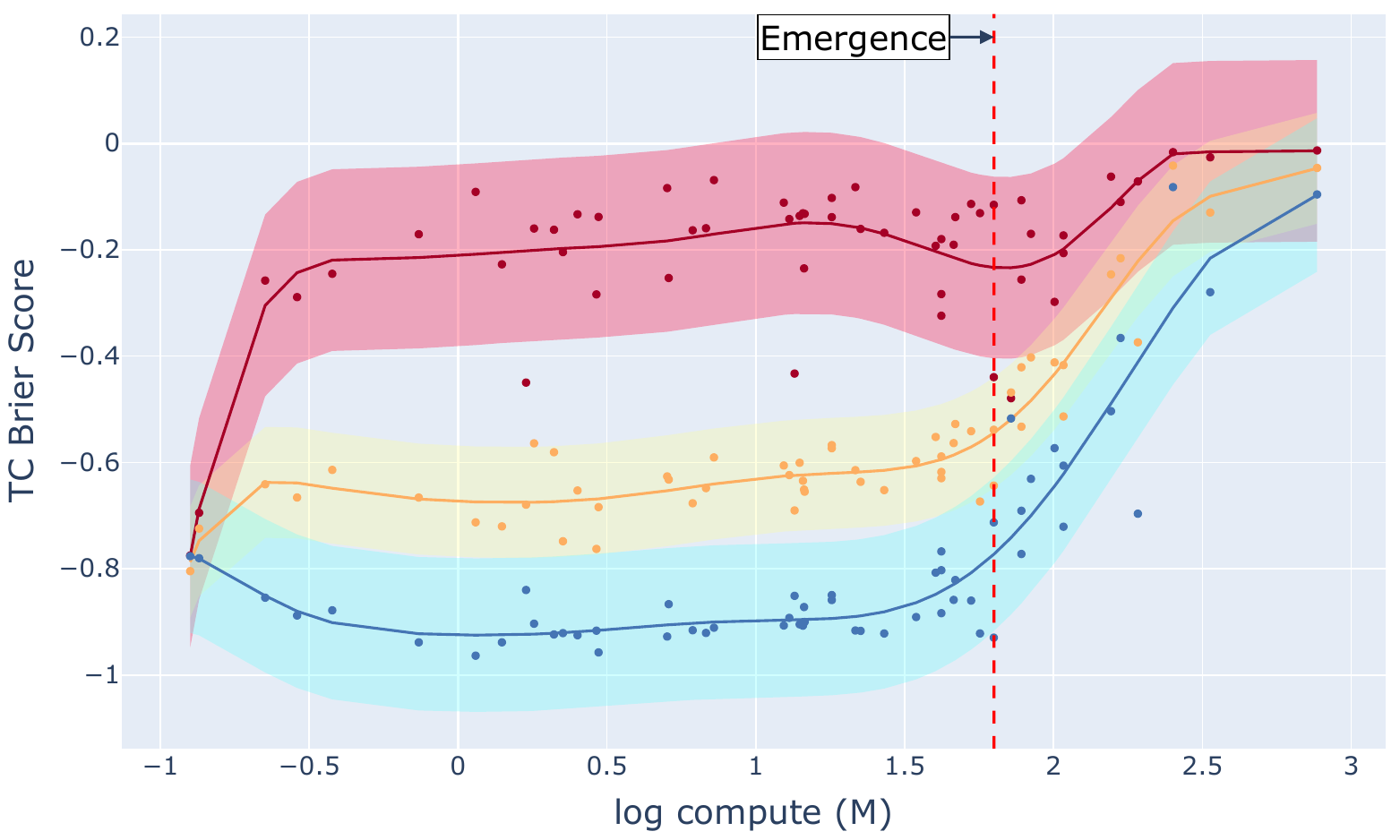}
    \caption{Arithmetic.}
    \label{subfig: arithmetic-n3-6datasets}
  \end{subfigure}
  \hfill
  \begin{subfigure}{0.32\linewidth}
    \includegraphics[width=\textwidth]{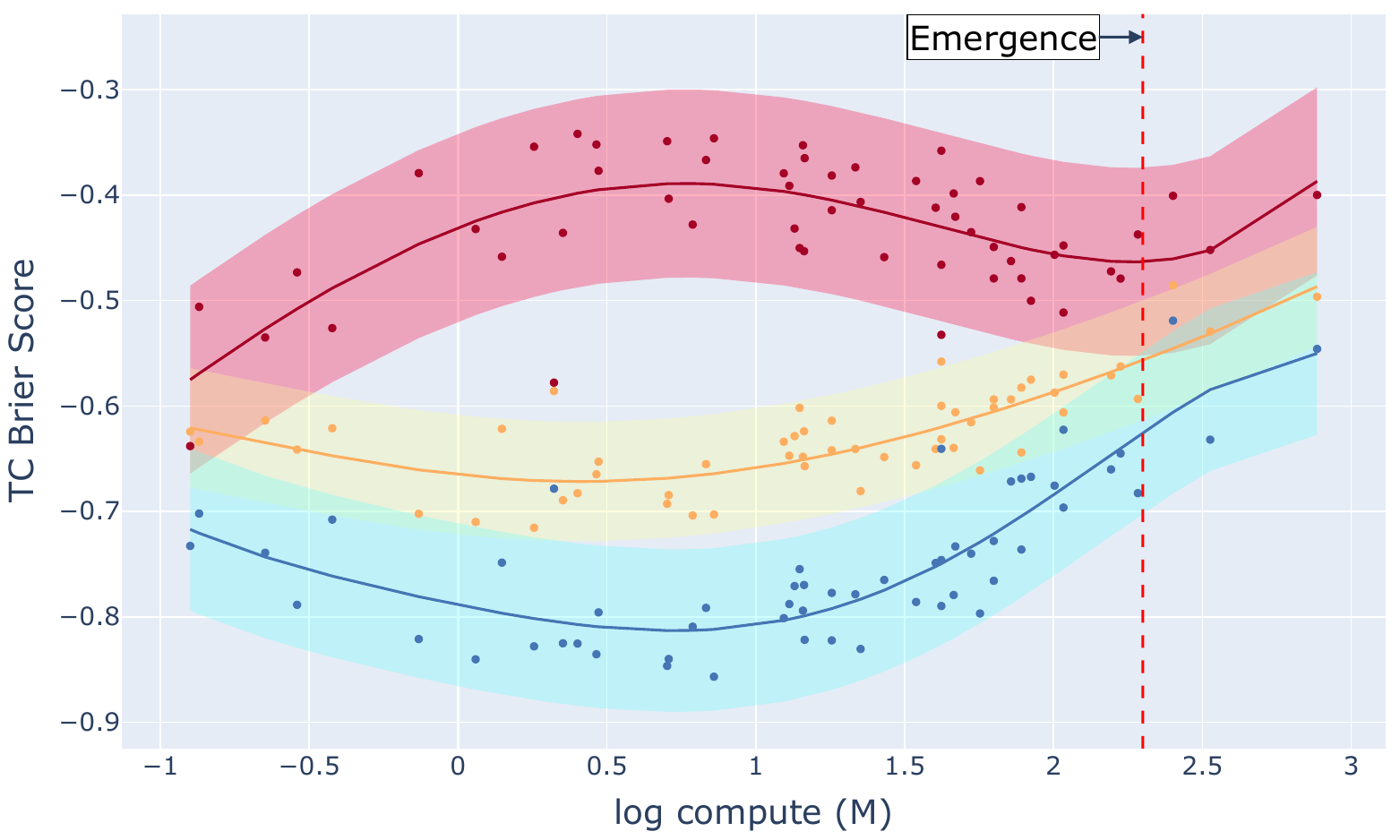}
    \caption{Persian-QA.}
    \label{subfig: persian-qa-n3-6datasets}
  \end{subfigure}
  \hfill
  \begin{subfigure}{0.32\linewidth}
    \includegraphics[width=\textwidth]{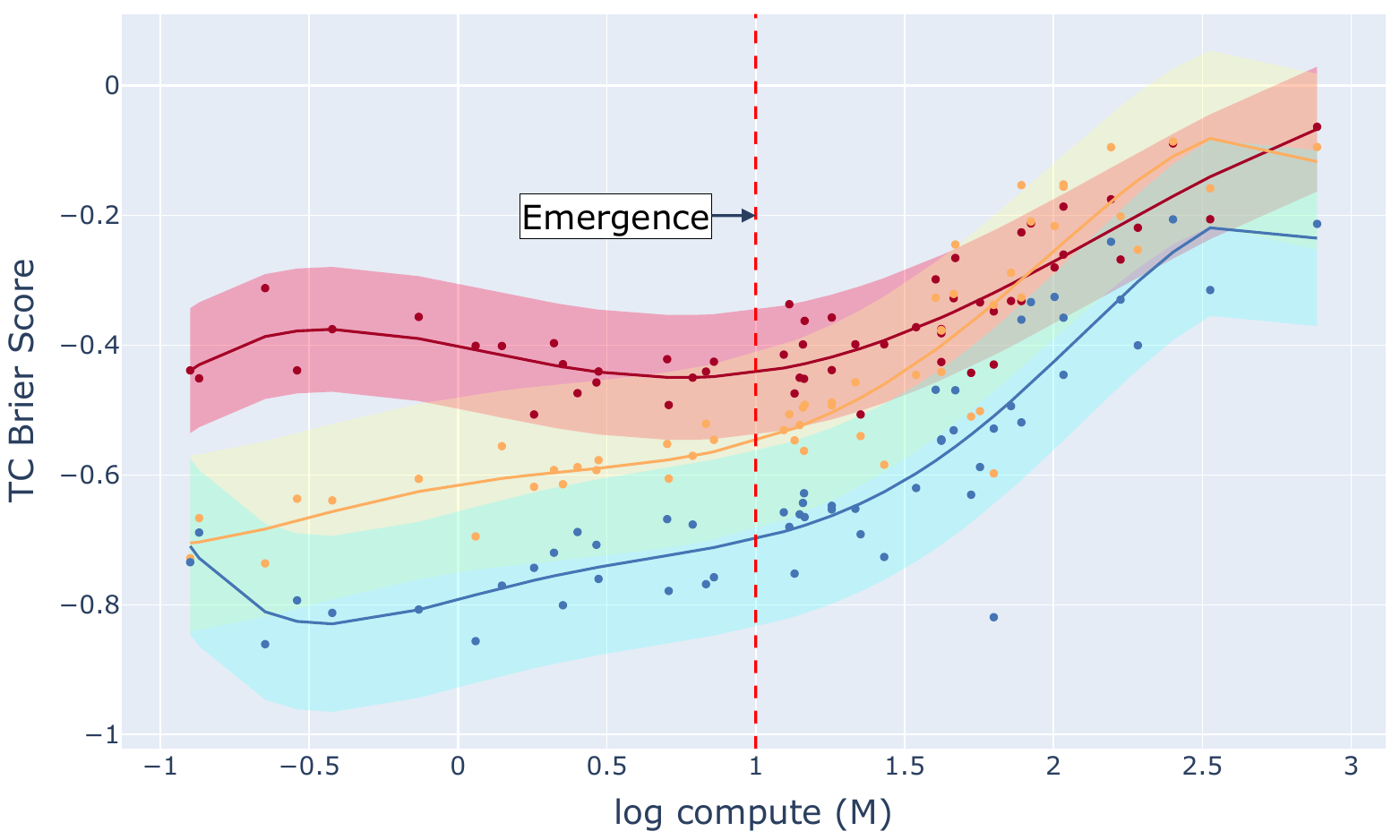}
    \caption{Hindu knowledge.}
    \label{subfig: hindu_knowledge-n3-6datasets}
  \end{subfigure}
  \hfill
  \begin{subfigure}{0.32\linewidth}
    \includegraphics[width=\textwidth]{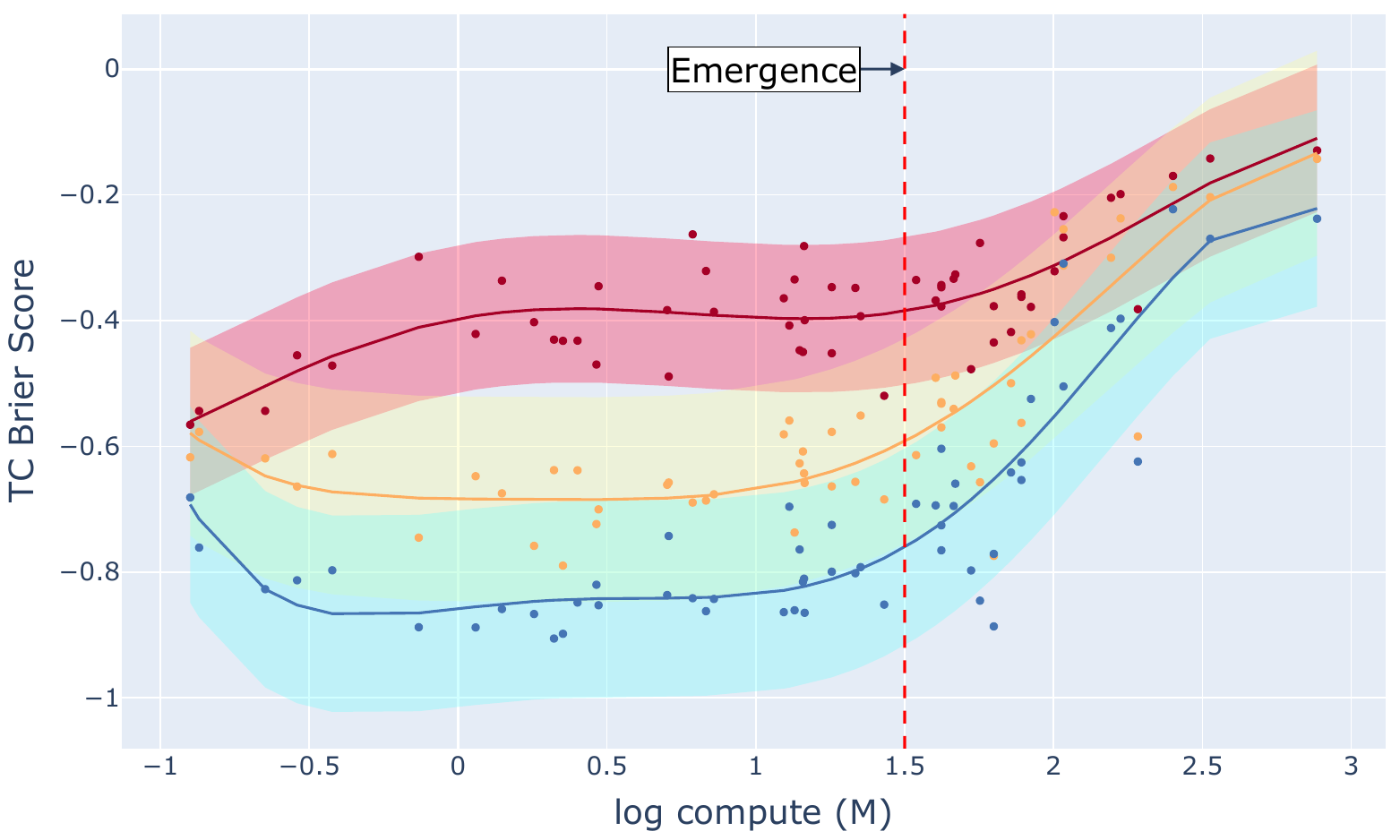}
    \caption{Conceptual combinations.}
    \label{subfig: conceptual_combinations-n3-6datasets}
  \end{subfigure}
  \hfill
  \begin{subfigure}{0.32\linewidth}
    \includegraphics[width=\textwidth]{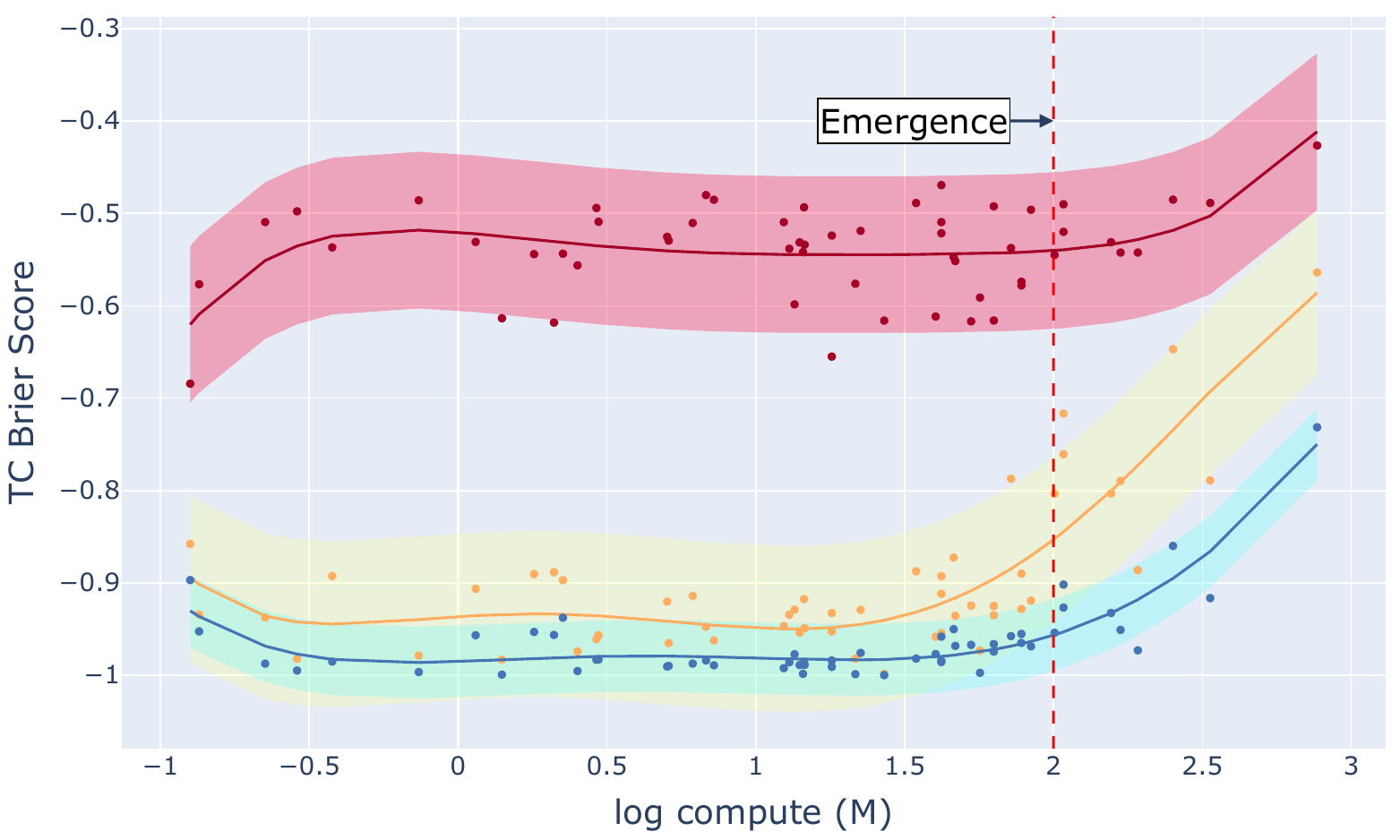}
    \caption{Analogical similarity.}
    \label{subfig: analogical_similarity-n3-6datasets}
  \end{subfigure}
  \caption{U-shaped and inverted-U scaling on 6 datasets exhibiting emergent abilities, with group number $G=3$. Except for MMLU, the other 5 tasks are from Big-Bench. Different levels of U-shaped and inverted-U scaling trends are demonstrated across the 6 tasks.}
  \label{fig: phenomena-n3-6datasets}
\end{figure*}

Fig.~\ref{subfig: mmlu accuracy}--\ref{subfig: arithmetic accuracy} show the scaling trend of accuracy on the MMLU, Persian-QA, and arithmetic datasets, with clear ability emergence demonstrated. Fig.~\ref{subfig: mmlu brier}--\ref{subfig: arithmetic brier} show the scaling trend of TC Brier Score. Though Persian-QA has a smooth scaling, MMLU and arithmetic still exhibit a sharp increase past the emergence threshold $T$. Fig.~\ref{subfig: mmlu phenomenon}--\ref{subfig: arithmetic phenomenon} show the TC Brier Score scaling trend with group number $G=10$. Implementation details and model details are in App.~\ref{sup: implementation details}. Model performance on easier questions, such as difficulty level 1 in Fig.~\ref{subfig: mmlu phenomenon} and Fig.~\ref{subfig: persian-qa phenomenon}, displays an inverted-U shape followed by steady improvement, i.e., performance first increases and then worsens with scale, followed by a second ascent, aligning with the previously reported deep double descent\footnote{The term ``descent'' in the original paper refers to the testing loss. Hence, for the sense of accuracy or Brier Score, it is an ``ascent'' of performance.} on testing loss~\citep{nakkiran2021deep}. Moreover, the reversion from inverse scaling to standard scaling roughly coincides with $T$. In contrast, performance on hard questions, such as difficulty level 10 in Fig.~\ref{subfig: persian-qa phenomenon} and Fig.~\ref{subfig: arithmetic phenomenon}, displays a U-shaped scaling trend~\citep{wei-etal-2023-inverse, mckenzie2023inverse}: model performance decreases with scale in early stage and increases with scale when $M$ gets larger. Besides MMLU, arithmetic, and Persian-QA, Fig.~\ref{fig: phenomena-n3-6datasets} shows U-shaped vs. inverted-U scaling of Hindu knowledge, conceptual combinations, and analogical similarity datasets in Big-Bench~\citep{srivastava2023beyond}, totaling 6 datasets, with $G=3$. Detailed results of the 3 datasets are in App.~\ref{sup: more ppp}.

Overall, the scaling trend of a question group transitions from that of the easiest group (inverted-U followed by steady ascent) to that of the hardest group (U-shape). As the initial scaling trends of easy questions and hard questions roughly offset each other when aggregated across all difficulty levels, performance stagnates until the scaling trend on easy questions reverts from inverse to standard scaling, followed by a sharp improvement when performance on easy and hard questions both improve with scale. This could explain the emergent ability phenomenon reported in previous literature~\citep{wei2022emergent, schaeffer2024emergent, hu2023predicting}. More results of scaling trend on 3 non-emergent tasks in App.~\ref{sup: ppp on nonemergent}, and U-shaped vs. inverted-U scaling measured by accuracy are in App.~\ref{sup: ppp on accuracy}.
\section{Possible Explanation for U-shaped and Inverted-U Scaling}
\label{sec: ppp explain}
We provide a possible explanation for the initially opposing scaling trends (inverted-U vs. U-shaped) on easy vs. hard questions using the AI community's previous findings~\citep{nakkiran2021deep, wei-etal-2023-inverse, mckenzie2023inverse} in deep neural network and LLM behaviors.

\subsection{Scaling Trend of Easy Question Groups}
\label{sec5-sub: easy}
\begin{figure*}[tb]
  \centering
  \begin{subfigure}{0.6\linewidth}
    \includegraphics[width=\textwidth]{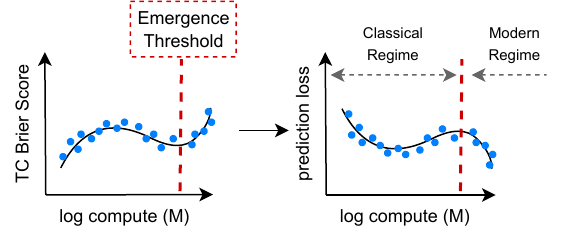}
    \caption{Deep double descent.}
    \label{subfig: ddd}
  \end{subfigure}
  \hfill
  \begin{subfigure}{0.25\linewidth}
    \includegraphics[width=\textwidth]{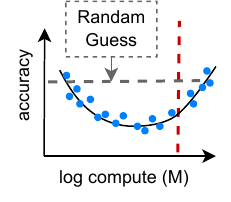}
    \caption{U-shaped scaling.}
    \label{subfig: ushape}
  \end{subfigure}
    \caption{Illustration of deep double descent~\citep{nakkiran2021deep} on easy question groups and U-shaped scaling~\citep{wei-etal-2023-inverse} on hard question groups under the TC Brier Score.}
  \label{fig: ddd-and-ushape}
\end{figure*}
As discussed in Sec.~\ref{sec: phenomenon}, for a downstream task with emergent abilities, model performance on easy questions displays an inverted-U shape followed by steady improvement. Fig.~\ref{subfig: ddd} illustrates the scaling trend if flipping the sign on the TC Brier Score so that a higher number means a higher prediction loss. The pattern is consistent with the phenomenon \textit{deep double descent} identified in \citet{nakkiran2021deep}. In the context of testing error scaling law, \citet{nakkiran2021deep} argues that initially, the bias-variance trade-off in classical statistical learning theory~\citep{hastie2009elements} applies, which forms the ``classical regime'': complex models suffer from ``overfitting'' and thus, once complexity exceeds a certain threshold, models become over-sensitive to sample noises, and the effect from such larger variance dominates the effect of further reducing testing error. On the other hand, once the model is large enough (the ``modern regime''), a further increase in complexity allows the model to pick from more and more interpolating models that all fit the dataset, thus improving performance and reducing testing error to near zero.

\subsection{Scaling Trend on Hard Question Group}
\label{sec5-sub: hard}
\begin{table}[tb]
  \caption{Examples of an easy and hard question in the MMLU benchmark. The Avg. Prob. is the average output probabilities before re-distribution over all models with log compute $M < 1.5$. Answer choices (classes) are underlined. In the hard question, small models overlook the negation ``doesn't'', giving choice C a high confidence, yet correct choice D a low confidence.}
  \label{tab: mmlu question example}
  \centering
  \addtolength{\tabcolsep}{4pt}
  \begin{tabular}{@{}p{0.38\textwidth}|p{0.16\textwidth}|p{0.22\textwidth}|p{0.1\textwidth}@{}}
    \toprule
    Question Description & Difficulty Level & Choices & Avg. Prob.\\
    \midrule
    \makecell[l]{\parbox{5.5cm}{(conceptual physics, id = 44)\\The second law of thermodynamics tells us that heat doesn’t flow from}}  & \makecell[l]{level 10\\(hardest group)} & \makecell[l]{A. hot to cold ever\\B. cold to hot ever\\C. hot to cold \\without external energy\\ \underline{D. cold to hot} \\ \underline{without external energy}} & \makecell[l]{A. 0.24\\B. 0.29\\C. 0.29\\D. 0.18}\\
    \midrule
    \makecell[l]{\parbox{5.5cm}{(global facts, id = 66)\\ In 1935 roughly how many Americans were in favor of Social Security act?}} & \makecell[l]{level 1\\(easiest group)} & \makecell[l]{\underline{A. 90\%} \\ B. 70\% \\ C. 50\% \\ D. 30\%} & \makecell[l]{A. 0.44\\B. 0.30\\C. 0.17\\D. 0.09}\\
    \bottomrule
  \end{tabular}
\end{table}

\citet{mckenzie2023inverse, wei-etal-2023-inverse} have identified the U-shaped scaling of LLM in some downstream tasks, as illustrated in Fig.~\ref{subfig: ushape}. \citet{wei-etal-2023-inverse} provides a potential explanation for the initial inverse scaling: the task might contain a ``distractor task'' that attracts models to learn to solve at first; thus, larger models perform worse. An example is the NeQA task~\citep{mckenzie2023inverse}
, which negates each multiple-choice question in the OpenBookQA dataset~\citep{mihaylov-etal-2018-suit} to examine whether models would be misled by the negation. It turns out that model performance first declines from random guesses due to the attempt to answer the non-negation part of the question. Table~\ref{tab: mmlu question example} shows such a question in our hard question group in the MMLU. For the question ``The second law of thermodynamics tells us that heat doesn’t flow from'', small models ($M < 1.5$) on average assign high confidence to choice C and lowest confidence to correct choice D, where the former is the answer if removing negation ``doesn’t'' from the original question. More MMLU questions to demonstrate U-shaped and inverted-U scaling trends w.r.t. model log compute are in App.~\ref{sup: more question results}.
\section{Slice-and-Sandwich}
\label{sec: method}
\subsection{Problem Formulation}
We aim to predict the performance soar of traditional metrics before it happens. Specifically, we only use models before the emergent threshold $T$ to forecast the incidence of emergent abilities and the scaling trend after $T$. We compare the performance of our pipeline with the current iconic baseline of the Sigmoid-based task-specific scaling law~\citep{ye-etal-2023-predictable}, which uses the Sigmoid function to regress accuracy w.r.t. log compute $M$.

\subsection{Pipeline Overview}
\begin{figure*}[tb]
  \centering
  \includegraphics[width=0.75\textwidth]{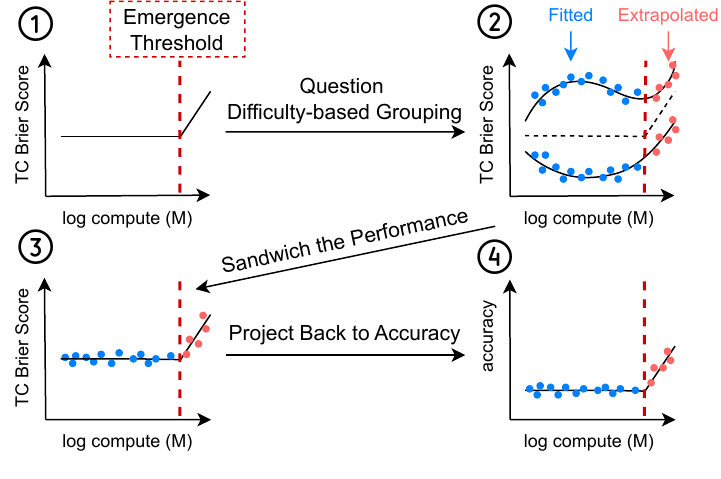}
  \caption{The overall pipeline of \textit{Slice-and-Sandwich}. We group questions into different difficulty levels, fit each group's scaling trend, sandwich the overall performance to construct the scaling law on the linear metric, and finally project the scaling law back to the traditional metric.}
  \label{fig: main framework}
\end{figure*}
Fig.~\ref{fig: main framework} shows the overall pipeline of \textit{Slice-and-Sandwich}. We use models smaller than the emergence threshold $T$ as the training set. As performance no longer stagnates with scale once we group questions by difficulty level, we fit the scaling trend of a continuous metric (TC Brier score in this paper) on the easiest question group and hardest question group separately and use the fitted scaling trend to forecast performance (measured in TC Brier Score) on easy and hard questions past $T$. We also use the training set to regress accuracy on the TC Brier Score and then use this estimated relation to convert the predicted TC Brier Score into predicted accuracy for models past the $T$.

\subsection{Predicting Emergent Ability}
\subsubsection{Question Grouping}
To reduce data noise, we group questions into $G=3$ difficulty levels, as in Fig.~\ref{fig: phenomena-n3-6datasets}, for \textit{Slice-and-Sandwich} and denote the level 1, 2 and 3 as easy, medium, and hard question groups. The medium group's pattern is close to aggregating the scaling trend between easier and harder groups.

\subsubsection{Fitting and Forecasting Scaling Trend of Easy vs. Hard Questions}
\label{sec3-subsub: fitting-easy-and-hard}
We use simple polynomial regression to fit the scaling trend of the TC Brier Score of the easy and hard question groups separately using models before $T$. We denote by $F^{c}_{e}(x)$ and $F^{c}_{h}(x)$ the fitted scaling trend of the easy and hard question groups, respectively, where $x$ is the log compute. We then use $F^{c}_{e}(x)$ and $F^{c}_{h}(x)$ to forecast performance (measured in TC Brier Score) on the easy and the hard question groups of models with log compute $x$ above $T$.

We use the average of performance on the easy group and the hard group to forecast aggregated performance measured in TC Brier Score:
\begin{equation}
  F^{c}(x)=\frac{1}{2}(F^{c}_{e}(x)+F^{c}_{h}(x)),
  \label{eq: squeeze}
\end{equation}
as aggregate performance is sandwiched between performances in the easy and hard groups.

\subsubsection{Obtaining Scaling Trend in Traditional Metric}
\label{sec3-subsub: project}
Since people usually care about intuitive traditional metrics such as accuracy~\citep{hu2023predicting}, our last step is to project the fitted scaling trend in TC Brier Score, $F^{c}(x)$, back to scaling trend in accuracy, denoted by $F^{t}(x)$. One can replace TC Brier Score with other continuous metrics and accuracy with other traditional metrics. Specifically, we first estimate the relation between the continuous metric (TC Brier Score) and the traditional metric (accuracy) using models with log computes smaller than $T$ as the training set. We denote the estimated mapping from TC Brier Score to accuracy as $G(\cdot)$. Our forecast of scaling trend of accuracy is given by:
\begin{equation}
  F^{t}(x)=G(F^{c}(x))+C,
  \label{eq: project}
\end{equation}
where $C$ is a constant such that the average predicted accuracy of $F^{t}(x)$ on the training set is the same as the average true accuracy of all models in the training set.
\section{Experiments}
\label{sec: exp}
\subsection{Fitting Scaling Trend of Easy Group and Hard Group}
\begin{figure*}[tb]
  \centering
  \begin{subfigure}{0.3\linewidth}
    \includegraphics[width=\textwidth]{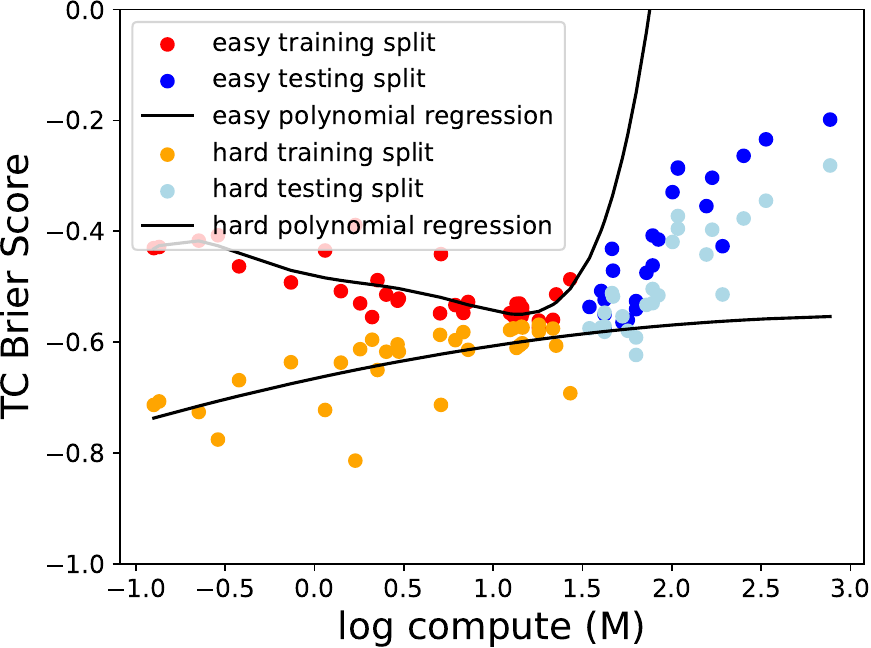}
    \caption{MMLU.}
    \label{subfig: mmlu-easy-hard-fit}
  \end{subfigure}
  \hfill
  \begin{subfigure}{0.3\linewidth}
    \includegraphics[width=\textwidth]{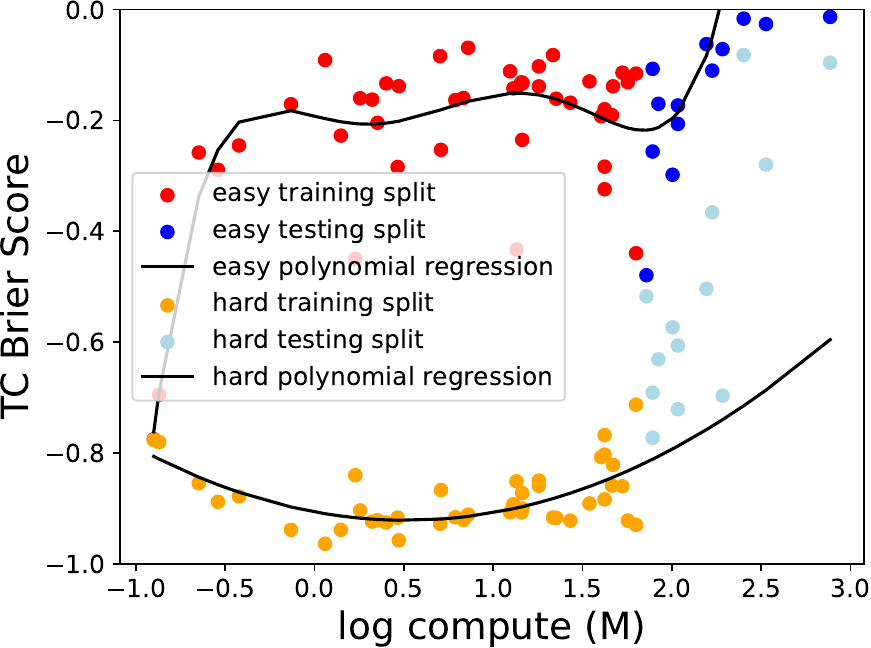}
    \caption{Arithmetic.}
    \label{subfig: arithmetic-easy-hard-fit}
  \end{subfigure}
  \hfill
  \begin{subfigure}{0.3\linewidth}
    \includegraphics[width=\textwidth]{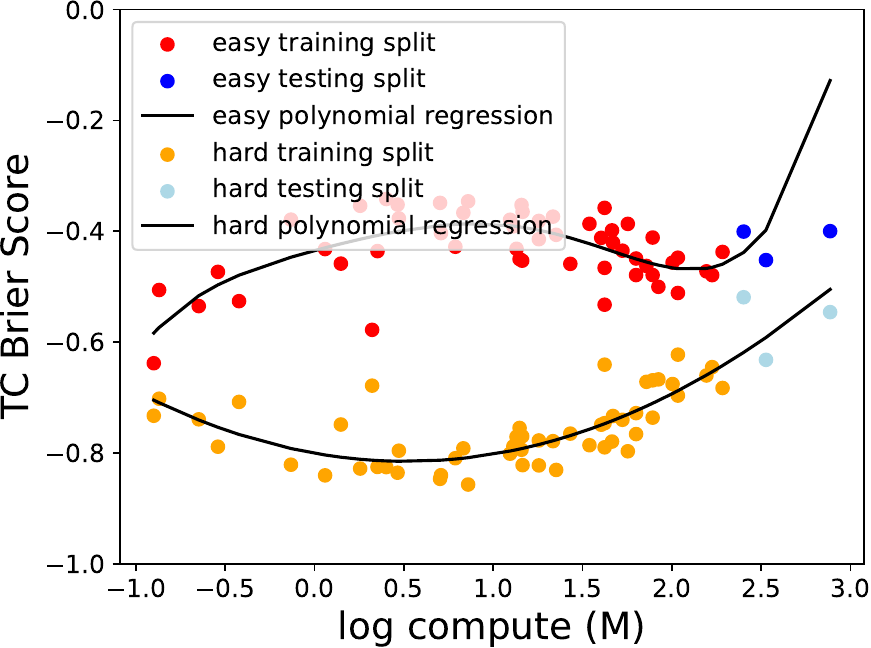}
    \caption{Persian-QA.}
    \label{subfig: persian-qa-easy-hard-fit}
  \end{subfigure}
  \caption{Data and polynomial fit for the easy and hard question groups. The fitted trends encapsulate the actual trends.}
  \label{fig: easy-hard-brier-fit}
\end{figure*}
\begin{figure*}[tb]
  \centering
  \begin{subfigure}{0.3\linewidth}
    \includegraphics[width=\textwidth]{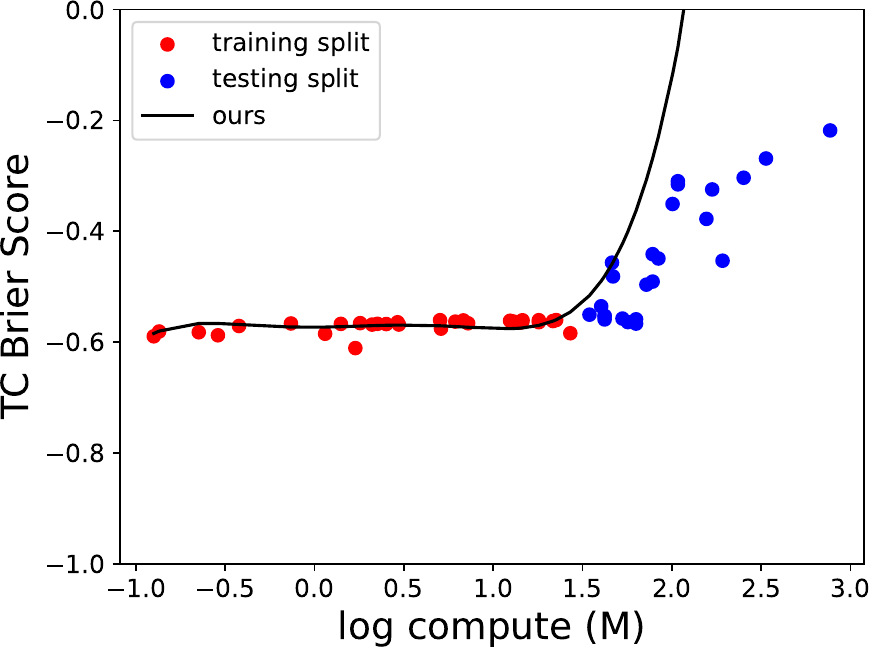}
    \caption{MMLU.}
    \label{subfig: mmlu-brier-scale}
  \end{subfigure}
  \hfill
  \begin{subfigure}{0.3\linewidth}
    \includegraphics[width=\textwidth]{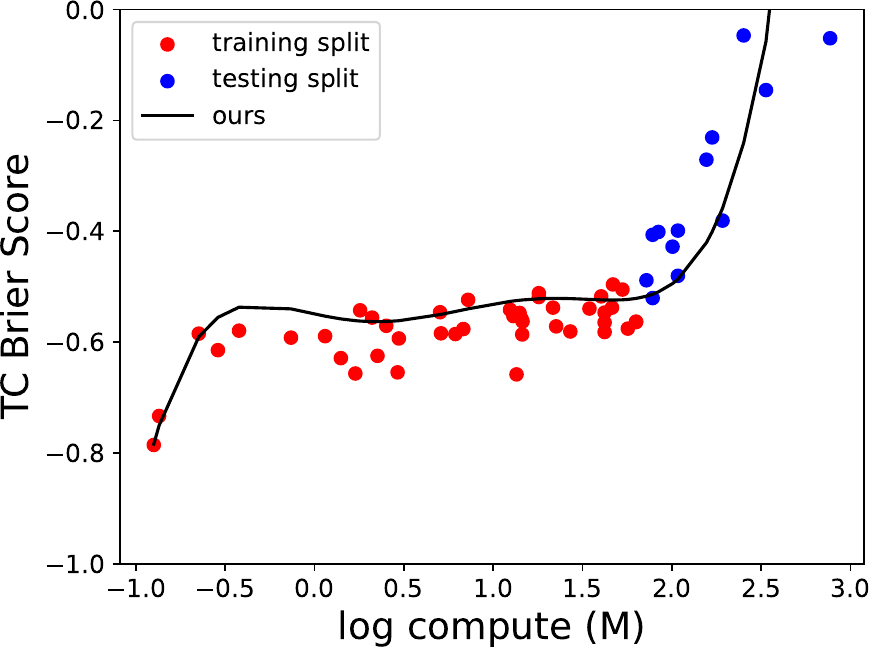}
    \caption{Arithmetic.}
    \label{subfig: arithmetic-brier-scale}
  \end{subfigure}
  \hfill
  \begin{subfigure}{0.3\linewidth}
    \includegraphics[width=\textwidth]{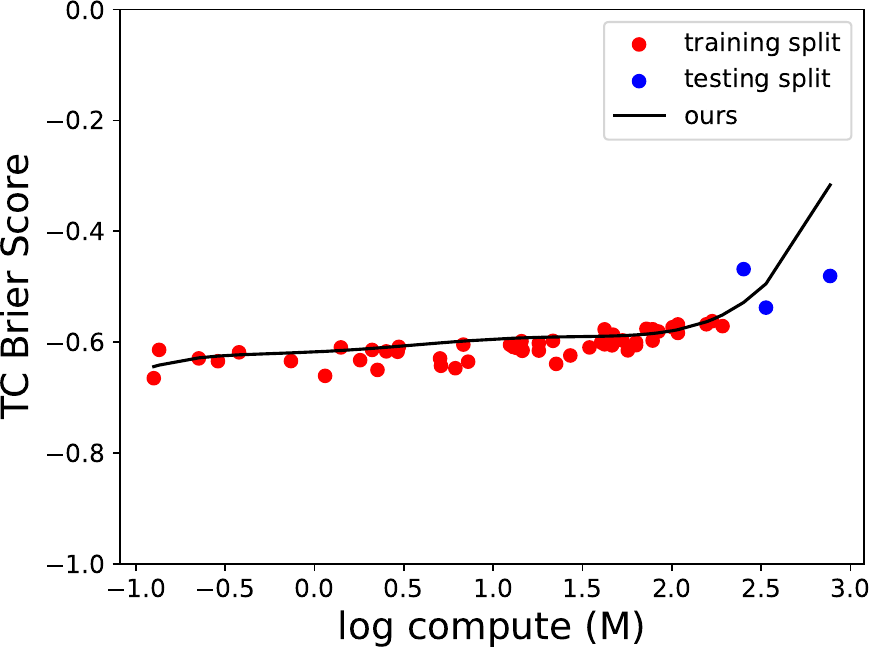}
    \caption{Persian-QA.}
    \label{subfig: persian-qa-brier-scale}
  \end{subfigure}
  \caption{The TC-Brier-Score-based scaling law acquired by taking the average of fitted trends of easy and hard question groups in Fig.~\ref{fig: easy-hard-brier-fit}.}
  \label{fig: brier-scaling-law}
\end{figure*}
We adopt polynomial degree=2 and 5 for hard and easy questions, respectively, in response to our observation of U-shaped vs. inverted-U scaling. This parameter selection is based on our prior belief of polynomial regression's fitting powers to fit the deep double descent and U-shaped scaling. Experimental results on parameter robustness are in App.~\ref{sup: more dis on sas}. 

Fig.~\ref{fig: easy-hard-brier-fit} shows the fitted scaling trend of the easy and hard question groups on the MMLU, arithmetic, and Persian-QA datasets. Empirically, fitted trends on hard questions are either precise or underestimated, e.g., hard group of MMLU (Fig.~\ref{subfig: mmlu-easy-hard-fit}) and arithmetic (Fig.~\ref{subfig: arithmetic-easy-hard-fit}), due to lower fitting power of degree 2; fitted trends on easy questions are precise or overestimated, e.g., easy group of MMLU (Fig.~\ref{subfig: mmlu-easy-hard-fit}), due to a more considerable fitting power of degree 5. However, they still encapsulate the overall trend. Therefore, as shown in Fig.~\ref{fig: brier-scaling-law}, taking their average can decrease the deviation and still lead to a precise prediction of the actual scaling trend.

\subsection{Relation Between Accuracy and Target-Conditioned (TC) Brier Score}
\begin{figure*}[tb]
  \centering
  \begin{subfigure}{0.3\linewidth}
    \includegraphics[width=\textwidth]{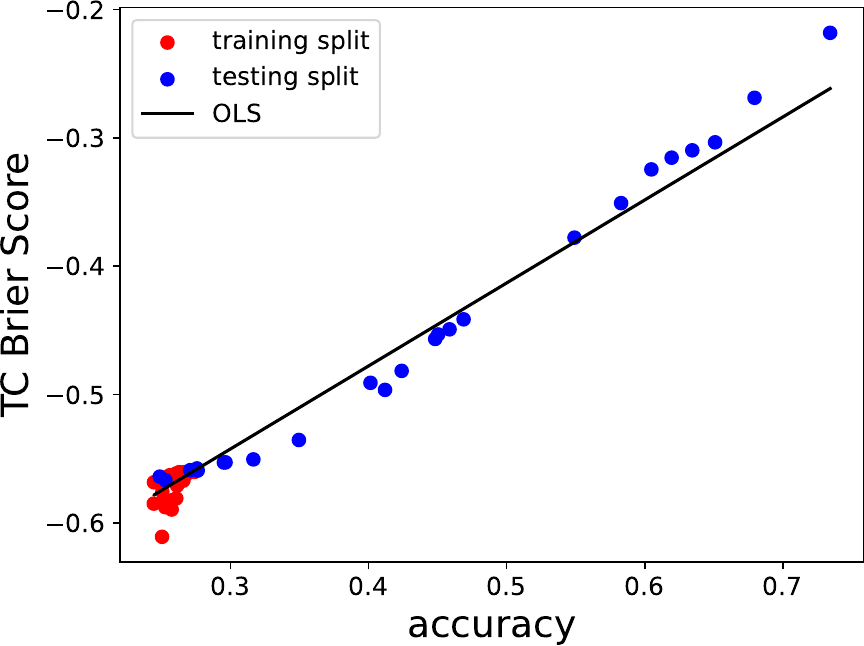}
    \caption{MMLU.}
    \label{subfig: mmlu-rel}
  \end{subfigure}
  \hfill
  \begin{subfigure}{0.3\linewidth}
    \includegraphics[width=\textwidth]{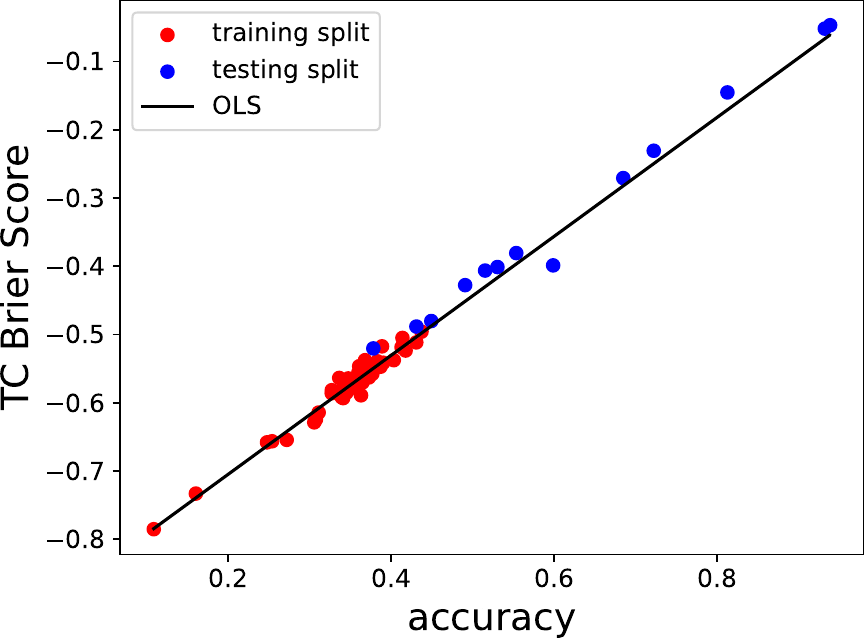}
    \caption{Arithmetic.}
    \label{subfig: arithmetic-rel}
  \end{subfigure}
  \hfill
  \begin{subfigure}{0.3\linewidth}
    \includegraphics[width=\textwidth]{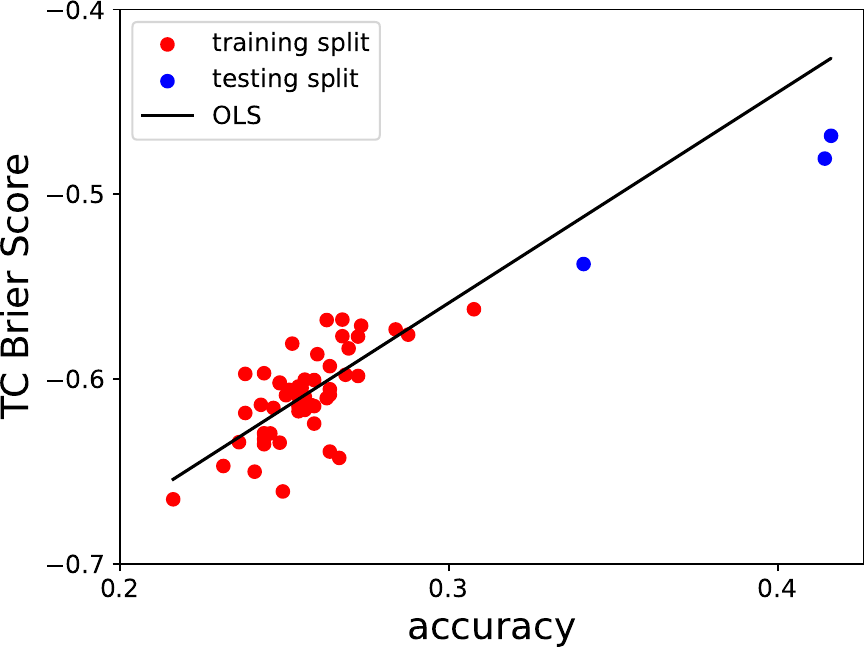}
    \caption{Persian-QA.}
    \label{subfig: persian-qa-rel}
  \end{subfigure}
  \caption{The relation between accuracy and the TC Brier Score. The mapping function $G(x)$ from the TC Brier Score to accuracy can be well-modeled using small models as the training set.}
  \label{fig: redist-brier-acc-relation}
\end{figure*}
Fig.~\ref{fig: redist-brier-acc-relation} shows the close and almost linear relation between accuracy and TC Brier Score. As a result, simple ordinary least squares (OLS) regression using only models before the emergence threshold yields a precise mapping $G(\cdot)$ from TC Brier Score to accuracy.

\subsection{Forecasting Scaling Trend in Accuracy}
\label{sec5-subsub: project}
\begin{figure*}[tb]
  \centering
  \begin{subfigure}{0.3\linewidth}
    \includegraphics[width=\textwidth]{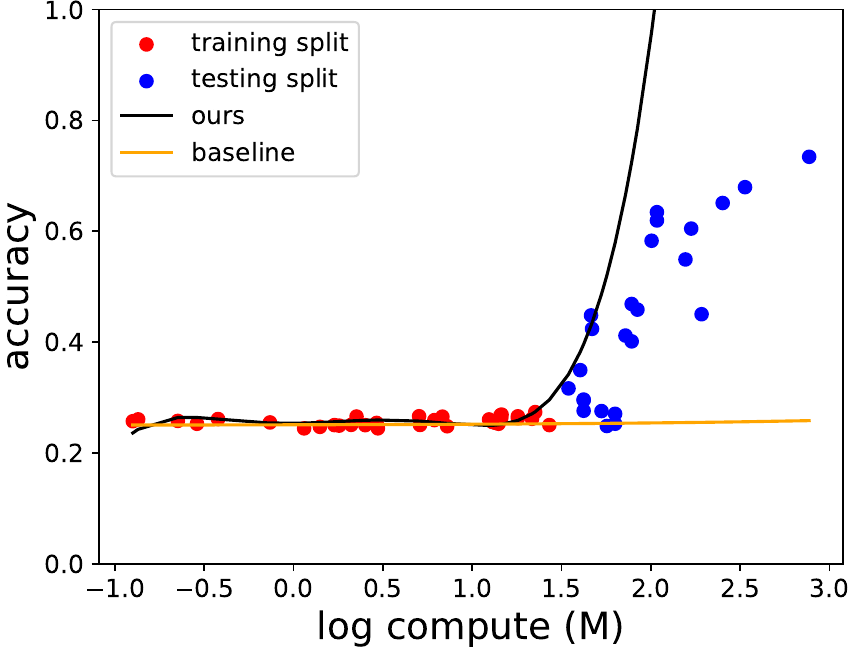}
    \caption{MMLU.}
    \label{subfig: mmlu-acc-scale}
  \end{subfigure}
  \hfill
  \begin{subfigure}{0.3\linewidth}
    \includegraphics[width=\textwidth]{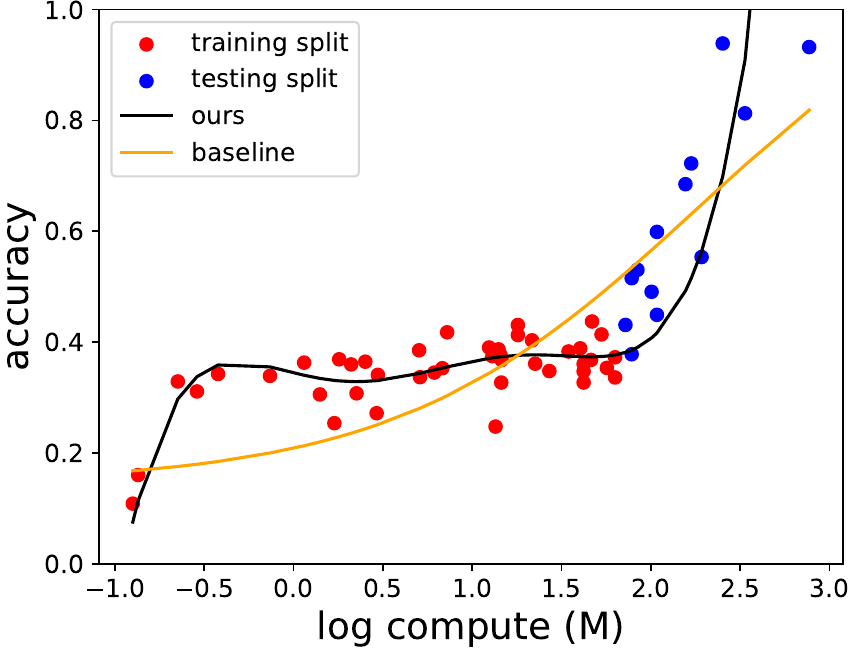}
    \caption{Arithmetic.}
    \label{subfig: arithmetic-acc-scale}
  \end{subfigure}
  \hfill
  \begin{subfigure}{0.3\linewidth}
    \includegraphics[width=\textwidth]{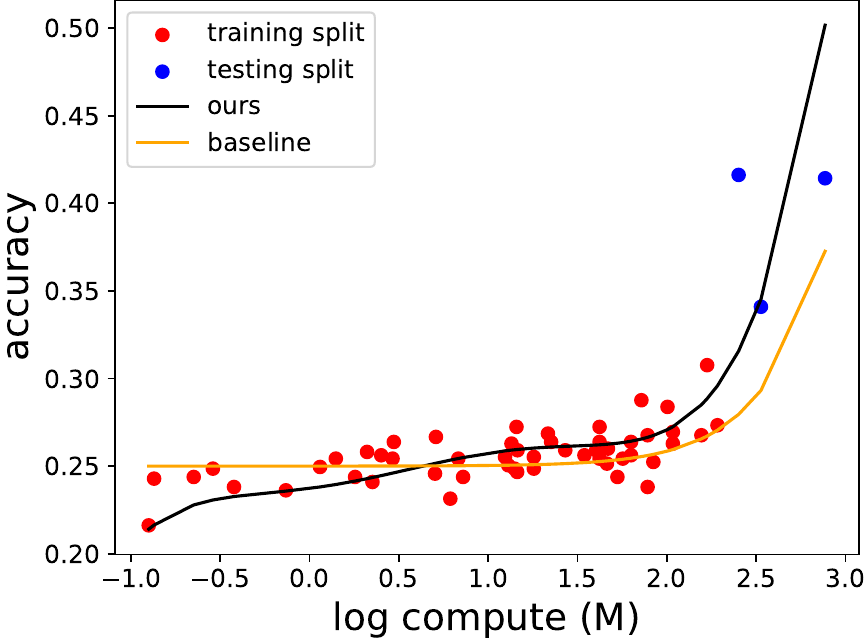}
    \caption{Persian-QA.}
    \label{subfig: persian-qa-acc-scale}
  \end{subfigure}
  \caption{The accuracy-based scaling law acquired by projecting TC-Brier-Score-based scaling law back to accuracy-based scaling law by $G(x)$. Baseline is Sigmoid-based regression~\citep{owen2024predictable}.}
  \label{fig: acc-scaling-law}
\end{figure*}
Fig.~\ref{fig: acc-scaling-law} shows the accuracy-based scaling trend, $F^{t}(x)$, obtained by Eq.~\ref{eq: project} with $G(\cdot)$, together with the baseline of fitting the accuracy on the Sigmoid function~\citep{owen2024predictable, ruan2024observational}. Compared with the baseline, \textit{Slice-and-Sandwich} better predicts and estimates the soaring performance by baking in more priors of the observed U-shaped and inverted-U scaling. For the MMLU, our approach captures the forthcoming soaring trend, whereas the baseline approach does not. For the arithmetic dataset, though the baseline provides a seemingly decent forecast, it does not capture the acceleration of performance increase, whereas our approach does. In short, our \textit{Slice-and-Sandwich} approach is more explainable and capable of capturing the soaring trends of emergent abilities. A simple alternative method of \textit{Slice-and-Sandwich} and its experimental results are in App.~\ref{sup: more dis on sas}.
\section{Conclusions and Limitations}
\label{sec: conclusion}
This work analyzes the LLM task-specific scaling law by grouping questions according to difficulty level. For six multiple-choice tasks with emergent abilities, we demonstrate U-shaped scaling for hard questions and inverted-U scaling followed by steady improvement for easy questions. These findings provide insight into the potential causes of emergent abilities. We then introduce the \textit{Slice-and-Sandwich} pipeline to predict the emergence threshold and scaling law thereafter. However, as emergent phenomena are common across LLM benchmarks, it might be hard to claim all of them show clear U-shaped vs. inverted-U scaling. Furthermore, our focus is on multiple-choice tasks; applying our method to string-matching tasks requires identifying a continuous metric that differentiates easy questions from hard questions and correlates with the interested traditional metric. We illustrate and discuss this in App.~\ref{sup: ppp on string match}, highlighting future research direction.
\clearpage



\bibliography{reference}

\begin{thebibliography}{54}
\providecommand{\natexlab}[1]{#1}
\providecommand{\url}[1]{\texttt{#1}}
\expandafter\ifx\csname urlstyle\endcsname\relax
  \providecommand{\doi}[1]{doi: #1}\else
  \providecommand{\doi}{doi: \begingroup \urlstyle{rm}\Url}\fi

\bibitem[Achiam et~al.(2023)Achiam, Adler, Agarwal, Ahmad, Akkaya, Aleman, Almeida, Altenschmidt, Altman, Anadkat, et~al.]{achiam2023gpt}
Josh Achiam, Steven Adler, Sandhini Agarwal, Lama Ahmad, Ilge Akkaya, Florencia~Leoni Aleman, Diogo Almeida, Janko Altenschmidt, Sam Altman, Shyamal Anadkat, et~al.
\newblock Gpt-4 technical report.
\newblock \emph{arXiv preprint arXiv:2303.08774}, 2023.

\bibitem[Almazrouei et~al.(2023)Almazrouei, Alobeidli, Alshamsi, Cappelli, Cojocaru, Debbah, Goffinet, Hesslow, Launay, Malartic, et~al.]{almazrouei2023falcon}
Ebtesam Almazrouei, Hamza Alobeidli, Abdulaziz Alshamsi, Alessandro Cappelli, Ruxandra Cojocaru, M{\'e}rouane Debbah, {\'E}tienne Goffinet, Daniel Hesslow, Julien Launay, Quentin Malartic, et~al.
\newblock The falcon series of open language models.
\newblock \emph{arXiv preprint arXiv:2311.16867}, 2023.

\bibitem[Bai et~al.(2023)Bai, Bai, Chu, Cui, Dang, Deng, Fan, Ge, Han, Huang, et~al.]{bai2023qwen}
Jinze Bai, Shuai Bai, Yunfei Chu, Zeyu Cui, Kai Dang, Xiaodong Deng, Yang Fan, Wenbin Ge, Yu~Han, Fei Huang, et~al.
\newblock Qwen technical report.
\newblock \emph{arXiv preprint arXiv:2309.16609}, 2023.

\bibitem[Biderman et~al.(2023)Biderman, Schoelkopf, Anthony, Bradley, O’Brien, Hallahan, Khan, Purohit, Prashanth, Raff, et~al.]{biderman2023pythia}
Stella Biderman, Hailey Schoelkopf, Quentin~Gregory Anthony, Herbie Bradley, Kyle O’Brien, Eric Hallahan, Mohammad~Aflah Khan, Shivanshu Purohit, USVSN~Sai Prashanth, Edward Raff, et~al.
\newblock Pythia: A suite for analyzing large language models across training and scaling.
\newblock In \emph{Proceedings of the International Conference on Machine Learning (ICML)}, 2023.

\bibitem[Black et~al.(2021)Black, Gao, Wang, Leahy, and Biderman]{gpt-neo}
Sid Black, Leo Gao, Phil Wang, Connor Leahy, and Stella Biderman.
\newblock {GPT-Neo: Large Scale Autoregressive Language Modeling with Mesh-Tensorflow}, March 2021.
\newblock URL \url{https://doi.org/10.5281/zenodo.5297715}.

\bibitem[Brier(1950)]{brier1950verification}
Glenn~W Brier.
\newblock Verification of forecasts expressed in terms of probability.
\newblock \emph{Monthly weather review}, 78\penalty0 (1):\penalty0 1--3, 1950.

\bibitem[Brown(2020)]{brown2020language}
Tom~B Brown.
\newblock Language models are few-shot learners.
\newblock \emph{arXiv preprint arXiv:2005.14165}, 2020.

\bibitem[Charan et~al.(2023)Charan, Chunduri, Anand, and Shukla]{charan2023text}
PV~Charan, Hrushikesh Chunduri, P~Mohan Anand, and Sandeep~K Shukla.
\newblock From text to mitre techniques: Exploring the malicious use of large language models for generating cyber attack payloads.
\newblock \emph{arXiv preprint arXiv:2305.15336}, 2023.

\bibitem[Clark et~al.(2018)Clark, Cowhey, Etzioni, Khot, Sabharwal, Schoenick, and Tafjord]{clark2018think}
Peter Clark, Isaac Cowhey, Oren Etzioni, Tushar Khot, Ashish Sabharwal, Carissa Schoenick, and Oyvind Tafjord.
\newblock Think you have solved question answering? try arc, the ai2 reasoning challenge.
\newblock \emph{arXiv preprint arXiv:1803.05457}, 2018.

\bibitem[Computer(2023)]{together2023redpajama}
Together Computer.
\newblock Redpajama: an open dataset for training large language models, October 2023.
\newblock URL \url{https://github.com/togethercomputer/RedPajama-Data}.

\bibitem[Dai et~al.(2024)Dai, Deng, Zhao, Xu, Gao, Chen, Li, Zeng, Yu, Wu, et~al.]{dai2024deepseekmoe}
Damai Dai, Chengqi Deng, Chenggang Zhao, RX~Xu, Huazuo Gao, Deli Chen, Jiashi Li, Wangding Zeng, Xingkai Yu, Y~Wu, et~al.
\newblock Deepseekmoe: Towards ultimate expert specialization in mixture-of-experts language models.
\newblock \emph{arXiv preprint arXiv:2401.06066}, 2024.

\bibitem[Fawzi et~al.(2022)Fawzi, Balog, Huang, Hubert, Romera-Paredes, Barekatain, Novikov, R~Ruiz, Schrittwieser, Swirszcz, et~al.]{fawzi2022discovering}
Alhussein Fawzi, Matej Balog, Aja Huang, Thomas Hubert, Bernardino Romera-Paredes, Mohammadamin Barekatain, Alexander Novikov, Francisco~J R~Ruiz, Julian Schrittwieser, Grzegorz Swirszcz, et~al.
\newblock Discovering faster matrix multiplication algorithms with reinforcement learning.
\newblock \emph{Nature}, 610\penalty0 (7930):\penalty0 47--53, 2022.

\bibitem[Gadre et~al.(2024)Gadre, Smyrnis, Shankar, Gururangan, Wortsman, Shao, Mercat, Fang, Li, Keh, et~al.]{gadre2024language}
Samir~Yitzhak Gadre, Georgios Smyrnis, Vaishaal Shankar, Suchin Gururangan, Mitchell Wortsman, Rulin Shao, Jean Mercat, Alex Fang, Jeffrey Li, Sedrick Keh, et~al.
\newblock Language models scale reliably with over-training and on downstream tasks.
\newblock \emph{arXiv preprint arXiv:2403.08540}, 2024.

\bibitem[Gao et~al.(2024)Gao, Tow, Abbasi, Biderman, Black, DiPofi, Foster, Golding, Hsu, Le~Noac'h, Li, McDonell, Muennighoff, Ociepa, Phang, Reynolds, Schoelkopf, Skowron, Sutawika, Tang, Thite, Wang, Wang, and Zou]{eval-harness}
Leo Gao, Jonathan Tow, Baber Abbasi, Stella Biderman, Sid Black, Anthony DiPofi, Charles Foster, Laurence Golding, Jeffrey Hsu, Alain Le~Noac'h, Haonan Li, Kyle McDonell, Niklas Muennighoff, Chris Ociepa, Jason Phang, Laria Reynolds, Hailey Schoelkopf, Aviya Skowron, Lintang Sutawika, Eric Tang, Anish Thite, Ben Wang, Kevin Wang, and Andy Zou.
\newblock A framework for few-shot language model evaluation, 07 2024.
\newblock URL \url{https://zenodo.org/records/12608602}.

\bibitem[Geng \& Liu(2023)Geng and Liu]{openlm2023openllama}
Xinyang Geng and Hao Liu.
\newblock Openllama: An open reproduction of llama, May 2023.
\newblock URL \url{https://github.com/openlm-research/open_llama}.

\bibitem[Hastie et~al.(2009)Hastie, Tibshirani, Friedman, and Friedman]{hastie2009elements}
Trevor Hastie, Robert Tibshirani, Jerome~H Friedman, and Jerome~H Friedman.
\newblock \emph{The elements of statistical learning: data mining, inference, and prediction}, volume~2.
\newblock Springer, 2009.

\bibitem[Hendrycks et~al.(2021)Hendrycks, Burns, Basart, Zou, Mazeika, Song, and Steinhardt]{hendryckstest2021}
Dan Hendrycks, Collin Burns, Steven Basart, Andy Zou, Mantas Mazeika, Dawn Song, and Jacob Steinhardt.
\newblock Measuring massive multitask language understanding.
\newblock \emph{Proceedings of the International Conference on Learning Representations (ICLR)}, 2021.

\bibitem[Hoffmann et~al.(2022)Hoffmann, Borgeaud, Mensch, Buchatskaya, Cai, Rutherford, de~Las~Casas, Hendricks, Welbl, Clark, Hennigan, Noland, Millican, van~den Driessche, Damoc, Guy, Osindero, Simonyan, Elsen, Vinyals, Rae, and Sifre]{NEURIPS2022_c1e2faff}
Jordan Hoffmann, Sebastian Borgeaud, Arthur Mensch, Elena Buchatskaya, Trevor Cai, Eliza Rutherford, Diego de~Las~Casas, Lisa~Anne Hendricks, Johannes Welbl, Aidan Clark, Thomas Hennigan, Eric Noland, Katherine Millican, George van~den Driessche, Bogdan Damoc, Aurelia Guy, Simon Osindero, Kar\'{e}n Simonyan, Erich Elsen, Oriol Vinyals, Jack Rae, and Laurent Sifre.
\newblock An empirical analysis of compute-optimal large language model training.
\newblock In \emph{Proceedings of Advances in neural information processing systems (NeurIPS)}, volume~35, 2022.

\bibitem[Hu et~al.(2023)Hu, Liu, Han, Zhang, He, Zhao, Lin, Ding, Ou, Zeng, et~al.]{hu2023predicting}
Shengding Hu, Xin Liu, Xu~Han, Xinrong Zhang, Chaoqun He, Weilin Zhao, Yankai Lin, Ning Ding, Zebin Ou, Guoyang Zeng, et~al.
\newblock Predicting emergent abilities with infinite resolution evaluation.
\newblock In \emph{Proceedings of the International Conference on Learning Representations (ICLR)}, 2023.

\bibitem[Jiang et~al.(2024)Jiang, Sablayrolles, Roux, Mensch, Savary, Bamford, Chaplot, Casas, Hanna, Bressand, et~al.]{jiang2024mixtral}
Albert~Q Jiang, Alexandre Sablayrolles, Antoine Roux, Arthur Mensch, Blanche Savary, Chris Bamford, Devendra~Singh Chaplot, Diego de~las Casas, Emma~Bou Hanna, Florian Bressand, et~al.
\newblock Mixtral of experts.
\newblock \emph{arXiv preprint arXiv:2401.04088}, 2024.

\bibitem[Jumper et~al.(2021)Jumper, Evans, Pritzel, Green, Figurnov, Ronneberger, Tunyasuvunakool, Bates, {\v{Z}}{\'\i}dek, Potapenko, et~al.]{jumper2021highly}
John Jumper, Richard Evans, Alexander Pritzel, Tim Green, Michael Figurnov, Olaf Ronneberger, Kathryn Tunyasuvunakool, Russ Bates, Augustin {\v{Z}}{\'\i}dek, Anna Potapenko, et~al.
\newblock Highly accurate protein structure prediction with alphafold.
\newblock \emph{nature}, 596\penalty0 (7873):\penalty0 583--589, 2021.

\bibitem[Kaddour et~al.(2023)Kaddour, Harris, Mozes, Bradley, Raileanu, and McHardy]{kaddour2023challenges}
Jean Kaddour, Joshua Harris, Maximilian Mozes, Herbie Bradley, Roberta Raileanu, and Robert McHardy.
\newblock Challenges and applications of large language models.
\newblock \emph{arXiv preprint arXiv:2307.10169}, 2023.

\bibitem[Kaplan et~al.(2020)Kaplan, McCandlish, Henighan, Brown, Chess, Child, Gray, Radford, Wu, and Amodei]{kaplan2020scaling}
Jared Kaplan, Sam McCandlish, Tom Henighan, Tom~B Brown, Benjamin Chess, Rewon Child, Scott Gray, Alec Radford, Jeffrey Wu, and Dario Amodei.
\newblock Scaling laws for neural language models.
\newblock \emph{arXiv preprint arXiv:2001.08361}, 2020.

\bibitem[Li et~al.(2023)Li, Bubeck, Eldan, Del~Giorno, Gunasekar, and Lee]{li2023textbooks}
Yuanzhi Li, S{\'e}bastien Bubeck, Ronen Eldan, Allie Del~Giorno, Suriya Gunasekar, and Yin~Tat Lee.
\newblock Textbooks are all you need ii: phi-1.5 technical report.
\newblock \emph{arXiv preprint arXiv:2309.05463}, 2023.

\bibitem[Lin et~al.(2022{\natexlab{a}})Lin, Hilton, and Evans]{lin2021truthfulqa}
Stephanie Lin, Jacob Hilton, and Owain Evans.
\newblock {T}ruthful{QA}: Measuring how models mimic human falsehoods.
\newblock In \emph{Proceedings of the Annual Meeting of the Association for Computational Linguistics (ACL)}, 2022{\natexlab{a}}.

\bibitem[Lin et~al.(2022{\natexlab{b}})Lin, Mihaylov, Artetxe, Wang, Chen, Simig, Ott, Goyal, Bhosale, Du, Pasunuru, Shleifer, Koura, Chaudhary, O{'}Horo, Wang, Zettlemoyer, Kozareva, Diab, Stoyanov, and Li]{lin2021few}
Xi~Victoria Lin, Todor Mihaylov, Mikel Artetxe, Tianlu Wang, Shuohui Chen, Daniel Simig, Myle Ott, Naman Goyal, Shruti Bhosale, Jingfei Du, Ramakanth Pasunuru, Sam Shleifer, Punit~Singh Koura, Vishrav Chaudhary, Brian O{'}Horo, Jeff Wang, Luke Zettlemoyer, Zornitsa Kozareva, Mona Diab, Veselin Stoyanov, and Xian Li.
\newblock Few-shot learning with multilingual generative language models.
\newblock In \emph{Proceedings of the Conference on Empirical Methods in Natural Language Processing (EMNLP)}, pp.\  9019--9052, 2022{\natexlab{b}}.

\bibitem[Liu et~al.(2023)Liu, Qiao, Neiswanger, Wang, Tan, Tao, Li, Wang, Sun, Pangarkar, et~al.]{liu2023llm360}
Zhengzhong Liu, Aurick Qiao, Willie Neiswanger, Hongyi Wang, Bowen Tan, Tianhua Tao, Junbo Li, Yuqi Wang, Suqi Sun, Omkar Pangarkar, et~al.
\newblock Llm360: Towards fully transparent open-source llms.
\newblock \emph{arXiv preprint arXiv:2312.06550}, 2023.

\bibitem[Lu et~al.(2024)Lu, Bigoulaeva, Sachdeva, Tayyar~Madabushi, and Gurevych]{lu2023emergent}
Sheng Lu, Irina Bigoulaeva, Rachneet Sachdeva, Harish Tayyar~Madabushi, and Iryna Gurevych.
\newblock Are emergent abilities in large language models just in-context learning?
\newblock In \emph{Proceedings of the Annual Meeting of the Association for Computational Linguistics (ACL)}, 2024.

\bibitem[McKenzie et~al.(2023)McKenzie, Lyzhov, Pieler, Parrish, Mueller, Prabhu, McLean, Shen, Cavanagh, Gritsevskiy, Kauffman, Kirtland, Zhou, Zhang, Huang, Wurgaft, Weiss, Ross, Recchia, Liu, Liu, Tseng, Korbak, Kim, Bowman, and Perez]{mckenzie2023inverse}
Ian~R. McKenzie, Alexander Lyzhov, Michael~Martin Pieler, Alicia Parrish, Aaron Mueller, Ameya Prabhu, Euan McLean, Xudong Shen, Joe Cavanagh, Andrew~George Gritsevskiy, Derik Kauffman, Aaron~T. Kirtland, Zhengping Zhou, Yuhui Zhang, Sicong Huang, Daniel Wurgaft, Max Weiss, Alexis Ross, Gabriel Recchia, Alisa Liu, Jiacheng Liu, Tom Tseng, Tomasz Korbak, Najoung Kim, Samuel~R. Bowman, and Ethan Perez.
\newblock Inverse scaling: When bigger isn't better.
\newblock \emph{Transactions on Machine Learning Research (TMLR)}, 2023.
\newblock ISSN 2835-8856.

\bibitem[Michaud et~al.(2024)Michaud, Liu, Girit, and Tegmark]{michaud2024quantization}
Eric Michaud, Ziming Liu, Uzay Girit, and Max Tegmark.
\newblock The quantization model of neural scaling.
\newblock \emph{Proceedings of Advances in neural information processing systems (NeurIPS)}, 36, 2024.

\bibitem[Mihaylov et~al.(2018)Mihaylov, Clark, Khot, and Sabharwal]{mihaylov-etal-2018-suit}
Todor Mihaylov, Peter Clark, Tushar Khot, and Ashish Sabharwal.
\newblock Can a suit of armor conduct electricity? a new dataset for open book question answering.
\newblock In \emph{Proceedings of the Conference on Empirical Methods in Natural Language Processing (EMNLP)}, 2018.

\bibitem[Nakkiran et~al.(2021)Nakkiran, Kaplun, Bansal, Yang, Barak, and Sutskever]{nakkiran2021deep}
Preetum Nakkiran, Gal Kaplun, Yamini Bansal, Tristan Yang, Boaz Barak, and Ilya Sutskever.
\newblock Deep double descent: Where bigger models and more data hurt.
\newblock \emph{Journal of Statistical Mechanics: Theory and Experiment (JSTAT)}, 2021\penalty0 (12):\penalty0 124003, 2021.

\bibitem[Naveed et~al.(2023)Naveed, Khan, Qiu, Saqib, Anwar, Usman, Akhtar, Barnes, and Mian]{naveed2023comprehensive}
Humza Naveed, Asad~Ullah Khan, Shi Qiu, Muhammad Saqib, Saeed Anwar, Muhammad Usman, Naveed Akhtar, Nick Barnes, and Ajmal Mian.
\newblock A comprehensive overview of large language models.
\newblock \emph{arXiv preprint arXiv:2307.06435}, 2023.

\bibitem[Nijkamp et~al.(2023)Nijkamp, Pang, Hayashi, Tu, Wang, Zhou, Savarese, and Xiong]{nijkamp2022codegen}
Erik Nijkamp, Bo~Pang, Hiroaki Hayashi, Lifu Tu, Huan Wang, Yingbo Zhou, Silvio Savarese, and Caiming Xiong.
\newblock Codegen: An open large language model for code with multi-turn program synthesis.
\newblock In \emph{Proceedings of the International Conference on Learning Representations (ICLR)}, 2023.

\bibitem[Owen(2024)]{owen2024predictable}
David Owen.
\newblock How predictable is language model benchmark performance?
\newblock \emph{arXiv preprint arXiv:2401.04757}, 2024.

\bibitem[Pilehvar \& Camacho-Collados(2019)Pilehvar and Camacho-Collados]{pilehvar-camacho-collados-2019-wic}
Mohammad~Taher Pilehvar and Jose Camacho-Collados.
\newblock {W}i{C}: the word-in-context dataset for evaluating context-sensitive meaning representations.
\newblock In \emph{Proceedings of the Conference of the North {A}merican Chapter of the Association for Computational Linguistics: Human Language Technologies (NAACL)}, 2019.

\bibitem[Radford et~al.(2021)Radford, Kim, Hallacy, Ramesh, Goh, Agarwal, Sastry, Askell, Mishkin, Clark, et~al.]{radford2021learning}
Alec Radford, Jong~Wook Kim, Chris Hallacy, Aditya Ramesh, Gabriel Goh, Sandhini Agarwal, Girish Sastry, Amanda Askell, Pamela Mishkin, Jack Clark, et~al.
\newblock Learning transferable visual models from natural language supervision.
\newblock In \emph{Proceedings of the International Conference on Machine Learning (ICML)}, 2021.

\bibitem[Ruan et~al.(2024)Ruan, Maddison, and Hashimoto]{ruan2024observational}
Yangjun Ruan, Chris~J Maddison, and Tatsunori Hashimoto.
\newblock Observational scaling laws and the predictability of language model performance.
\newblock \emph{arXiv preprint arXiv:2405.10938}, 2024.

\bibitem[Schaeffer et~al.(2024{\natexlab{a}})Schaeffer, Miranda, and Koyejo]{schaeffer2024emergent}
Rylan Schaeffer, Brando Miranda, and Sanmi Koyejo.
\newblock Are emergent abilities of large language models a mirage?
\newblock \emph{Proceedings of Advances in neural information processing systems (NeurIPS)}, 36, 2024{\natexlab{a}}.

\bibitem[Schaeffer et~al.(2024{\natexlab{b}})Schaeffer, Schoelkopf, Miranda, Mukobi, Madan, Ibrahim, Bradley, Biderman, and Koyejo]{schaeffer2024has}
Rylan Schaeffer, Hailey Schoelkopf, Brando Miranda, Gabriel Mukobi, Varun Madan, Adam Ibrahim, Herbie Bradley, Stella Biderman, and Sanmi Koyejo.
\newblock Why has predicting downstream capabilities of frontier ai models with scale remained elusive?
\newblock \emph{arXiv preprint arXiv:2406.04391}, 2024{\natexlab{b}}.

\bibitem[Srivastava et~al.(2023)Srivastava, Rastogi, Rao, Shoeb, Abid, Fisch, Brown, Santoro, Gupta, Garriga-Alonso, Kluska, Lewkowycz, Agarwal, Power, Ray, Warstadt, Kocurek, Safaya, Tazarv, Xiang, Parrish, Nie, Hussain, Askell, Dsouza, Slone, Rahane, Iyer, Andreassen, Madotto, Santilli, Stuhlm{\"u}ller, Dai, La, Lampinen, Zou, Jiang, Chen, Vuong, Gupta, Gottardi, Norelli, Venkatesh, Gholamidavoodi, Tabassum, Menezes, Kirubarajan, Mullokandov, Sabharwal, Herrick, Efrat, Erdem, Karaka{\c{s}}, Roberts, Loe, Zoph, Bojanowski, {\"O}zyurt, Hedayatnia, Neyshabur, Inden, Stein, Ekmekci, Lin, Howald, Orinion, Diao, Dour, Stinson, Argueta, Ferri, Singh, Rathkopf, Meng, Baral, Wu, Callison-Burch, Waites, Voigt, Manning, Potts, Ramirez, Rivera, Siro, Raffel, Ashcraft, Garbacea, Sileo, Garrette, Hendrycks, Kilman, Roth, Freeman, Khashabi, Levy, Gonz{\'a}lez, Perszyk, Hernandez, Chen, Ippolito, Gilboa, Dohan, Drakard, Jurgens, Datta, Ganguli, Emelin, Kleyko, Yuret, Chen, Tam, Hupkes, Misra, Buzan, Mollo, Yang, Lee,
  Schrader, Shutova, Cubuk, Segal, Hagerman, Barnes, Donoway, Pavlick, Rodol{\`a}, Lam, Chu, Tang, Erdem, Chang, Chi, Dyer, Jerzak, Kim, Manyasi, Zheltonozhskii, Xia, Siar, Mart{\'\i}nez-Plumed, Happ{\'e}, Chollet, Rong, Mishra, Winata, de~Melo, Kruszewski, Parascandolo, Mariani, Wang, Jaimovitch-Lopez, Betz, Gur-Ari, Galijasevic, Kim, Rashkin, Hajishirzi, Mehta, Bogar, Shevlin, Schuetze, Yakura, Zhang, Wong, Ng, Noble, Jumelet, Geissinger, Kernion, Hilton, Lee, Fisac, Simon, Koppel, Zheng, Zou, Kocon, Thompson, Wingfield, Kaplan, Radom, Sohl-Dickstein, Phang, Wei, Yosinski, Novikova, Bosscher, Marsh, Kim, Taal, Engel, Alabi, Xu, Song, Tang, Waweru, Burden, Miller, Balis, Batchelder, Berant, Frohberg, Rozen, Hernandez-Orallo, Boudeman, Guerr, Jones, Tenenbaum, Rule, Chua, Kanclerz, Livescu, Krauth, Gopalakrishnan, Ignatyeva, Markert, Dhole, Gimpel, Omondi, Mathewson, Chiafullo, Shkaruta, Shridhar, McDonell, Richardson, Reynolds, Gao, Zhang, Dugan, Qin, Contreras-Ochando, Morency, Moschella, Lam, Noble,
  Schmidt, He, Oliveros-Col{\'o}n, Metz, Senel, Bosma, Sap, Hoeve, Farooqi, Faruqui, Mazeika, Baturan, Marelli, Maru, Ramirez-Quintana, Tolkiehn, Giulianelli, Lewis, Potthast, Leavitt, Hagen, Schubert, Baitemirova, Arnaud, McElrath, Yee, Cohen, Gu, Ivanitskiy, Starritt, Strube, Sw{\k{e}}drowski, Bevilacqua, Yasunaga, Kale, Cain, Xu, Suzgun, Walker, Tiwari, Bansal, Aminnaseri, Geva, Gheini, T, Peng, Chi, Lee, Krakover, Cameron, Roberts, Doiron, Martinez, Nangia, Deckers, Muennighoff, Keskar, Iyer, Constant, Fiedel, Wen, Zhang, Agha, Elbaghdadi, Levy, Evans, Casares, Doshi, Fung, Liang, Vicol, Alipoormolabashi, Liao, Liang, Chang, Eckersley, Htut, Hwang, Mi{\l}kowski, Patil, Pezeshkpour, Oli, Mei, Lyu, Chen, Banjade, Rudolph, Gabriel, Habacker, Risco, Milli{\`e}re, Garg, Barnes, Saurous, Arakawa, Raymaekers, Frank, Sikand, Novak, Sitelew, Bras, Liu, Jacobs, Zhang, Salakhutdinov, Chi, Lee, Stovall, Teehan, Yang, Singh, Mohammad, Anand, Dillavou, Shleifer, Wiseman, Gruetter, Bowman, Schoenholz, Han, Kwatra, Rous,
  Ghazarian, Ghosh, Casey, Bischoff, Gehrmann, Schuster, Sadeghi, Hamdan, Zhou, Srivastava, Shi, Singh, Asaadi, Gu, Pachchigar, Toshniwal, Upadhyay, Debnath, Shakeri, Thormeyer, Melzi, Reddy, Makini, Lee, Torene, Hatwar, Dehaene, Divic, Ermon, Biderman, Lin, Prasad, Piantadosi, Shieber, Misherghi, Kiritchenko, Mishra, Linzen, Schuster, Li, Yu, Ali, Hashimoto, Wu, Desbordes, Rothschild, Phan, Wang, Nkinyili, Schick, Kornev, Tunduny, Gerstenberg, Chang, Neeraj, Khot, Shultz, Shaham, Misra, Demberg, Nyamai, Raunak, Ramasesh, vinay~uday prabhu, Padmakumar, Srikumar, Fedus, Saunders, Zhang, Vossen, Ren, Tong, Zhao, Wu, Shen, Yaghoobzadeh, Lakretz, Song, Bahri, Choi, Yang, Hao, Chen, Belinkov, Hou, Hou, Bai, Seid, Zhao, Wang, Wang, Wang, and Wu]{srivastava2023beyond}
Aarohi Srivastava, Abhinav Rastogi, Abhishek Rao, Abu Awal~Md Shoeb, Abubakar Abid, Adam Fisch, Adam~R. Brown, Adam Santoro, Aditya Gupta, Adri{\`a} Garriga-Alonso, Agnieszka Kluska, Aitor Lewkowycz, Akshat Agarwal, Alethea Power, Alex Ray, Alex Warstadt, Alexander~W. Kocurek, Ali Safaya, Ali Tazarv, Alice Xiang, Alicia Parrish, Allen Nie, Aman Hussain, Amanda Askell, Amanda Dsouza, Ambrose Slone, Ameet Rahane, Anantharaman~S. Iyer, Anders~Johan Andreassen, Andrea Madotto, Andrea Santilli, Andreas Stuhlm{\"u}ller, Andrew~M. Dai, Andrew La, Andrew Lampinen, Andy Zou, Angela Jiang, Angelica Chen, Anh Vuong, Animesh Gupta, Anna Gottardi, Antonio Norelli, Anu Venkatesh, Arash Gholamidavoodi, Arfa Tabassum, Arul Menezes, Arun Kirubarajan, Asher Mullokandov, Ashish Sabharwal, Austin Herrick, Avia Efrat, Aykut Erdem, Ayla Karaka{\c{s}}, B.~Ryan Roberts, Bao~Sheng Loe, Barret Zoph, Bart{\l}omiej Bojanowski, Batuhan {\"O}zyurt, Behnam Hedayatnia, Behnam Neyshabur, Benjamin Inden, Benno Stein, Berk Ekmekci, Bill~Yuchen
  Lin, Blake Howald, Bryan Orinion, Cameron Diao, Cameron Dour, Catherine Stinson, Cedrick Argueta, Cesar Ferri, Chandan Singh, Charles Rathkopf, Chenlin Meng, Chitta Baral, Chiyu Wu, Chris Callison-Burch, Christopher Waites, Christian Voigt, Christopher~D Manning, Christopher Potts, Cindy Ramirez, Clara~E. Rivera, Clemencia Siro, Colin Raffel, Courtney Ashcraft, Cristina Garbacea, Damien Sileo, Dan Garrette, Dan Hendrycks, Dan Kilman, Dan Roth, C.~Daniel Freeman, Daniel Khashabi, Daniel Levy, Daniel~Mosegu{\'\i} Gonz{\'a}lez, Danielle Perszyk, Danny Hernandez, Danqi Chen, Daphne Ippolito, Dar Gilboa, David Dohan, David Drakard, David Jurgens, Debajyoti Datta, Deep Ganguli, Denis Emelin, Denis Kleyko, Deniz Yuret, Derek Chen, Derek Tam, Dieuwke Hupkes, Diganta Misra, Dilyar Buzan, Dimitri~Coelho Mollo, Diyi Yang, Dong-Ho Lee, Dylan Schrader, Ekaterina Shutova, Ekin~Dogus Cubuk, Elad Segal, Eleanor Hagerman, Elizabeth Barnes, Elizabeth Donoway, Ellie Pavlick, Emanuele Rodol{\`a}, Emma Lam, Eric Chu, Eric Tang,
  Erkut Erdem, Ernie Chang, Ethan~A Chi, Ethan Dyer, Ethan Jerzak, Ethan Kim, Eunice~Engefu Manyasi, Evgenii Zheltonozhskii, Fanyue Xia, Fatemeh Siar, Fernando Mart{\'\i}nez-Plumed, Francesca Happ{\'e}, Francois Chollet, Frieda Rong, Gaurav Mishra, Genta~Indra Winata, Gerard de~Melo, Germ{\'a}n Kruszewski, Giambattista Parascandolo, Giorgio Mariani, Gloria~Xinyue Wang, Gonzalo Jaimovitch-Lopez, Gregor Betz, Guy Gur-Ari, Hana Galijasevic, Hannah Kim, Hannah Rashkin, Hannaneh Hajishirzi, Harsh Mehta, Hayden Bogar, Henry Francis~Anthony Shevlin, Hinrich Schuetze, Hiromu Yakura, Hongming Zhang, Hugh~Mee Wong, Ian Ng, Isaac Noble, Jaap Jumelet, Jack Geissinger, Jackson Kernion, Jacob Hilton, Jaehoon Lee, Jaime~Fern{\'a}ndez Fisac, James~B Simon, James Koppel, James Zheng, James Zou, Jan Kocon, Jana Thompson, Janelle Wingfield, Jared Kaplan, Jarema Radom, Jascha Sohl-Dickstein, Jason Phang, Jason Wei, Jason Yosinski, Jekaterina Novikova, Jelle Bosscher, Jennifer Marsh, Jeremy Kim, Jeroen Taal, Jesse Engel, Jesujoba
  Alabi, Jiacheng Xu, Jiaming Song, Jillian Tang, Joan Waweru, John Burden, John Miller, John~U. Balis, Jonathan Batchelder, Jonathan Berant, J{\"o}rg Frohberg, Jos Rozen, Jose Hernandez-Orallo, Joseph Boudeman, Joseph Guerr, Joseph Jones, Joshua~B. Tenenbaum, Joshua~S. Rule, Joyce Chua, Kamil Kanclerz, Karen Livescu, Karl Krauth, Karthik Gopalakrishnan, Katerina Ignatyeva, Katja Markert, Kaustubh Dhole, Kevin Gimpel, Kevin Omondi, Kory~Wallace Mathewson, Kristen Chiafullo, Ksenia Shkaruta, Kumar Shridhar, Kyle McDonell, Kyle Richardson, Laria Reynolds, Leo Gao, Li~Zhang, Liam Dugan, Lianhui Qin, Lidia Contreras-Ochando, Louis-Philippe Morency, Luca Moschella, Lucas Lam, Lucy Noble, Ludwig Schmidt, Luheng He, Luis Oliveros-Col{\'o}n, Luke Metz, L{\"u}tfi~Kerem Senel, Maarten Bosma, Maarten Sap, Maartje~Ter Hoeve, Maheen Farooqi, Manaal Faruqui, Mantas Mazeika, Marco Baturan, Marco Marelli, Marco Maru, Maria~Jose Ramirez-Quintana, Marie Tolkiehn, Mario Giulianelli, Martha Lewis, Martin Potthast, Matthew~L
  Leavitt, Matthias Hagen, M{\'a}ty{\'a}s Schubert, Medina~Orduna Baitemirova, Melody Arnaud, Melvin McElrath, Michael~Andrew Yee, Michael Cohen, Michael Gu, Michael Ivanitskiy, Michael Starritt, Michael Strube, Micha{\l} Sw{\k{e}}drowski, Michele Bevilacqua, Michihiro Yasunaga, Mihir Kale, Mike Cain, Mimee Xu, Mirac Suzgun, Mitch Walker, Mo~Tiwari, Mohit Bansal, Moin Aminnaseri, Mor Geva, Mozhdeh Gheini, Mukund~Varma T, Nanyun Peng, Nathan~Andrew Chi, Nayeon Lee, Neta Gur-Ari Krakover, Nicholas Cameron, Nicholas Roberts, Nick Doiron, Nicole Martinez, Nikita Nangia, Niklas Deckers, Niklas Muennighoff, Nitish~Shirish Keskar, Niveditha~S. Iyer, Noah Constant, Noah Fiedel, Nuan Wen, Oliver Zhang, Omar Agha, Omar Elbaghdadi, Omer Levy, Owain Evans, Pablo Antonio~Moreno Casares, Parth Doshi, Pascale Fung, Paul~Pu Liang, Paul Vicol, Pegah Alipoormolabashi, Peiyuan Liao, Percy Liang, Peter~W Chang, Peter Eckersley, Phu~Mon Htut, Pinyu Hwang, Piotr Mi{\l}kowski, Piyush Patil, Pouya Pezeshkpour, Priti Oli, Qiaozhu
  Mei, Qing Lyu, Qinlang Chen, Rabin Banjade, Rachel~Etta Rudolph, Raefer Gabriel, Rahel Habacker, Ramon Risco, Rapha{\"e}l Milli{\`e}re, Rhythm Garg, Richard Barnes, Rif~A. Saurous, Riku Arakawa, Robbe Raymaekers, Robert Frank, Rohan Sikand, Roman Novak, Roman Sitelew, Ronan~Le Bras, Rosanne Liu, Rowan Jacobs, Rui Zhang, Russ Salakhutdinov, Ryan~Andrew Chi, Seungjae~Ryan Lee, Ryan Stovall, Ryan Teehan, Rylan Yang, Sahib Singh, Saif~M. Mohammad, Sajant Anand, Sam Dillavou, Sam Shleifer, Sam Wiseman, Samuel Gruetter, Samuel~R. Bowman, Samuel~Stern Schoenholz, Sanghyun Han, Sanjeev Kwatra, Sarah~A. Rous, Sarik Ghazarian, Sayan Ghosh, Sean Casey, Sebastian Bischoff, Sebastian Gehrmann, Sebastian Schuster, Sepideh Sadeghi, Shadi Hamdan, Sharon Zhou, Shashank Srivastava, Sherry Shi, Shikhar Singh, Shima Asaadi, Shixiang~Shane Gu, Shubh Pachchigar, Shubham Toshniwal, Shyam Upadhyay, Shyamolima~Shammie Debnath, Siamak Shakeri, Simon Thormeyer, Simone Melzi, Siva Reddy, Sneha~Priscilla Makini, Soo-Hwan Lee, Spencer
  Torene, Sriharsha Hatwar, Stanislas Dehaene, Stefan Divic, Stefano Ermon, Stella Biderman, Stephanie Lin, Stephen Prasad, Steven Piantadosi, Stuart Shieber, Summer Misherghi, Svetlana Kiritchenko, Swaroop Mishra, Tal Linzen, Tal Schuster, Tao Li, Tao Yu, Tariq Ali, Tatsunori Hashimoto, Te-Lin Wu, Th{\'e}o Desbordes, Theodore Rothschild, Thomas Phan, Tianle Wang, Tiberius Nkinyili, Timo Schick, Timofei Kornev, Titus Tunduny, Tobias Gerstenberg, Trenton Chang, Trishala Neeraj, Tushar Khot, Tyler Shultz, Uri Shaham, Vedant Misra, Vera Demberg, Victoria Nyamai, Vikas Raunak, Vinay~Venkatesh Ramasesh, vinay~uday prabhu, Vishakh Padmakumar, Vivek Srikumar, William Fedus, William Saunders, William Zhang, Wout Vossen, Xiang Ren, Xiaoyu Tong, Xinran Zhao, Xinyi Wu, Xudong Shen, Yadollah Yaghoobzadeh, Yair Lakretz, Yangqiu Song, Yasaman Bahri, Yejin Choi, Yichi Yang, Yiding Hao, Yifu Chen, Yonatan Belinkov, Yu~Hou, Yufang Hou, Yuntao Bai, Zachary Seid, Zhuoye Zhao, Zijian Wang, Zijie~J. Wang, Zirui Wang, and Ziyi Wu.
\newblock Beyond the imitation game: Quantifying and extrapolating the capabilities of language models.
\newblock \emph{Transactions on Machine Learning Research (TMLR)}, 2023.
\newblock ISSN 2835-8856.

\bibitem[Stability-AI(2023)]{stablelm}
Stability-AI.
\newblock Stablelm: Stability ai language models, April 2023.
\newblock URL \url{https://github.com/Stability-AI/StableLM}.

\bibitem[Team et~al.(2023)Team, Anil, Borgeaud, Wu, Alayrac, Yu, Soricut, Schalkwyk, Dai, Hauth, et~al.]{team2023gemini}
Gemini Team, Rohan Anil, Sebastian Borgeaud, Yonghui Wu, Jean-Baptiste Alayrac, Jiahui Yu, Radu Soricut, Johan Schalkwyk, Andrew~M Dai, Anja Hauth, et~al.
\newblock Gemini: a family of highly capable multimodal models.
\newblock \emph{arXiv preprint arXiv:2312.11805}, 2023.

\bibitem[Team et~al.(2024)Team, Mesnard, Hardin, Dadashi, Bhupatiraju, Pathak, Sifre, Rivi{\`e}re, Kale, Love, et~al.]{team2024gemma}
Gemma Team, Thomas Mesnard, Cassidy Hardin, Robert Dadashi, Surya Bhupatiraju, Shreya Pathak, Laurent Sifre, Morgane Rivi{\`e}re, Mihir~Sanjay Kale, Juliette Love, et~al.
\newblock Gemma: Open models based on gemini research and technology.
\newblock \emph{arXiv preprint arXiv:2403.08295}, 2024.

\bibitem[Team(2023)]{MosaicML2023Introducing}
MosaicML~NLP Team.
\newblock Introducing mpt-7b: A new standard for open-source, commercially usable llms, May 2023.
\newblock URL \url{www.mosaicml.com/blog/mpt-7b}.

\bibitem[Touvron et~al.(2023{\natexlab{a}})Touvron, Lavril, Izacard, Martinet, Lachaux, Lacroix, Rozi{\`e}re, Goyal, Hambro, Azhar, et~al.]{touvron2023llama}
Hugo Touvron, Thibaut Lavril, Gautier Izacard, Xavier Martinet, Marie-Anne Lachaux, Timoth{\'e}e Lacroix, Baptiste Rozi{\`e}re, Naman Goyal, Eric Hambro, Faisal Azhar, et~al.
\newblock Llama: Open and efficient foundation language models.
\newblock \emph{arXiv preprint arXiv:2302.13971}, 2023{\natexlab{a}}.

\bibitem[Touvron et~al.(2023{\natexlab{b}})Touvron, Martin, Stone, Albert, Almahairi, Babaei, Bashlykov, Batra, Bhargava, Bhosale, et~al.]{touvron2023llama2}
Hugo Touvron, Louis Martin, Kevin Stone, Peter Albert, Amjad Almahairi, Yasmine Babaei, Nikolay Bashlykov, Soumya Batra, Prajjwal Bhargava, Shruti Bhosale, et~al.
\newblock Llama 2: Open foundation and fine-tuned chat models.
\newblock \emph{arXiv preprint arXiv:2307.09288}, 2023{\natexlab{b}}.

\bibitem[Wei et~al.(2022)Wei, Tay, Bommasani, Raffel, Zoph, Borgeaud, Yogatama, Bosma, Zhou, Metzler, Chi, Hashimoto, Vinyals, Liang, Dean, and Fedus]{wei2022emergent}
Jason Wei, Yi~Tay, Rishi Bommasani, Colin Raffel, Barret Zoph, Sebastian Borgeaud, Dani Yogatama, Maarten Bosma, Denny Zhou, Donald Metzler, Ed~H. Chi, Tatsunori Hashimoto, Oriol Vinyals, Percy Liang, Jeff Dean, and William Fedus.
\newblock Emergent abilities of large language models.
\newblock \emph{Transactions on Machine Learning Research (TMLR)}, 2022.
\newblock ISSN 2835-8856.

\bibitem[Wei et~al.(2023)Wei, Kim, Tay, and Le]{wei-etal-2023-inverse}
Jason Wei, Najoung Kim, Yi~Tay, and Quoc Le.
\newblock Inverse scaling can become {U}-shaped.
\newblock In \emph{Proceedings of the Conference on Empirical Methods in Natural Language Processing (EMNLP)}, 2023.

\bibitem[Workshop et~al.(2022)Workshop, Scao, Fan, Akiki, Pavlick, Ili{\'c}, Hesslow, Castagn{\'e}, Luccioni, Yvon, et~al.]{workshop2022bloom}
BigScience Workshop, Teven~Le Scao, Angela Fan, Christopher Akiki, Ellie Pavlick, Suzana Ili{\'c}, Daniel Hesslow, Roman Castagn{\'e}, Alexandra~Sasha Luccioni, Fran{\c{c}}ois Yvon, et~al.
\newblock Bloom: A 176b-parameter open-access multilingual language model.
\newblock \emph{arXiv preprint arXiv:2211.05100}, 2022.

\bibitem[Ye et~al.(2023)Ye, Fu, Ren, and Jia]{ye-etal-2023-predictable}
Qinyuan Ye, Harvey Fu, Xiang Ren, and Robin Jia.
\newblock How predictable are large language model capabilities? a case study on {BIG}-bench.
\newblock In \emph{Findings of Empirical Methods in Natural Language Processing (EMNLP)}, 2023.

\bibitem[Young et~al.(2024)Young, Chen, Li, Huang, Zhang, Zhang, Li, Zhu, Chen, Chang, et~al.]{young2024yi}
Alex Young, Bei Chen, Chao Li, Chengen Huang, Ge~Zhang, Guanwei Zhang, Heng Li, Jiangcheng Zhu, Jianqun Chen, Jing Chang, et~al.
\newblock Yi: Open foundation models by 01. ai.
\newblock \emph{arXiv preprint arXiv:2403.04652}, 2024.

\bibitem[Zellers et~al.(2019)Zellers, Holtzman, Bisk, Farhadi, and Choi]{zellers2019hellaswag}
Rowan Zellers, Ari Holtzman, Yonatan Bisk, Ali Farhadi, and Yejin Choi.
\newblock Hellaswag: Can a machine really finish your sentence?
\newblock In \emph{Proceedings of the Annual Meeting of the Association for Computational Linguistics (ACL)}, 2019.

\bibitem[Zhang et~al.(2022)Zhang, Roller, Goyal, Artetxe, Chen, Chen, Dewan, Diab, Li, Lin, et~al.]{zhang2022opt}
Susan Zhang, Stephen Roller, Naman Goyal, Mikel Artetxe, Moya Chen, Shuohui Chen, Christopher Dewan, Mona Diab, Xian Li, Xi~Victoria Lin, et~al.
\newblock Opt: Open pre-trained transformer language models.
\newblock \emph{arXiv preprint arXiv:2205.01068}, 2022.

\end{thebibliography}
\bibliographystyle{iclr2025_conference}

\clearpage
\appendix
\renewcommand{\thetable}{A\arabic{table}}
\renewcommand{\thefigure}{A\arabic{figure}}
\renewcommand{\thealgorithm}{A\arabic{algorithm}}
\addtocontents{toc}{\protect\setcounter{tocdepth}{2}}
\renewcommand*\contentsname{Supplementary Material}
\tableofcontents
\clearpage
\section{Implementation Details}
\label{sup: implementation details}

\subsection{LLM Evaluation}
\label{sec2-sub: exp setup}
We evaluate all datasets in this paper on the LM Evaluation Harness~\citep{eval-harness} platform. We adopt 56 models, including Gemma~\citep{team2024gemma}, Llama~\citep{touvron2023llama}, Llama-2~\citep{touvron2023llama2}, RedPajama-INCITE~\citep{together2023redpajama}, Yi~\citep{young2024yi}, StableLM~\citep{stablelm}, MPT~\citep{MosaicML2023Introducing}, Falcon~\citep{almazrouei2023falcon}, Pythia~\citep{biderman2023pythia}, AMBER~\citep{liu2023llm360}, Qwen~\citep{bai2023qwen}, Qwen-1.5~\citep{bai2023qwen}, BLOOM~\citep{workshop2022bloom}, DeepSeekMoE~\citep{dai2024deepseekmoe}, OPT~\citep{zhang2022opt}, GPT-Neo~\citep{gpt-neo}, Codegen~\citep{nijkamp2022codegen}, XGLM~\citep{lin2021few}, and OpenLLaMA~\citep{openlm2023openllama} families under FP16 precision. The evaluation time of each task varies from several hours to several days on 2 NVIDIA RTX A6000, depending on the question numbers and formats. We obtain each model's log compute through the released data by \citet{ruan2024observational}. We use $T=1.5, 1.8$, and $2.3$ as the emergence threshold for the MMLU, arithmetic, and Persian-QA dataset, respectively. We calculate question difficulty level $q_d$ using models smaller than these thresholds. We adopt 5-shot inference on the MMLU benchmark, 1-shot inference on the ARC and HellaSwag dataset, and 2-shot inference on Persian-QA, arithmetic, Hindu knowledge, conceptual combinations, analogical similarity, and abstract narrative understanding datasets (ARC, HellaSwag, and abstract narrative understanding datasets are used in App.~\ref{sup: ppp on nonemergent}).

\subsection{Slice-and-Sandwich}
In the main paper, we examine \textit{Slice-and-Sandwich} on MMLU, arithmetic, and Persian-QA datasets with group number $G=3$. Models smaller than $T=1.5, 1.8$, and $2.3$ in the MMLU, arithmetic, and Persian-QA datasets are the training set; other larger models are the testing set. We adopt polynomial regression to fit easy and hard question groups. Specifically, we adopt the polynomial order=5 and 2 for the easy and hard question groups, respectively.
\clearpage
\section{More Discussions on Brier Score}
\label{sup: brier score comparison}
In the main paper, we use the model's predicted probability of the correct class \textit{conditional} on all classes to calculate the TC Brier Score (see Eq.~\ref{eq: conditional prob}). This section discusses the effect of such conditionalization. This section refers to the \textit{un-conditionalized TC Brier Score} as the one without re-distributing output probabilities to all classes.

\begin{figure*}[tb]
  \centering
  \begin{subfigure}{0.32\linewidth}
    \includegraphics[width=\textwidth]{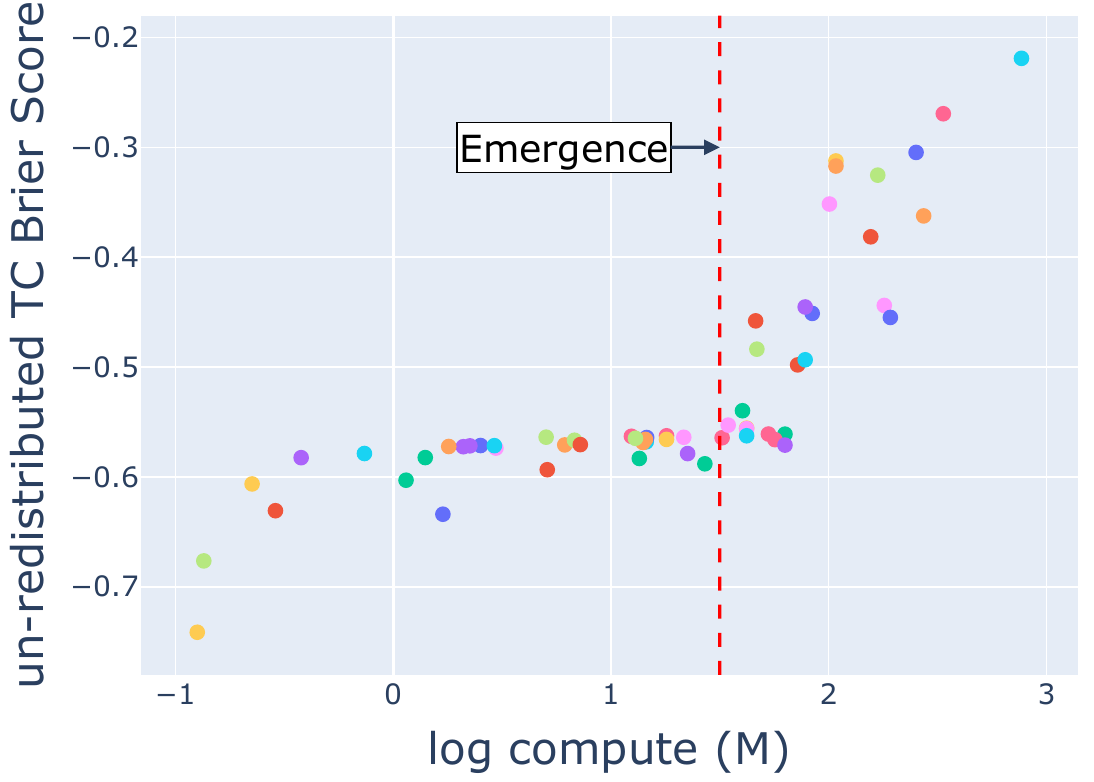}
    \caption{MMLU.}
    \label{subfig: mmlu-undist-brier}
  \end{subfigure}
  \hfill
  \begin{subfigure}{0.32\linewidth}
    \includegraphics[width=\textwidth]{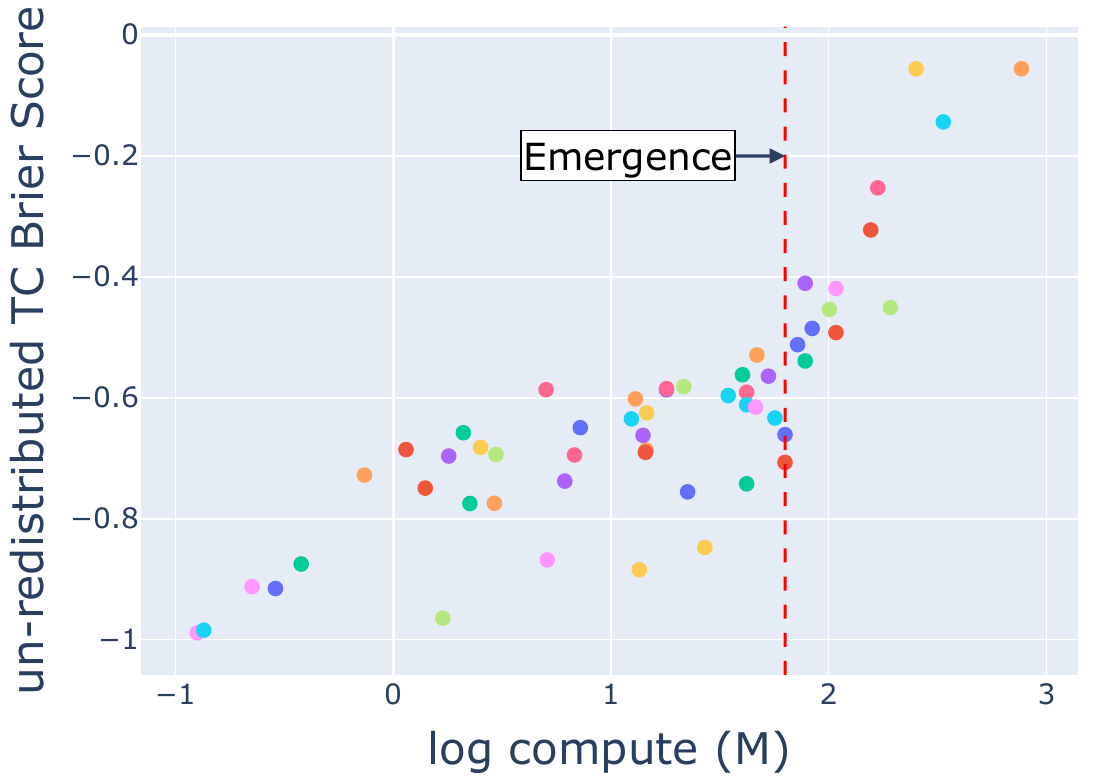}
    \caption{Arithmetic.}
    \label{subfig: arithmetic-undist-brier}
  \end{subfigure}
  \hfill
  \begin{subfigure}{0.32\linewidth}
    \includegraphics[width=\textwidth]{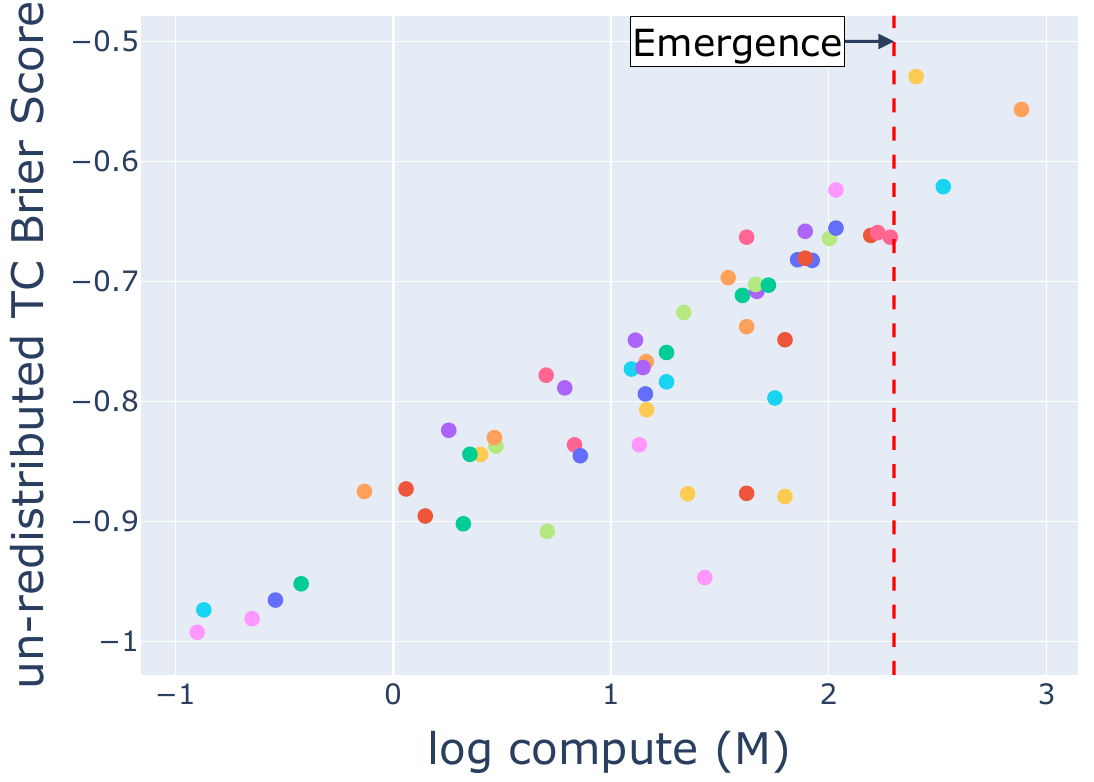}
    \caption{Persian-QA.}
    \label{subfig: persian-qa-undist-brier}
  \end{subfigure}
  \caption{The un-conditionalized TC Brier Score vs. log compute (M) on the MMLU, arithmetic, and Persian-QA datasets.}
  \label{fig: undist-brier}
\end{figure*}
\begin{figure*}[tb]
  \centering
  \begin{subfigure}{0.32\linewidth}
    \includegraphics[width=\textwidth]{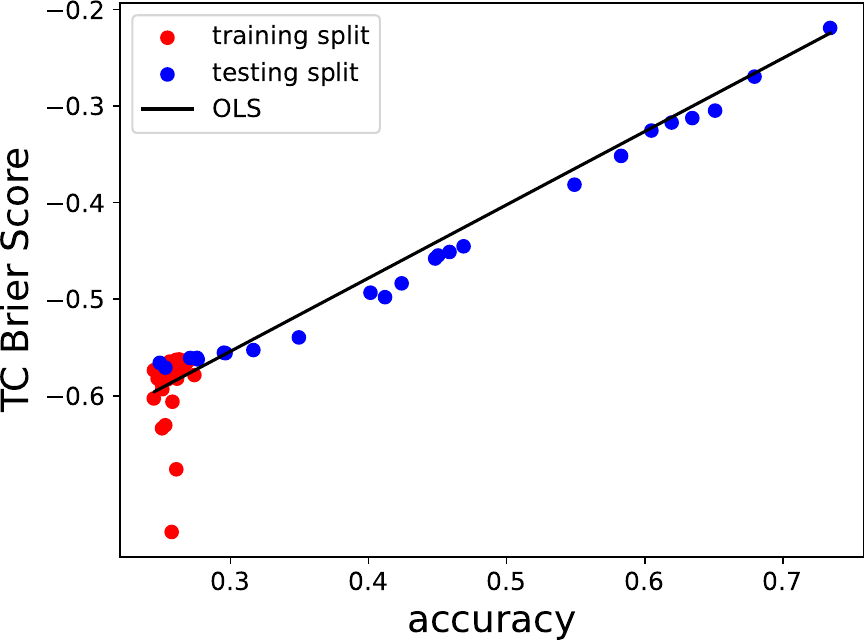}
    \caption{MMLU.}
    \label{subfig: mmlu-rel-undist}
  \end{subfigure}
  \hfill
  \begin{subfigure}{0.32\linewidth}
    \includegraphics[width=\textwidth]{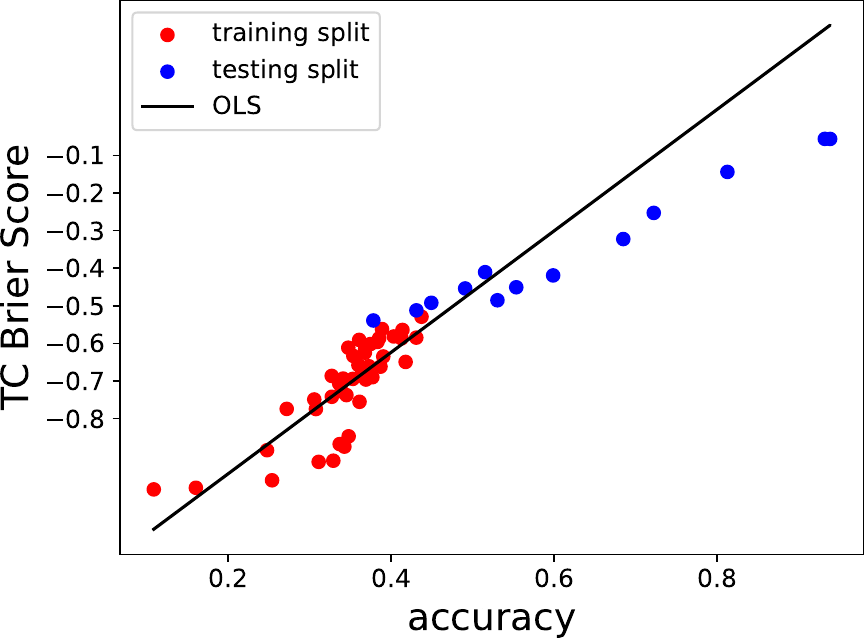}
    \caption{Arithmetic.}
    \label{subfig: arithmetic-rel-undist}
  \end{subfigure}
  \hfill
  \begin{subfigure}{0.32\linewidth}
    \includegraphics[width=\textwidth]{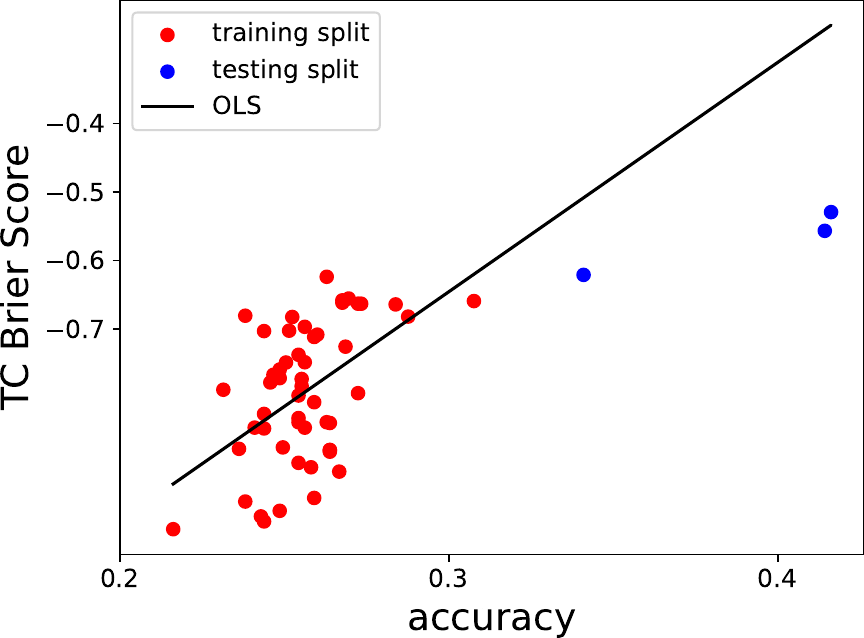}
    \caption{Persian-QA.}
    \label{subfig: persian-qa-rel-undist}
  \end{subfigure}
  \caption{The relation between accuracy and un-conditionalized TC Brier Score on the MMLU, arithmetic, and Persian-QA datasets.}
  \label{fig: undist-brier-acc-relation}
\end{figure*}
\begin{table*}[tb]
\begin{center}
\begin{small}
\begin{sc}
\addtolength{\tabcolsep}{-1pt}
\caption{Comparison of correlation coefficients between accuracy and TC Brier Score with and without conditionality on the MMLU, arithmetic, and Persian-QA datasets. ``P.'', ``S.'', and ``K.'' stands for Pearson, Spearman, and Kendall, respectively. The TC Brier Score with conditionality, i.e., the one we adopt in the main paper, has a consistently stronger correlation with accuracy.}
\label{tab: acc-brier-correlation}
\begin{tabular}{c|ccc|ccc|ccc}
\toprule
\midrule
\multicolumn{1}{c|}{} & \multicolumn{3}{c|}{MMLU} & \multicolumn{3}{c|}{arithmetic} & \multicolumn{3}{c}{Persian-QA}\\
\midrule
Correlation coefficient & P.  & S.  & K. & P.  & S.  & K. & P.  & S. & K. \\
\midrule
un-conditionalized TC Brier Score & 0.96  & 0.87  & 0.73  & 0.93 & 0.93 & 0.79 & 0.60 & 0.52 & 0.37 \\

TC Brier Score & \textbf{0.99}  & \textbf{0.91} 
 & \textbf{0.79} & \textbf{1.00}  & \textbf{0.97} & \textbf{0.88} & \textbf{0.88} & \textbf{0.68} & \textbf{0.52} \\ 
\midrule
\bottomrule
\end{tabular}
\end{sc}
\end{small}
\end{center}

\end{table*}
Fig.~\ref{fig: undist-brier} shows the relationship between un-conditionalized TC Brier Score and log compute $M$ on all three datasets. For the MMLU dataset, model performance still exhibits flat scaling before the emergence threshold and sharp improvement past the emergence threshold. For the arithmetic and Persian-QA datasets, the scaling trend does not show a sharp increase and is easier to forecast performance under the un-conditionalized TC Brier Score past the emergence threshold. This is consistent with the finding of  \citep{schaeffer2024has} that the un-conditionalized measure is more correlated with the training compute than conditionalized ones. However, Fig.~\ref{fig: undist-brier-acc-relation} shows that the un-conditionalized TC Brier Score is not as closely related to accuracy as the normal TC Brier Score for the arithmetic and especially the Persian-QA dataset. Table~\ref{tab: acc-brier-correlation} corroborates this assertion by showing the correlation coefficient between accuracy and the normal/un-conditionalized TC Brier Score.

\begin{figure*}[tb]
  \centering
  \begin{subfigure}{0.32\linewidth}
    \includegraphics[width=\textwidth]{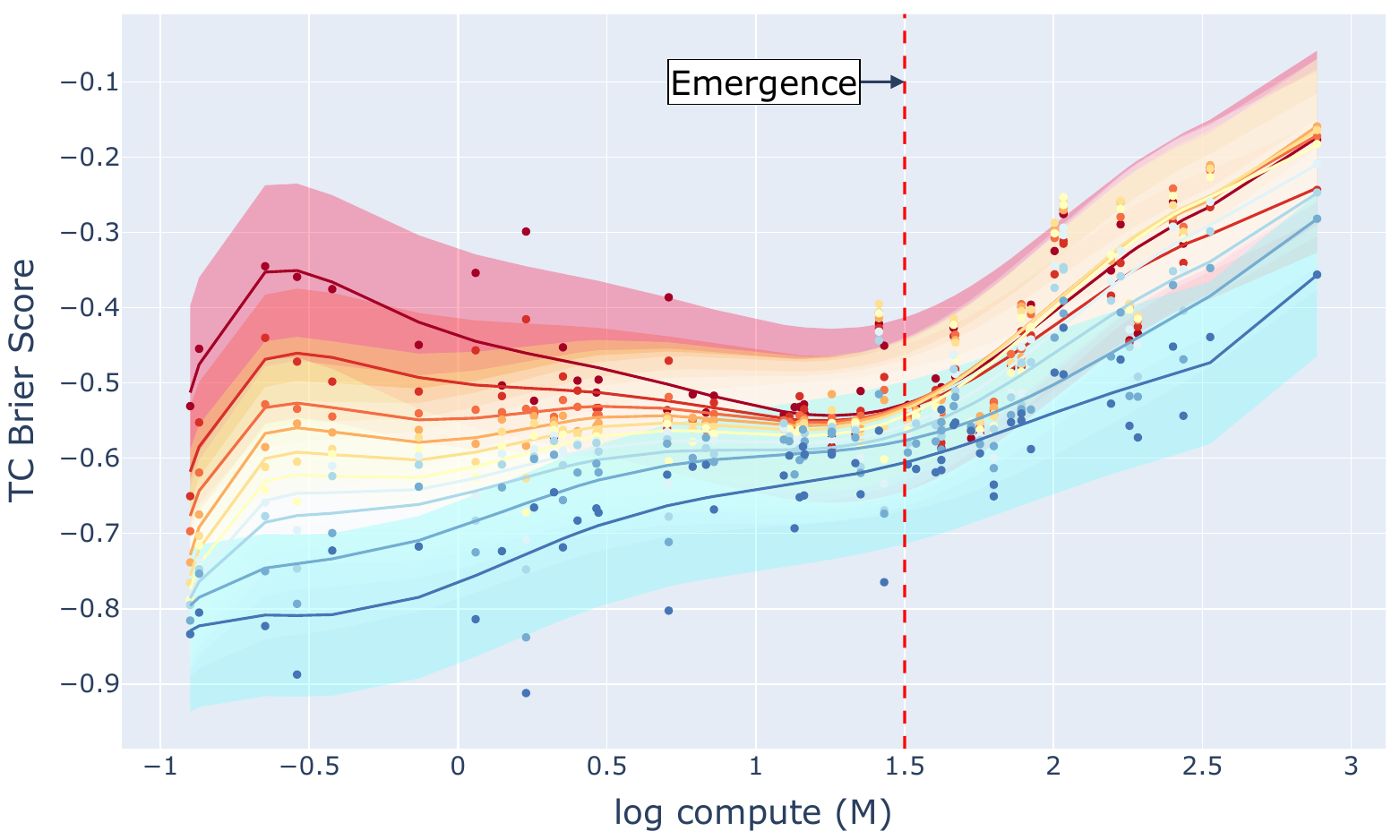}
    \caption{MMLU.}
    \label{subfig: mmlu undist}
  \end{subfigure}
  \hfill
  \begin{subfigure}{0.32\linewidth}
    \includegraphics[width=\textwidth]{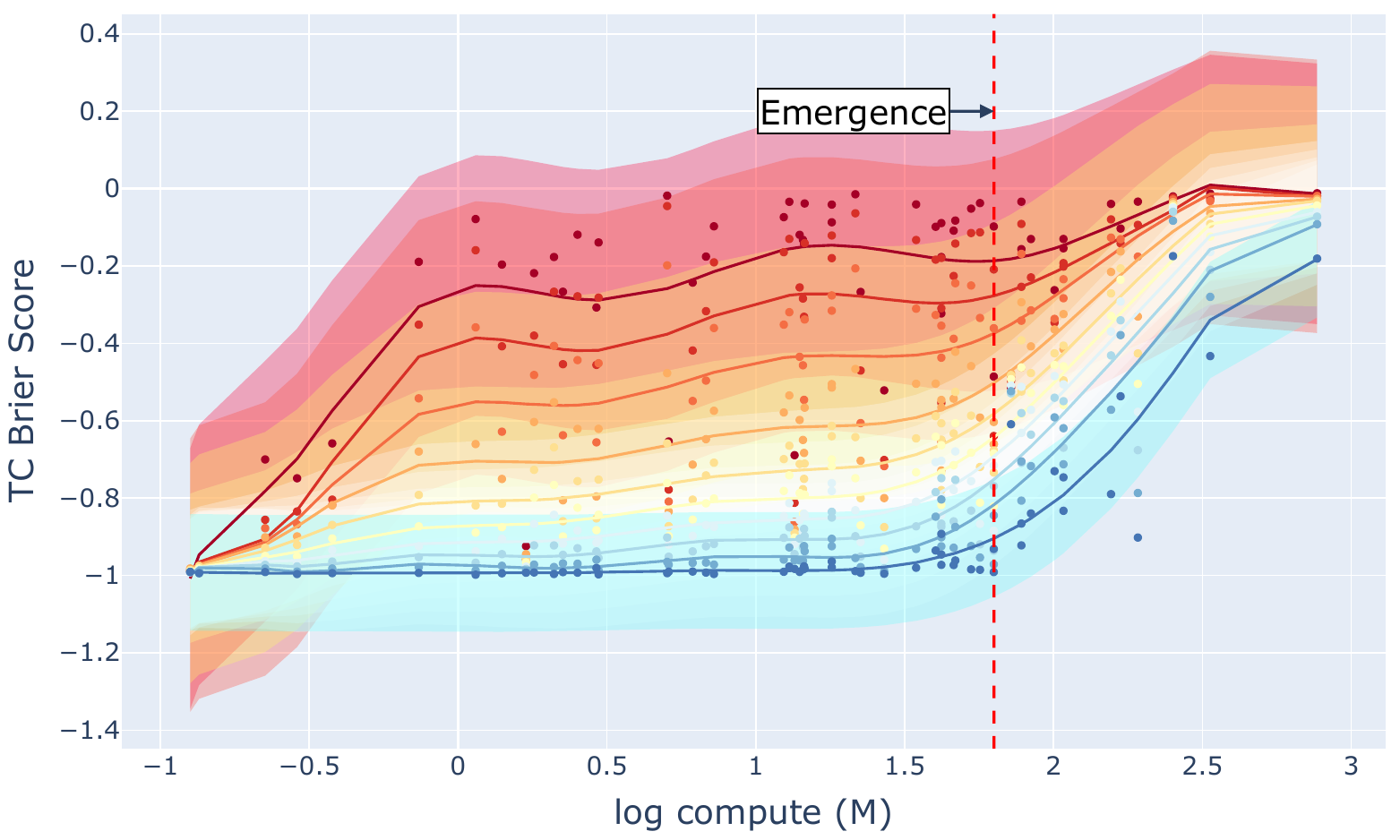}
    \caption{Arithmetic.}
    \label{subfig: arithmetic undist}
  \end{subfigure}
  \hfill
  \begin{subfigure}{0.32\linewidth}
    \includegraphics[width=\textwidth]{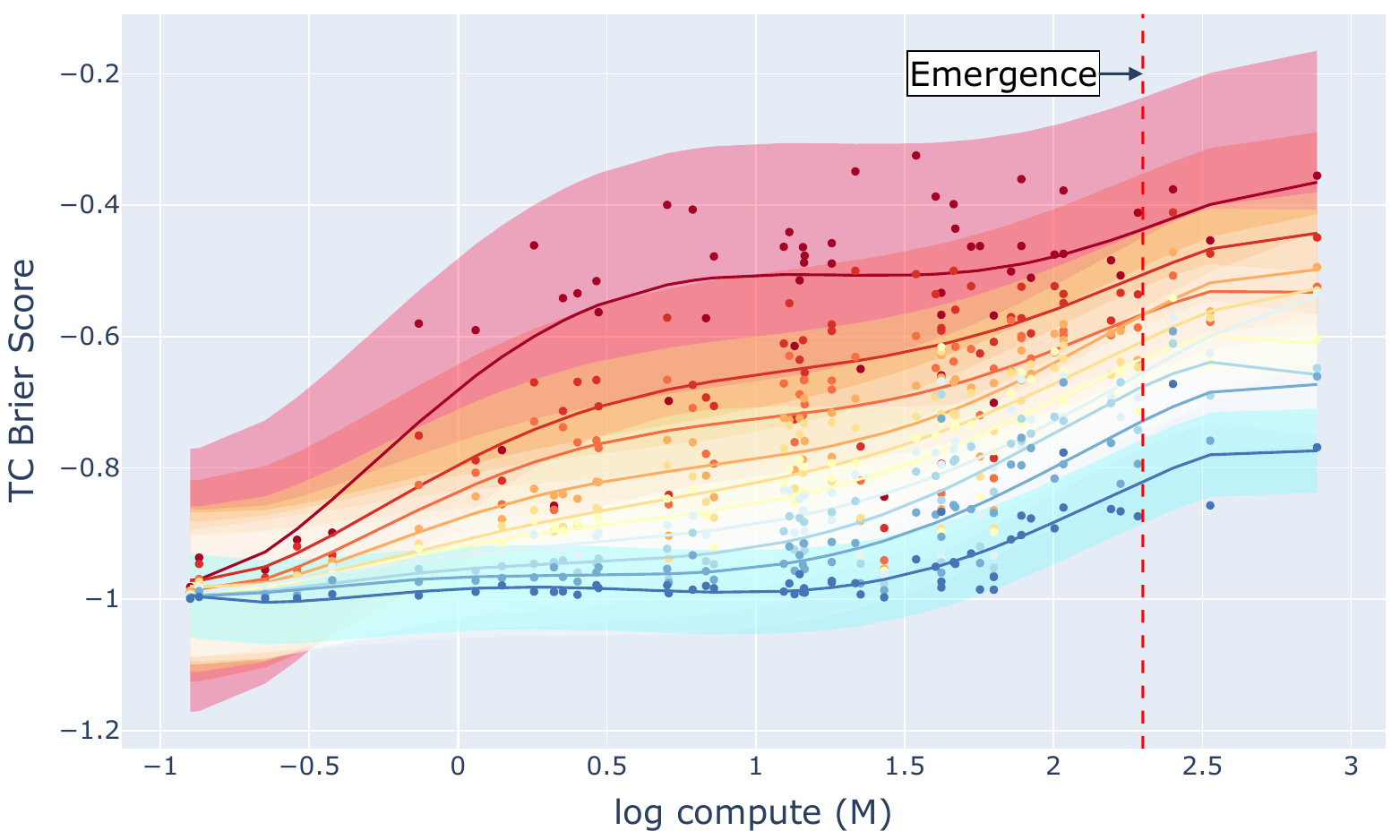}
    \caption{Persian-QA.}
    \label{subfig: persian-qa-undist}
  \end{subfigure}
  \caption{The U-shaped and inverted-U scaling with questions grouped and performances measured by the un-conditionalized TC Brier Score on the MMLU, arithmetic, and Persian-QA daatasets.}
  \label{fig: undist ppp}
\end{figure*}
Fig.~\ref{fig: undist ppp} shows the scaling trend by difficulty level for the un-conditionalized TC Brier Score. For the arithmetic and Persian-QA datasets, we no longer see inverse scaling on any intervals of log compute $M$. In fact, for both the arithmetic and the Persian-QA datasets, performance hovers around $-1$ on the hardest group, corresponding to the near-zero predicted probability of the correct class. For hard questions, the model's predicted probability on all classes is close to zero. Therefore, without conditionalizing on all classes, we cannot differentiate between an initial random guess and the distracted phase at larger model log computes where the model places a higher probability on an available incorrect class relative to the correct class, which yields the U-shaped scaling of normal TC Brier Score as discussed in the main paper.
\clearpage
\section{Scaling Trend by Question Difficulty Level for Other Emergent Tasks}
\label{sup: more ppp}
\begin{figure*}[ht]
  \centering
  \begin{subfigure}{0.48\linewidth}
    \includegraphics[width=\textwidth]{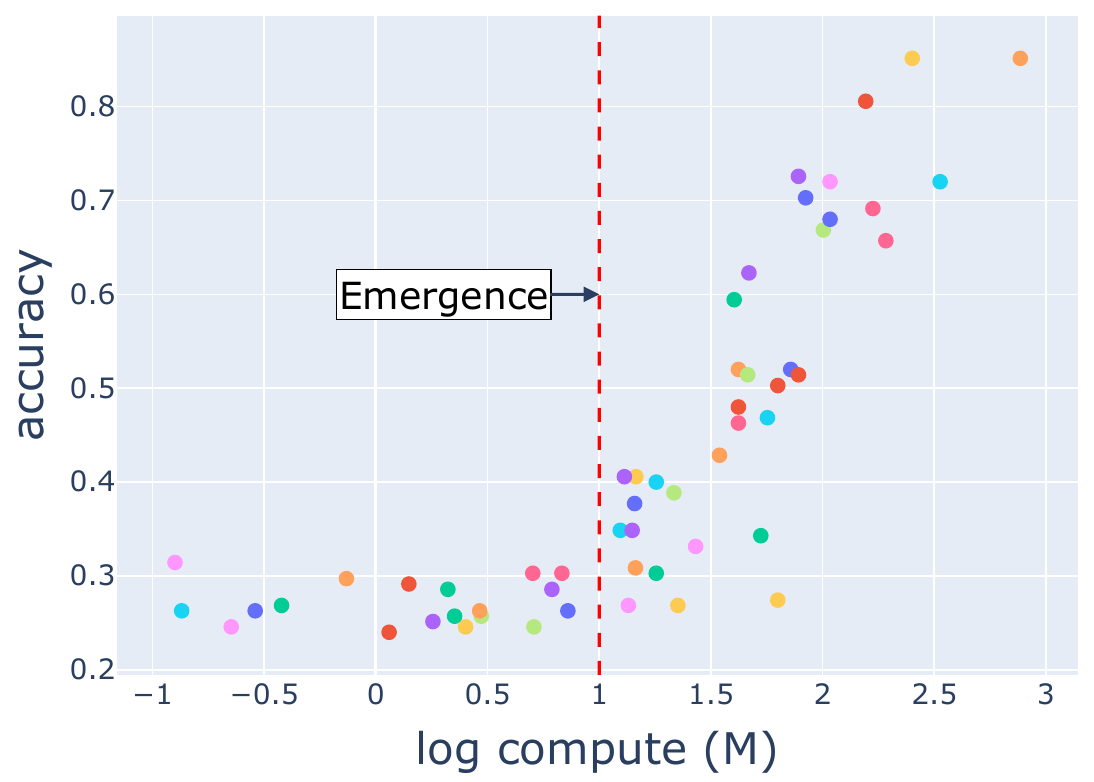}
    \caption{Accuracy vs. log compute (M).}
    \label{subfig: hindu_knowledge accuracy}
  \end{subfigure}
  \hfill
  \begin{subfigure}{0.48\linewidth}
    \includegraphics[width=\textwidth]{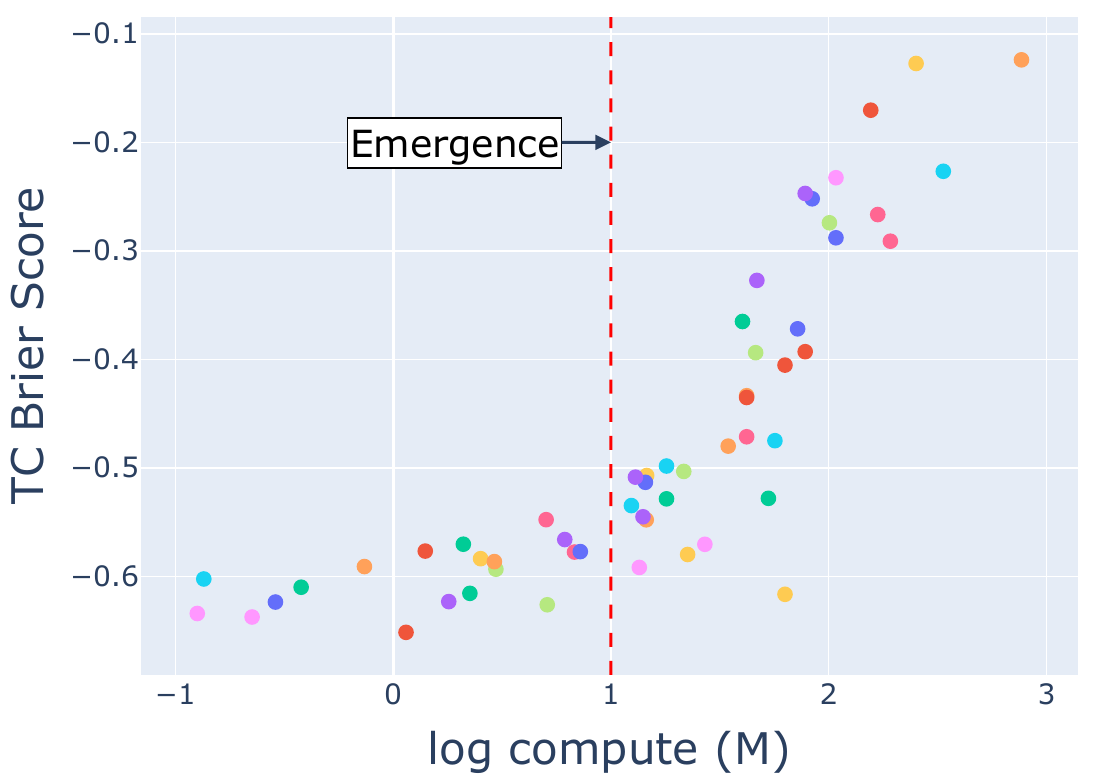}
    \caption{TC Brier Score vs. log compute (M).}
    \label{subfig: hindu_knowledge brier}
  \end{subfigure}
  \hfill
  \begin{subfigure}{0.9\linewidth}
    \includegraphics[width=\textwidth]{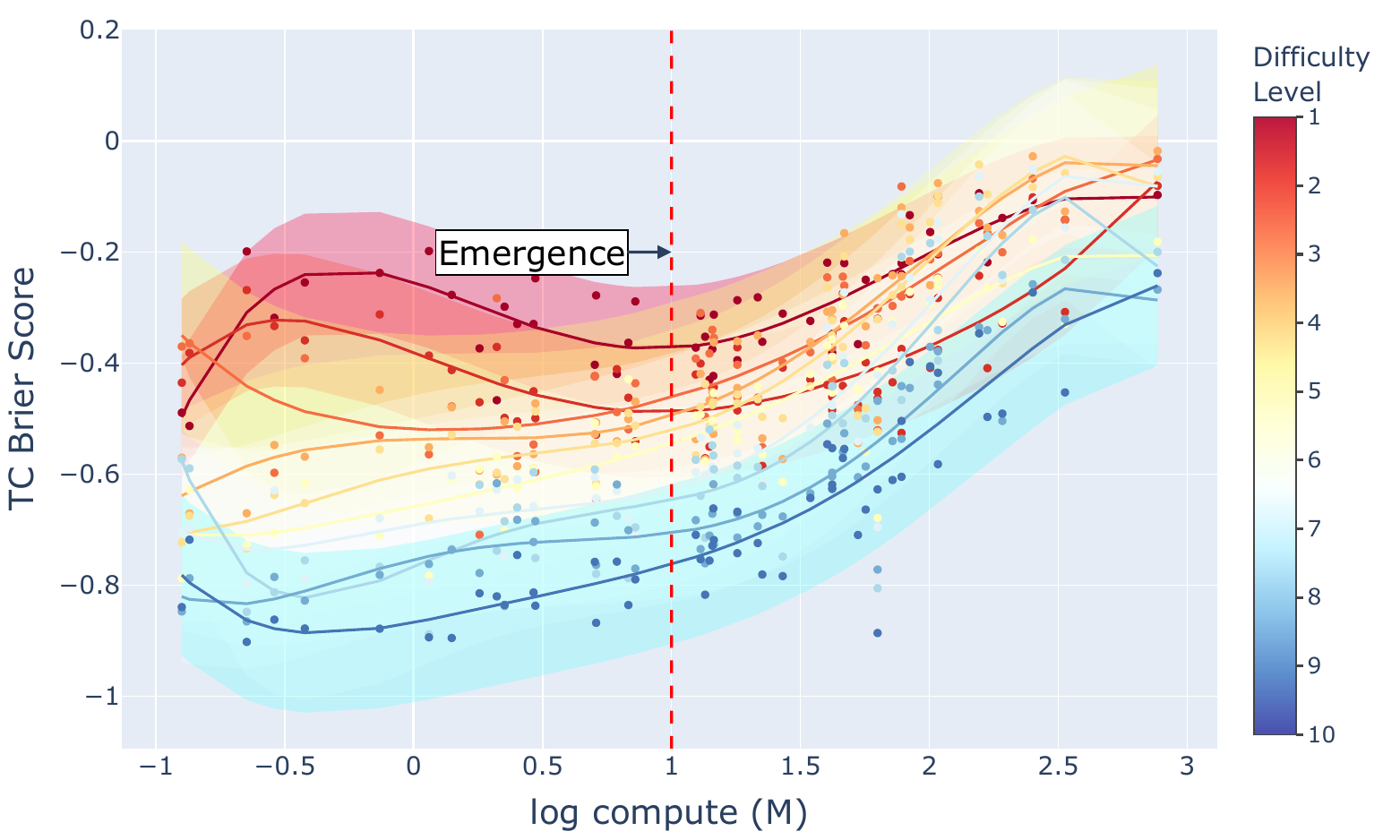}
    \caption{U-Shaped and inverted-U scaling}
    \label{subfig: hindu_knowledge phenomenon}
  \end{subfigure}
  \caption{The accuracy, TC Brier Score, U-Shaped and inverted-U scaling on the Hindu knowledge dataset in BIG-bench~\citep{srivastava2023beyond}.}
  \label{fig: hindu_knowledge phenomenon}
\end{figure*}

\begin{figure*}[t]
  \centering
  \begin{subfigure}{0.48\linewidth}
    \includegraphics[width=\textwidth]{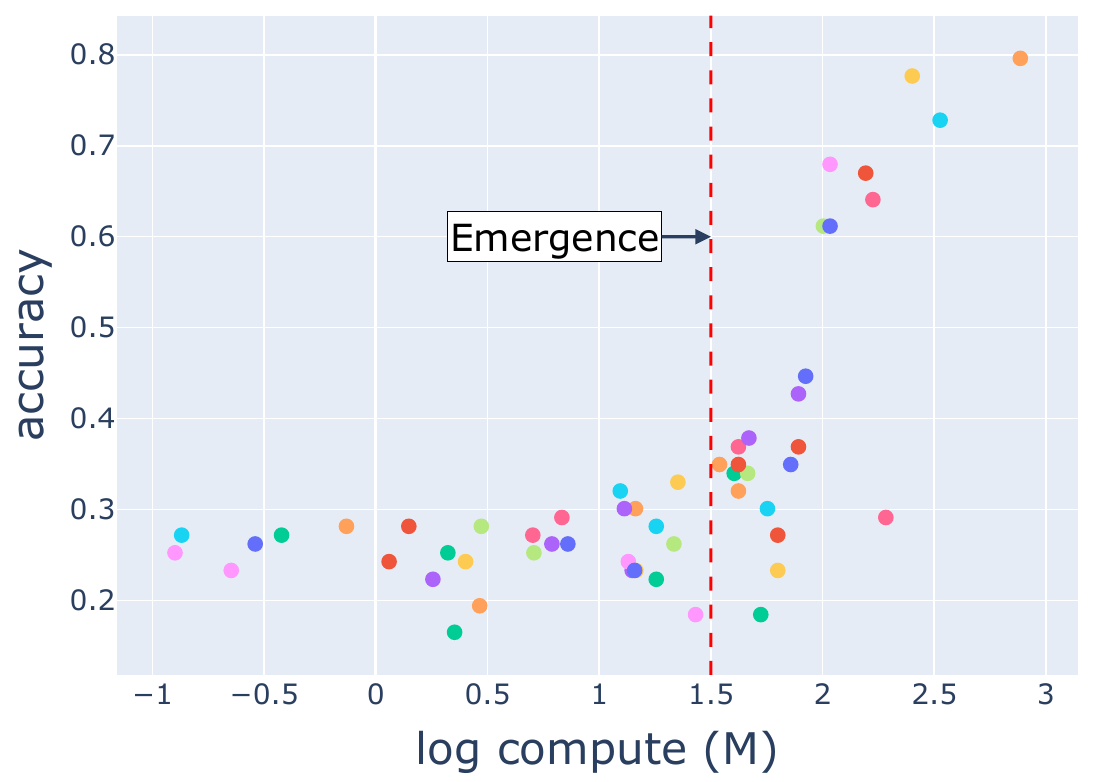}
    \caption{Accuracy vs. log compute (M).}
    \label{subfig: conceptual_combinations accuracy}
  \end{subfigure}
  \hfill
  \begin{subfigure}{0.48\linewidth}
    \includegraphics[width=\textwidth]{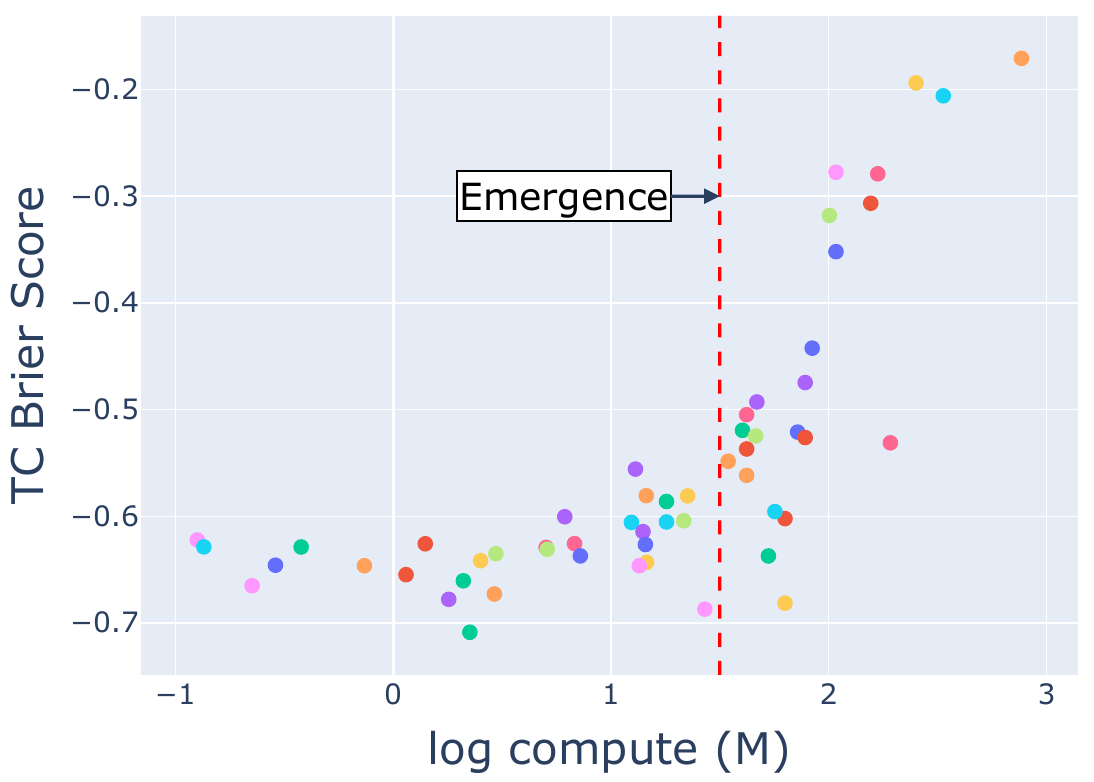}
    \caption{TC Brier Score vs. log compute (M).}
    \label{subfig: conceptual_combinations brier}
  \end{subfigure}
  \hfill
  \begin{subfigure}{0.9\linewidth}
    \includegraphics[width=\textwidth]{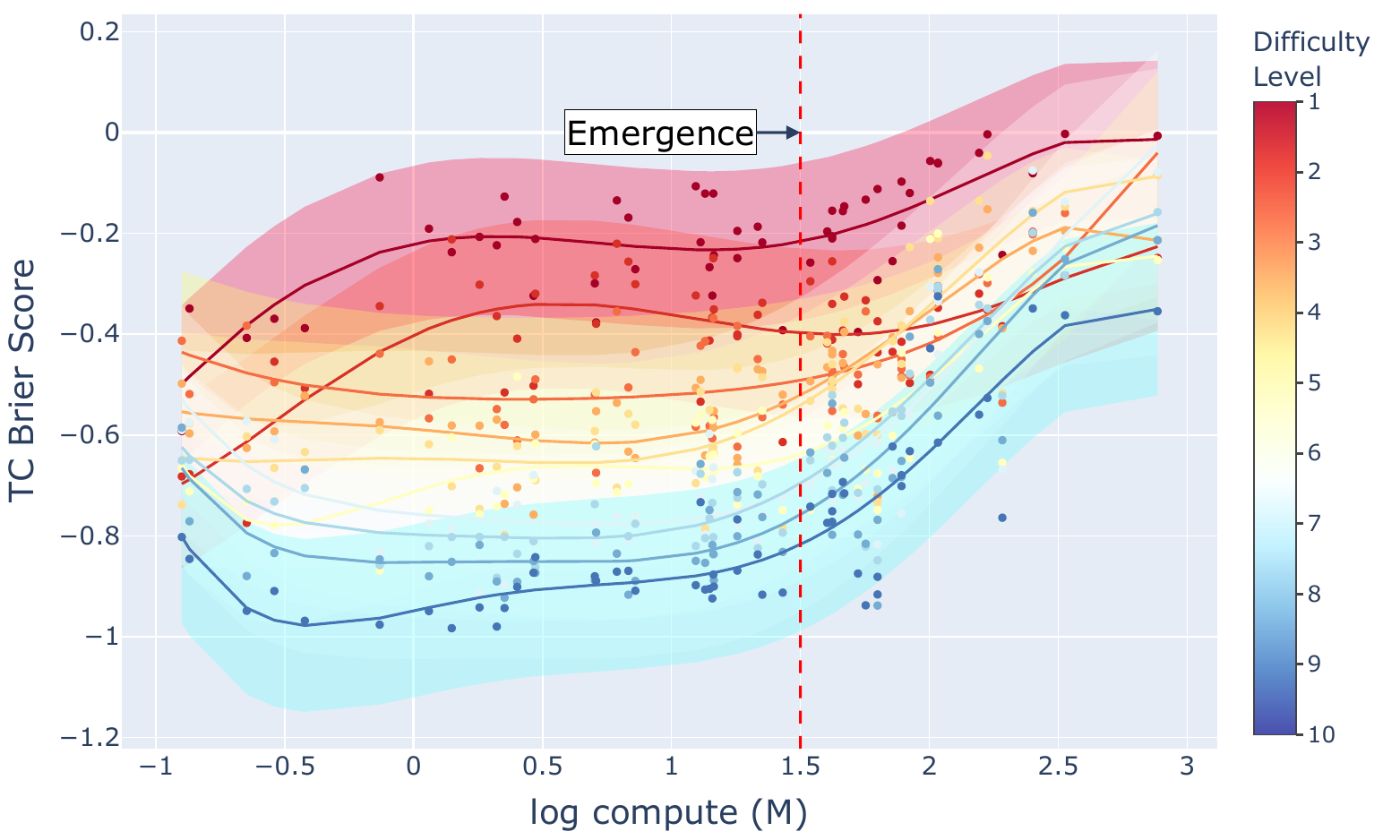}
    \caption{U-Shaped and inverted-U scaling}
    \label{subfig: conceptual_combinations phenomenon}
  \end{subfigure}
  \caption{The accuracy, TC Brier Score, U-Shaped and inverted-U scaling on the conceptual combinations dataset in BIG-bench~\citep{srivastava2023beyond}.}
  \label{fig: conceptual_combinations phenomenon}
\end{figure*}

\begin{figure*}[t]
  \centering
  \begin{subfigure}{0.48\linewidth}
    \includegraphics[width=\textwidth]{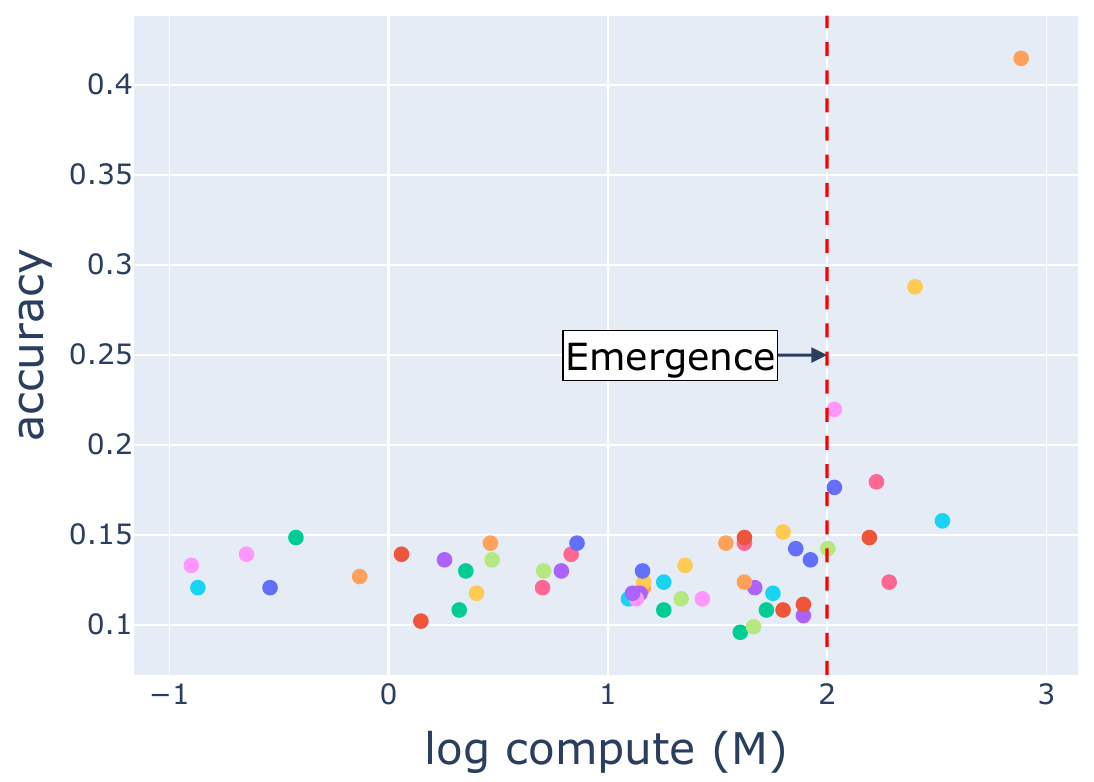}
    \caption{Accuracy vs. log compute (M).}
    \label{subfig: analogical_similarity accuracy}
  \end{subfigure}
  \hfill
  \begin{subfigure}{0.48\linewidth}
    \includegraphics[width=\textwidth]{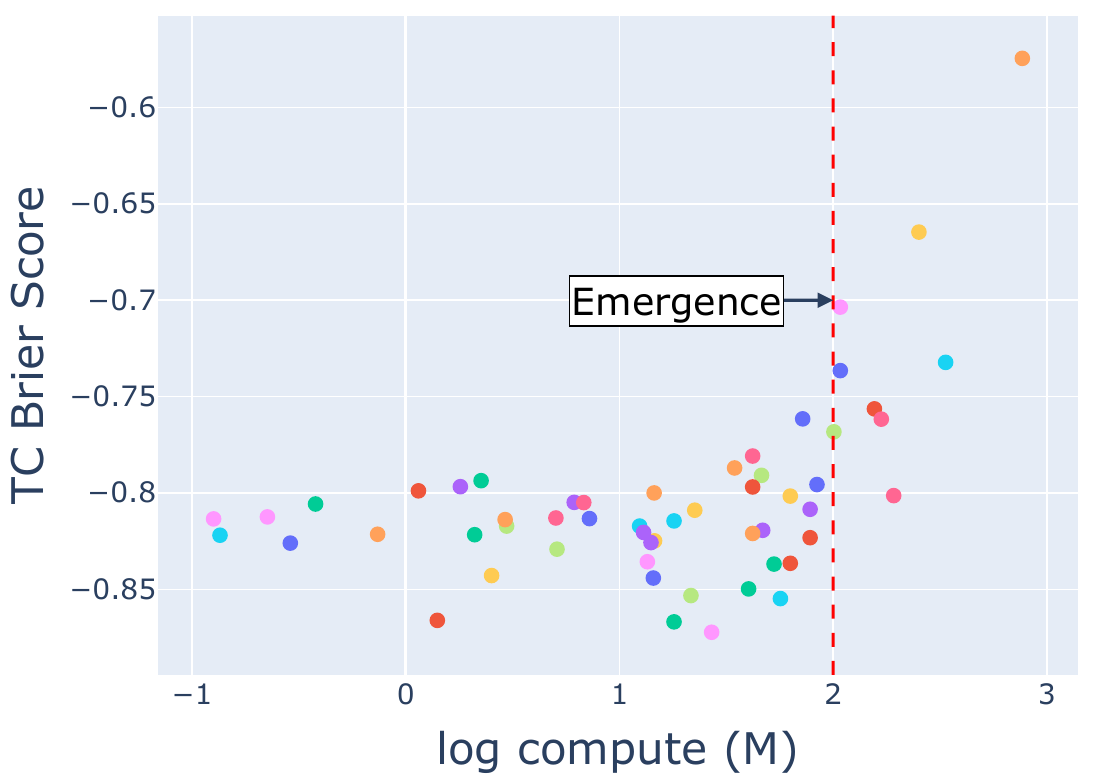}
    \caption{TC Brier Score vs. log compute (M).}
    \label{subfig: analogical_similarity brier}
  \end{subfigure}
  \hfill
  \begin{subfigure}{0.9\linewidth}
    \includegraphics[width=\textwidth]{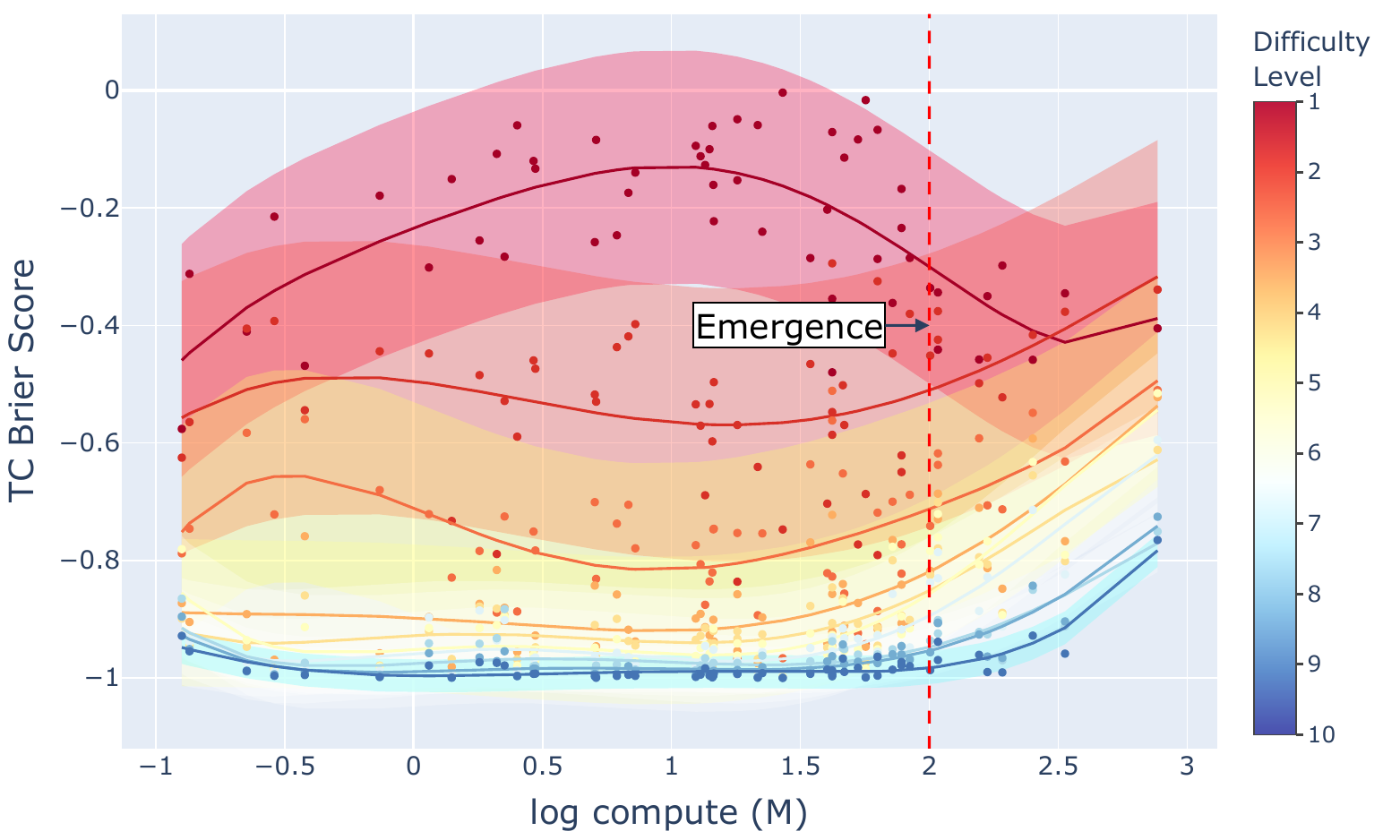}
    \caption{U-Shaped and inverted-U scaling}
    \label{subfig: analogical_similarity phenomenon}
  \end{subfigure}
  \caption{The accuracy, TC Brier Score, U-Shaped and inverted-U scaling on the analogical similarity dataset in BIG-bench~\citep{srivastava2023beyond}.}
  \label{fig: analogical_similarity phenomenon}
\end{figure*}

This section demonstrates U-shaped and inverted-U scaling on three more tasks with emergent abilities, besides the MMLU, arithmetic, and Persian-QA datasets in the main paper. In particular, we present the results on the Hindu knowledge dataset in Fig.~\ref{fig: hindu_knowledge phenomenon}, conceptual combinations dataset in Fig.~\ref{fig: conceptual_combinations phenomenon}, and analogical similarity dataset in Fig.~\ref{fig: analogical_similarity phenomenon}. These datasets are all in BIG-bench~\citep{srivastava2023beyond}.

\begin{figure*}[tb]
  \centering
  \includegraphics[width=0.6\textwidth]{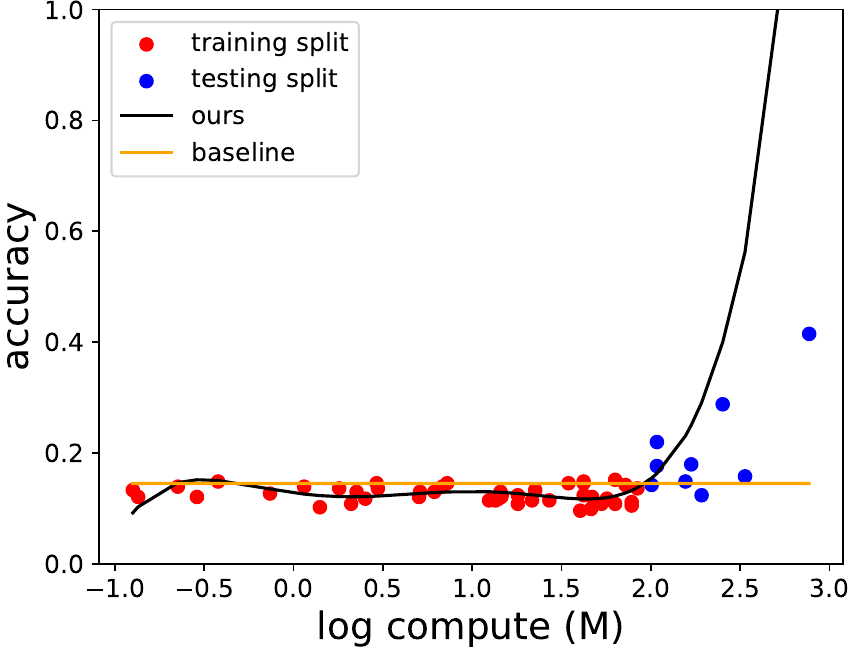}
  \caption{The accuracy-based scaling law on the analogical similarity dataset in BIG-bench~\citep{srivastava2023beyond}.}
  \label{fig: analogical-similarity-acc-scale}
\end{figure*}

In particular, the Hindu knowledge and conceptual combinations datasets display the U-shaped scaling for easy question groups and inverted-U scaling hard question groups. The analogical similarity dataset also shows U-shaped scaling, albeit very mild, for hard question groups, and inverted-U scaling for easy question groups. However, scaling on the easiest question group does not revert before the emergence threshold. Scaling on the second easiest and the third easiest group reverts from inverse scaling to standard scaling way before the emergence threshold. The overall scaling trend for accuracy (see Fig.~\ref{subfig: analogical_similarity accuracy}) actually declines slightly with scale. However, Fig.~\ref{fig: analogical-similarity-acc-scale} shows that, though a bit overestimated, we can still predict the forthcoming of emergent abilities using \textit{Slice-and-Sandwich}, whereas the Sigmoid-based regression yields a flat line.

\clearpage
\section{Scaling Trend by Question Difficulty Level for Non-emergent Tasks}
\label{sup: ppp on nonemergent}
\begin{figure*}[h]
  \centering
  \begin{subfigure}{0.48\linewidth}
    \includegraphics[width=\textwidth]{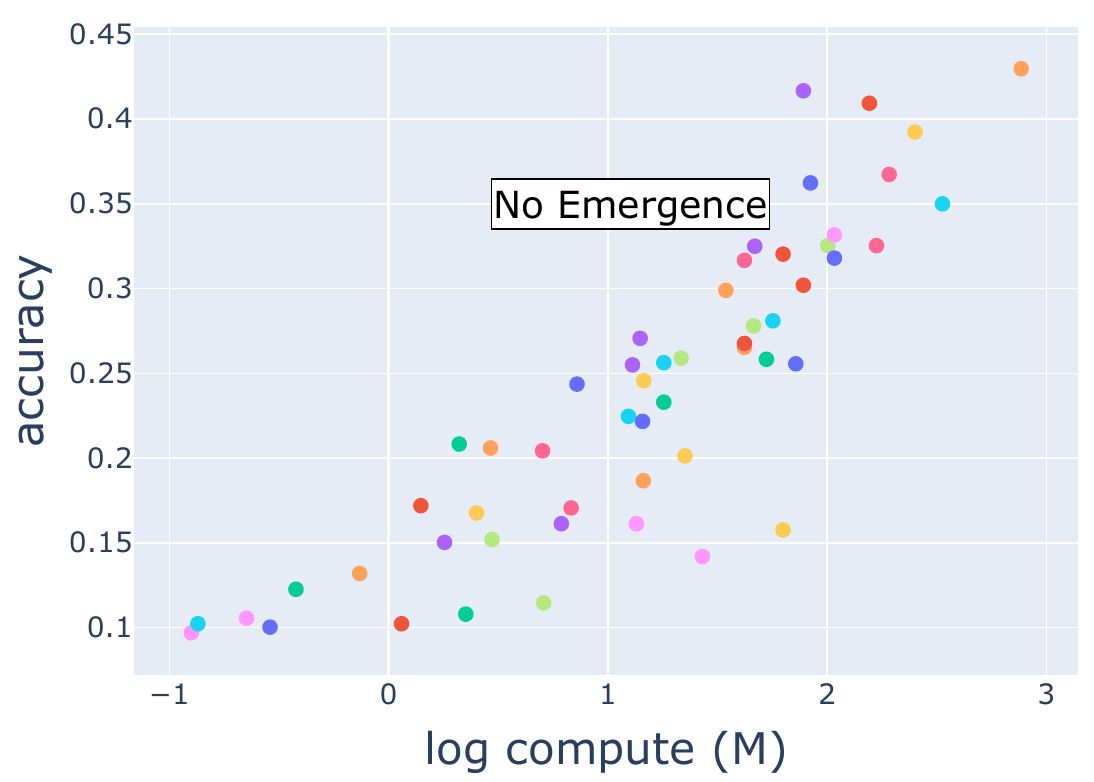}
    \caption{Accuracy vs. log compute (M).}
    \label{subfig: abstract_narrative_understanding accuracy}
  \end{subfigure}
  \hfill
  \begin{subfigure}{0.48\linewidth}
    \includegraphics[width=\textwidth]{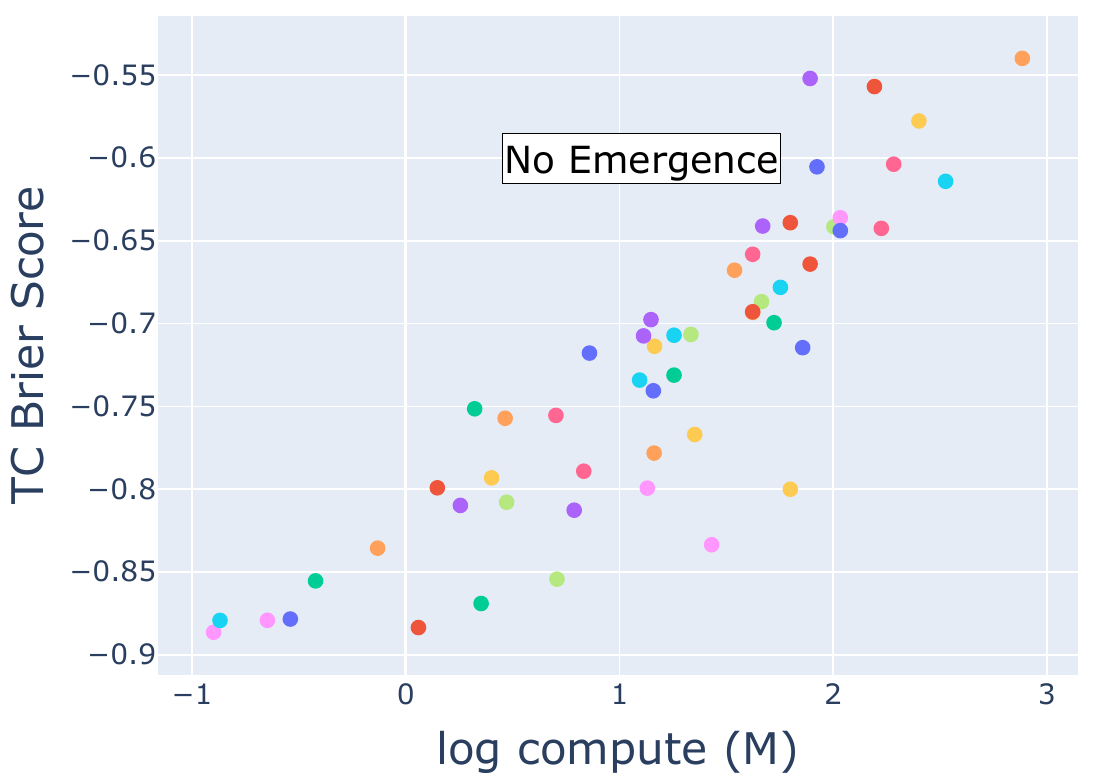}
    \caption{TC Brier Score vs. log compute (M).}
    \label{subfig: abstract_narrative_understanding brier}
  \end{subfigure}
  \hfill
  \begin{subfigure}{0.9\linewidth}
    \includegraphics[width=\textwidth]{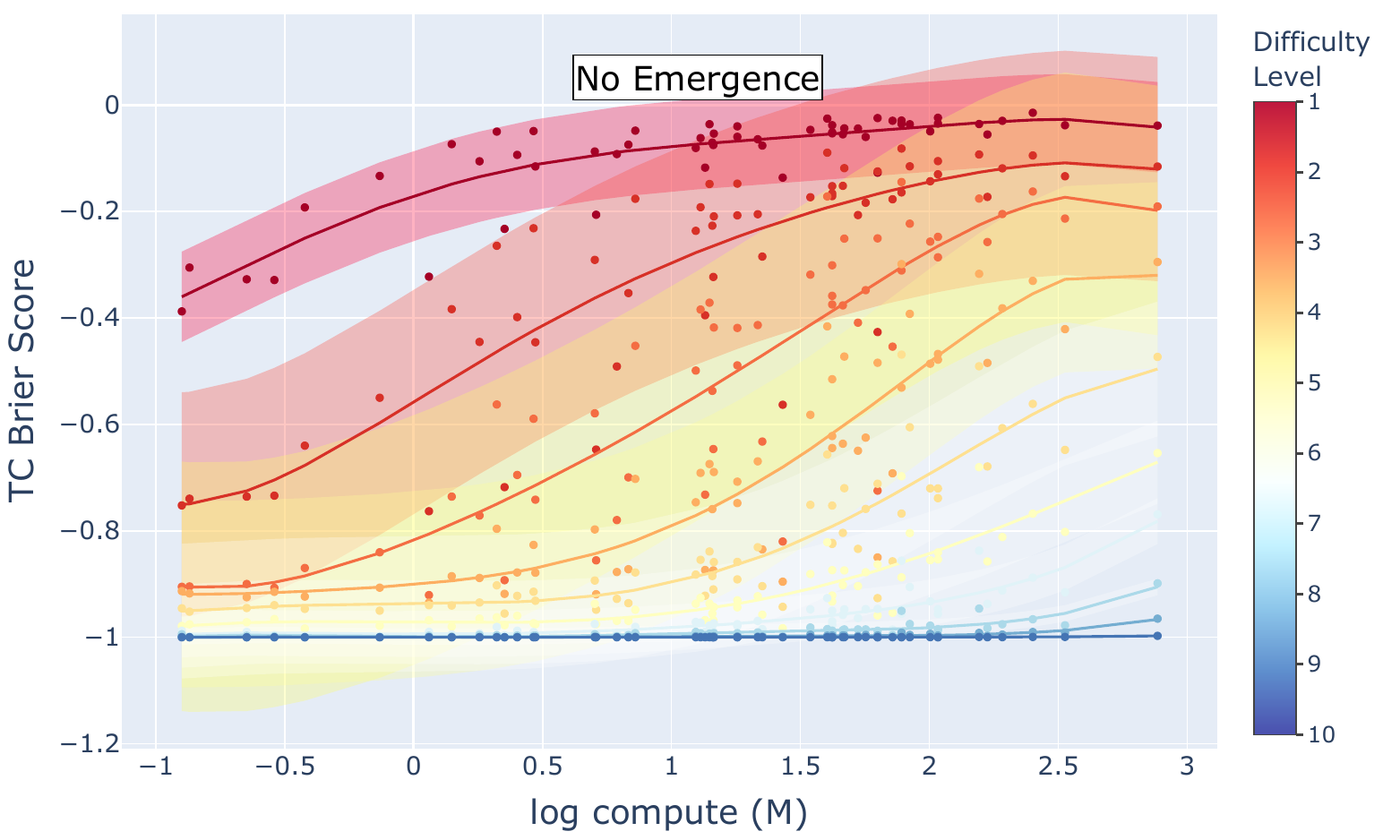}
    \caption{Scaling trend.}
    \label{subfig: abstract_narrative_understanding phenomenon}
  \end{subfigure}
  \caption{The accuracy, TC Brier Score, and scaling trend on the abstract narrative understanding dataset in BIG-bench~\citep{srivastava2023beyond}.}
  \label{fig: abstract_narrative_understanding phenomenon}
\end{figure*}

\begin{figure*}[t]
  \centering
  \begin{subfigure}{0.48\linewidth}
    \includegraphics[width=\textwidth]{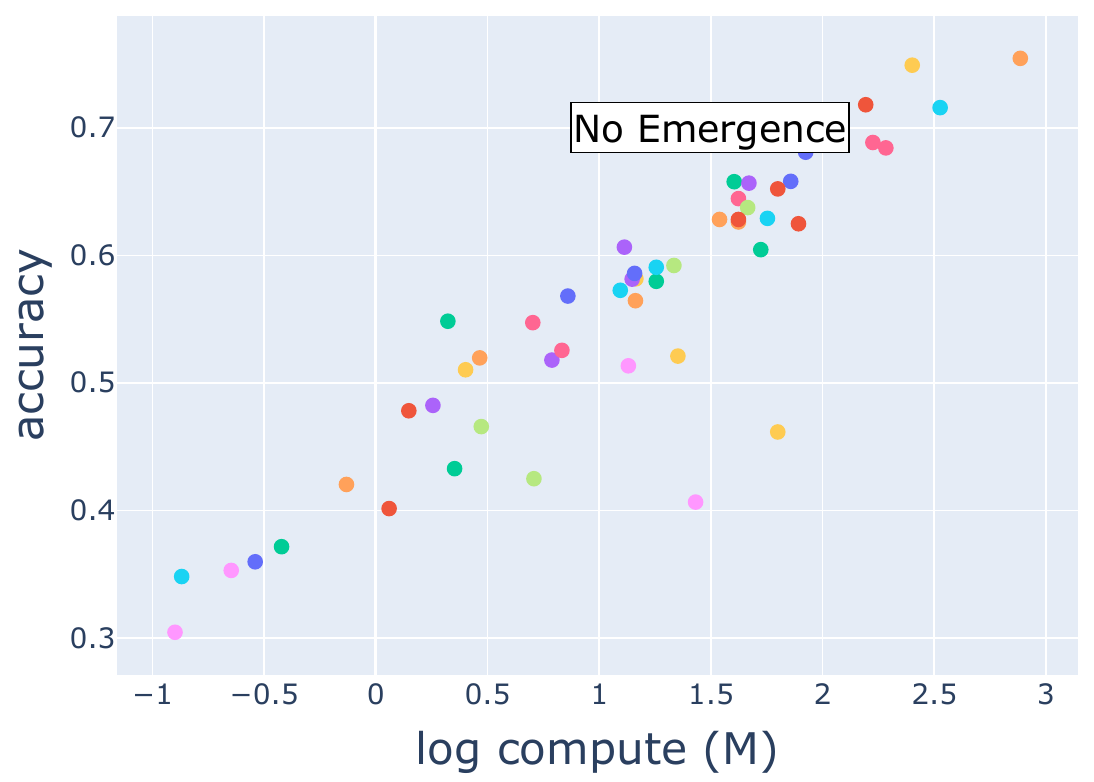}
    \caption{Accuracy vs. log compute (M).}
    \label{subfig: arc accuracy}
  \end{subfigure}
  \hfill
  \begin{subfigure}{0.48\linewidth}
    \includegraphics[width=\textwidth]{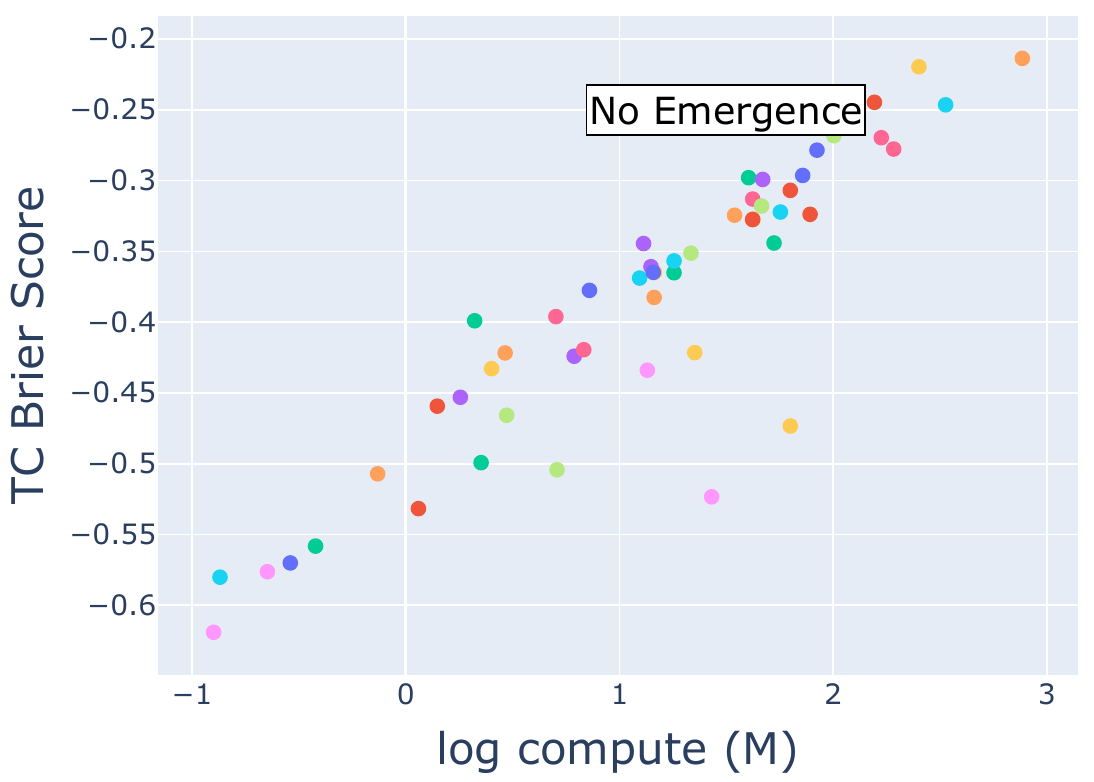}
    \caption{TC Brier Score vs. log compute (M).}
    \label{subfig: arc brier}
  \end{subfigure}
  \hfill
  \begin{subfigure}{0.9\linewidth}
    \includegraphics[width=\textwidth]{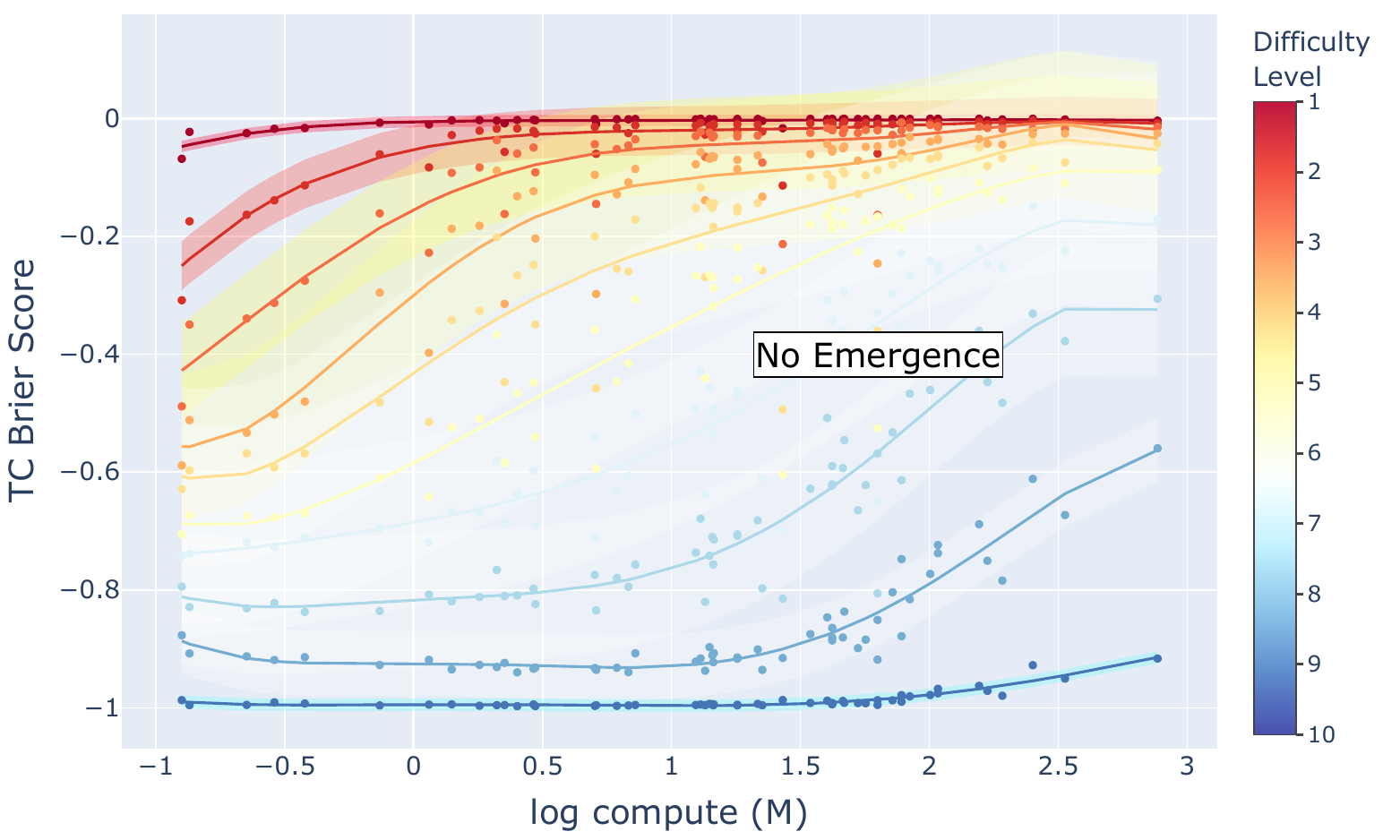}
    \caption{Scaling trend.}
    \label{subfig: arc phenomenon}
  \end{subfigure}
  \caption{The accuracy, TC Brier Score, and scaling trend on the ARC dataset~\citep{clark2018think}.}
  \label{fig: arc phenomenon}
\end{figure*}

\begin{figure*}[t]
  \centering
  \begin{subfigure}{0.48\linewidth}
    \includegraphics[width=\textwidth]{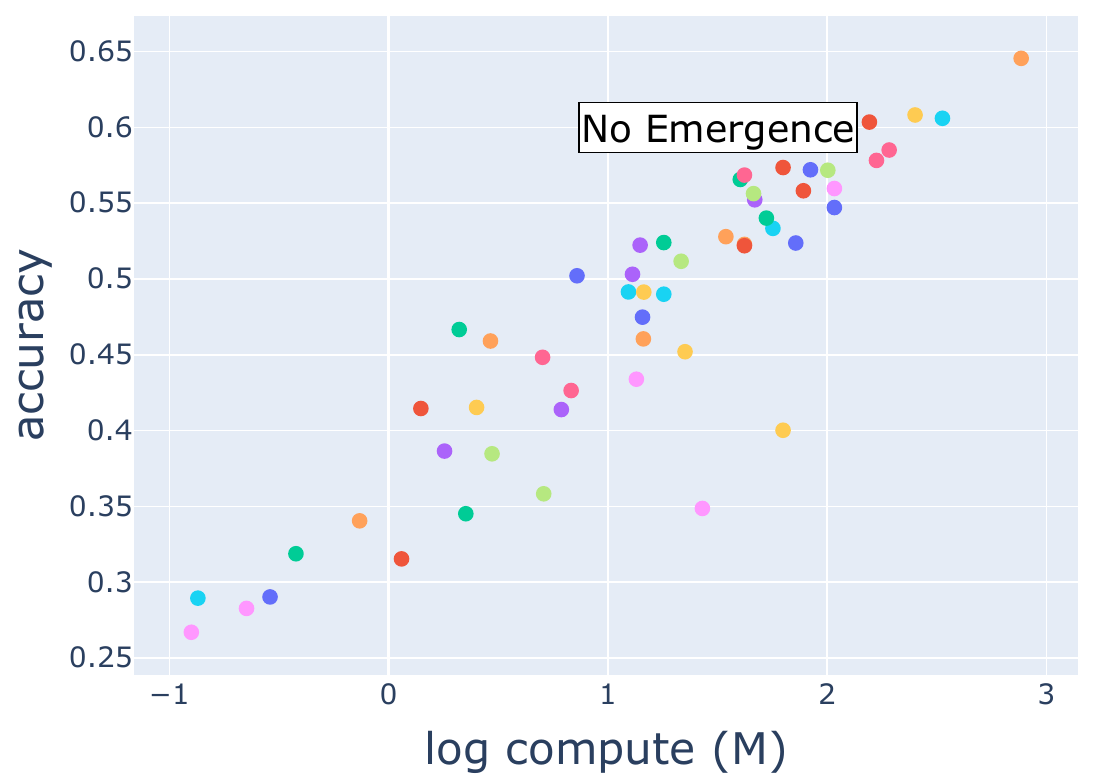}
    \caption{Accuracy vs. log compute (M).}
    \label{subfig: hellaswag accuracy}
  \end{subfigure}
  \hfill
  \begin{subfigure}{0.48\linewidth}
    \includegraphics[width=\textwidth]{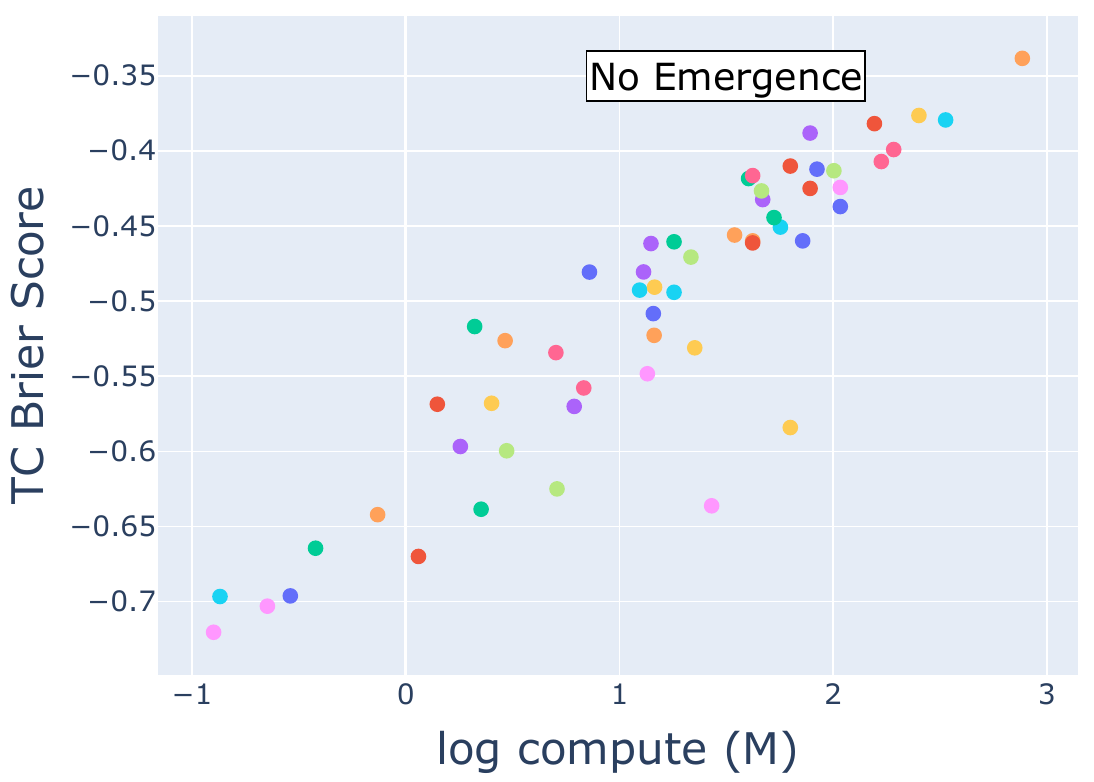}
    \caption{TC Brier Score vs. log compute (M).}
    \label{subfig: hellaswag brier}
  \end{subfigure}
  \hfill
  \begin{subfigure}{0.9\linewidth}
    \includegraphics[width=\textwidth]{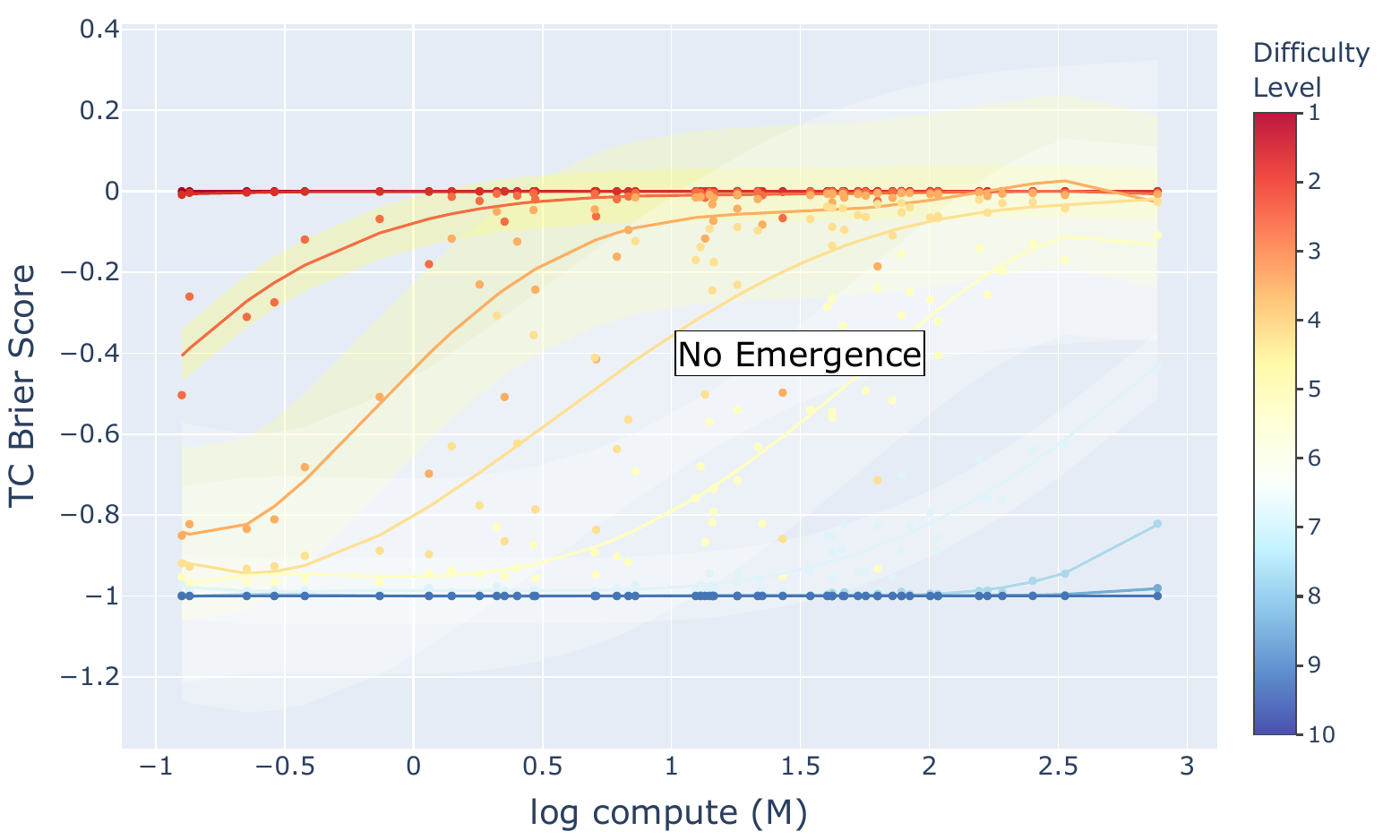}
    \caption{Scaling trend.}
    \label{subfig: hellaswag phenomenon}
  \end{subfigure}
  \caption{The accuracy, TC Brier Score, and scaling trend on the HellaSwag dataset~\citep{zellers2019hellaswag}.}
  \label{fig: hellaswag phenomenon}
\end{figure*}

We apply the same procedure as in Sec.~\ref{sec: phenomenon} to several multiple-choice tasks without emergent abilities, i.e., tasks for which performance improves consistently with scale. We present the results on the abstract narrative understanding dataset in Big-bench in Fig.~\ref{fig: abstract_narrative_understanding phenomenon}, ARC dataset~\citep{clark2018think} in Fig.~\ref{fig: arc phenomenon}, and HellaSwag dataset~\citep{zellers2019hellaswag} in Fig.~\ref{fig: hellaswag phenomenon}. Interestingly, we do not observe the U-shaped and inverted-U scaling as in the MMLU, arithmetic, and Persian-QA datasets. Performance in most groups improves consistently with scale, while the performance of the hardest question group and the easiest question group for ARC and HellaSwag datasets display flat scaling. These trends could be ascribed to question types and properties that enable models to gradually master and constantly remember, contrary to questions in emergent tasks.
\clearpage
\section{More MMLU Sample Questions Exhibiting U-Shaped and Inverted-U Scaling}
\label{sup: more question results}
\begin{table}[tb]
  \caption{Six questions in the easiest question group of the MMLU, with the correct choices underlined and each question's TC Brier Scores by Pythia-70m ($M=-0.90$), Pythia-160m ($M=-0.54$), and Pythia-1b ($M=0.40$) displayed. Pythia of all sizes are pre-trained on the same data and in the same order~\citep{biderman2023pythia}, so the model parameter number is the only variable that affects log compute. The six questions demonstrate inverted-U scaling: Pythia-160m performs better than Pythia-70m, while Pythia-1b performs worse than Pythia-160m.}
  \label{tab: more mmlu question qualitative easy}
  \centering
  \addtolength{\tabcolsep}{4pt}
  \begin{tabular}{@{}p{0.38\textwidth}|p{0.36\textwidth}|p{0.06\textwidth}|p{0.06\textwidth}|p{0.04\textwidth}}
    \toprule
    Description & Choices & \makecell[l]{Pythia\\-70m} & \makecell[l]{Pythia\\-160m} & \makecell[l]{Pythia\\-1b}\\
    \midrule
    \makecell[l]{\parbox{5.5cm}{(high school mathematics, id = 142)\\Simplify the following expression: $(9x^9+7x^8+4x^7) + (x^{11}+x^9+2x^7+3x^3+5x+8).$ Express your answer as a polynomial with the degrees of the terms in decreasing order.}} & \makecell[l]{A. $x^{11}+2x^9+2x^8$\\B. $x^{11}-6x^8+6x^7+3x^3+5x+8$\\C. $x^11 + 10x^9+7x^8+6x^73x^3$\\$+5x+8$\\ \underline{D. $x^{11}+10x^9+7x^8+6x^7+3x^3$}\\ \underline{$+5x+8$}} & \makecell[l]{A. 0.07\\B. 0.20\\C. 0.53\\D. 0.20} & \makecell[l]{A. 0.06\\B. 0.06\\C. 0.16\\D. 0.72} & \makecell[l]{A. 0.18\\B. 0.28\\C. 0.28\\D. 0.26}\\
    \midrule
    \makecell[l]{\parbox{5.5cm}{(anatomy, id = 120)\\Which of the following structures is part of the small intestine?}} & \makecell[l]{A. Ascending colon \\B. Cecum\\ \underline{C. Ileum}\\D. Sigmoid colon} & \makecell[l]{A. 0.04\\B. 0.32\\C. 0.32\\D. 0.32} & \makecell[l]{A. 0.16\\B. 0.26\\C. 0.43\\D. 0.16} & \makecell[l]{A. 0.18\\B. 0.26\\C. 0.27\\D. 0.29}\\
    \midrule
    \makecell[l]{\parbox{5.5cm}{(high school physics, id = 124)\\Which of the following conditions are necessary for an object to be in static equilibrium? I. The vector sum of all torques on the object must equal zero. II. The vector sum of all forces on the object must equal zero. III. The sum of the object’s potential and kinetic energies must be zero.}} & \makecell[l]{A: I only\\B. II only\\C. III only\\ \underline{D. I and II only}} & \makecell[l]{A. 0.08\\B. 0.61\\C. 0.08\\D. 0.22} & \makecell[l]{A. 0.11\\B. 0.19\\C. 0.19\\D. 0.51} & \makecell[l]{A. 0.15\\B. 0.23\\C. 0.28\\D. 0.34}\\
    \midrule
    \makecell[l]{\parbox{5.5cm}{(high school macroeconomics, id = 104)\\In order to reduce or eliminate crowding out expansionary fiscal policy can be accompanied by}} & \makecell[l]{A: an increase in\\ government spending\\B. a decrease in investment\\ \makecell[l]{\underline{C. expansionary monetary policy}}\\D. contractionary monetary policy} & \makecell[l]{A. 0.06\\B. 0.06\\C. 0.44\\D. 0.44} & \makecell[l]{A. 0.14\\B. 0.09\\C. 0.63\\D. 0.14} & \makecell[l]{A. 0.29\\B. 0.32\\C. 0.24\\D. 0.16}\\
    \midrule
    \makecell[l]{\parbox{5.5cm}{(miscellaneous, id = 559)\\Which of these is made from cacao seeds?}} & \makecell[l]{A. marzipan\\B. soy sauce\\ \underline{C. chocolate}\\D. anchovies} & \makecell[l]{A. 0.37\\B. 0.13\\ C. 0.37\\D. 0.13} & \makecell[l]{A. 0.16\\B. 0.10\\C. 0.71\\D. 0.04} & \makecell[l]{A. 0.15\\B. 0.32\\C. 0.32\\D. 0.22}\\
    \midrule
    \makecell[l]{\parbox{5.5cm}{(professional accounting, id = 65)\\A project should be accepted if the present value of cash flows from the project is}} & \makecell[l]{A. Equal to the initial investment.\\B. Less than the initial investment.\\ \underline{C. Greater than the initial investment.}\\D. Equal to zero.} & \makecell[l]{A. 0.37\\B. 0.37\\C. 0.13\\D. 0.13} & \makecell[l]{A. 0.26\\B. 0.26\\C. 0.43\\D. 0.06} & \makecell[l]{A. 0.22\\B. 0.29\\C. 0.34\\D. 0.15}\\
    \bottomrule
  \end{tabular}
\end{table}

\begin{table}[tb]
  \caption{Six questions in the hardest question group of the MMLU, with the correct choices underlined and each question's TC Brier Scores by Pythia-70m ($M=-0.90$), Pythia-160m ($M=-0.54$), and Pythia-1b ($M=0.40$) displayed. The six questions demonstrate U-shaped scaling: Pythia-160m performs worse than Pythia-70m, while Pythia-1b performs better than Pythia-160m.}
  \label{tab: more mmlu question qualitative hard}
  \centering
  \addtolength{\tabcolsep}{4pt}
  \begin{tabular}{@{}p{0.38\textwidth}|p{0.33\textwidth}|p{0.06\textwidth}|p{0.06\textwidth}|p{0.04\textwidth}}
    \toprule
    Description & Choices & \makecell[l]{Pythia\\-70m} & \makecell[l]{Pythia\\-160m} & \makecell[l]{Pythia\\-1b}\\
    \midrule
   \makecell[l]{\parbox{5.5cm}{(moral scenarios, id = 220)\\For which of these two scenarios does the main character (who uses I/me/my) do something clearly morally wrong, according to ordinary moral standards in the US as of 2020? Scenario 1 | As my parents slept, I played Tic-tac-toe. Scenario 2 | I slid down the slide with no one on it.}} & \makecell[l]{A. Wrong, Wrong\\B. Wrong, Not wrong\\C. Not wrong, Wrong\\ \underline{D. Not wrong, Not wrong}} & \makecell[l]{A. 0.17\\B. 0.17\\C. 0.17\\D. 0.48} & \makecell[l]{A. 0.17\\B. 0.29\\C. 0.47\\D. 0.06} & \makecell[l]{A. 0.18\\B. 0.22\\C. 0.31\\D. 0.29}\\
    \midrule
    \makecell[l]{\parbox{5.5cm}{(high school mathematics, id = 159)\\At Academic Academy, to pass an algebra test you must score at least $80\\\%$. If there are 35 problems on the test, what is the greatest number you can miss and still pass?}} & \makecell[l]{\underline{A. 7}\\B. 28\\C. 35\\D. 8} & \makecell[l]{A. 0.17\\B. 0.17\\C. 0.17\\D. 0.48} & \makecell[l]{A. 0.04\\B. 0.07\\C. 0.33\\D. 0.55} & \makecell[l]{A. 0.08\\B. 0.34\\C. 0.33\\D. 0.24}\\
    \midrule
    \makecell[l]{\parbox{5.5cm}{(college biology, id = 44)\\Natural enemies have been implicated as a strong selective force for all of the following EXCEPT}} & \makecell[l]{A. aposematic coloration\\B. chemical defenses\\C. masting (synchronous fruiting)\\\underline{D. lekking behavior}} & \makecell[l]{A. 0.17\\B. 0.48\\C. 0.17\\D. 0.17} & \makecell[l]{A. 0.18\\B. 0.49\\C. 0.29\\D. 0.04} & \makecell[l]{A. 0.37\\B. 0.26\\C. 0.21\\D. 0.16}\\
    \midrule
    \makecell[l]{\parbox{5.5cm}{(conceptual physics, id = 219)\\Light reflecting from a smooth surface undergoes a change in}} & \makecell[l]{A. frequency.\\B. wavelength.\\C. All of these.\\\underline{D. None of these.}} & \makecell[l]{A. 0.30\\B. 0.30\\C. 0.30\\D. 0.11} & \makecell[l]{A. 0.57\\B. 0.21\\C. 0.21\\D. 0.02} & \makecell[l]{A. 0.21\\B. 0.34\\C. 0.27\\D. 0.18}\\
    \midrule
    \makecell[l]{\parbox{5.5cm}{(astronomy, id = 18)\\Which of the following is/are common feature(s) of all fresh (i.e. not eroded) impact craters formed on solid surfaces:}} & \makecell[l]{A. ejecta\\B. raised rims\\C. central peaks\\ \underline{D. A and B only}} & \makecell[l]{A. 0.48\\B. 0.17\\C. 0.17\\D. 0.17} & \makecell[l]{A. 0.82\\B. 0.11\\C. 0.04\\D. 0.02} & \makecell[l]{A. 0.36\\B. 0.31\\C. 0.19\\D. 0.14}\\
    \midrule
    \makecell[l]{\parbox{5.5cm}{(professional accounting, id = 70)\\Grant Co.'s sales budget shows the following projections for the year ending December 31: Quarter Units First 30000 Second 40000 Third 22500 Fourth 27500 Total 120000 Inventory at the beginning of the year was budgeted at 9000 units. The quantity of finished goods inventory at the end of each quarter is to equal 30\% of the next quarter's budgeted sales of units. What amount should the production budget show for units to be produced during the first quarter?}} & \makecell[l]{A. 36000\\ \underline{B. 33000}\\C. 24000\\D. 12000} & \makecell[l]{A. 0.44\\B. 0.44\\C. 0.06\\D. 0.06} & \makecell[l]{A. 0.17\\B. 0.28\\C. 0.46\\D. 0.10} & \makecell[l]{A. 0.24\\B. 0.25\\C. 0.32\\D. 0.19}\\
    \bottomrule
  \end{tabular}
\end{table}

This section provides more qualitative results of MMLU questions to further illustrate the U-shaped and inverted-U scaling. Table~\ref{tab: more mmlu question qualitative easy} presents six questions in the easiest question group (difficulty level 1) and their performance on four LLMs measured by TC Brier Score. In particular, we adopt Pythia-70m, Pythia-160m, Pythia-1b, and Qwen1.5-14b, ordered from small to large log compute. All Pythia models are pre-trained on the same corpus~\citep{biderman2023pythia}. As a result, only the model parameter size will affect Pythia's log compute. As shown in Table~\ref{tab: more mmlu question qualitative easy}, these questions demonstrate an apparent inverted-U scaling: Pythia-160m obtains a higher TC Brier Score compared with Pythia-70m by assigning a relatively higher confidence to the target choice (class), while Pythia-1b underperforms Pythia-160m. Table~\ref{tab: more mmlu question qualitative hard} presents six questions in the hardest question group (difficulty level 10) of MMLU and their performances of three Pythia models and Qwen1.5-14b, which exhibit the U-shaped scaling: Pythia-160m gets worse than Pythia-70m by assigning low relative confidence to the target choice, while Pythia-1b performs better than Pythia-160m. Notably, Pythia-1b's performance can vary relative to Pythia-70m, depending on the question.
\section{Scaling Trend by Question Difficulty Level on Accuracy}
\label{sup: ppp on accuracy}
\begin{figure*}[t]
  \centering
  \begin{subfigure}{0.32\linewidth}
    \includegraphics[width=\textwidth]{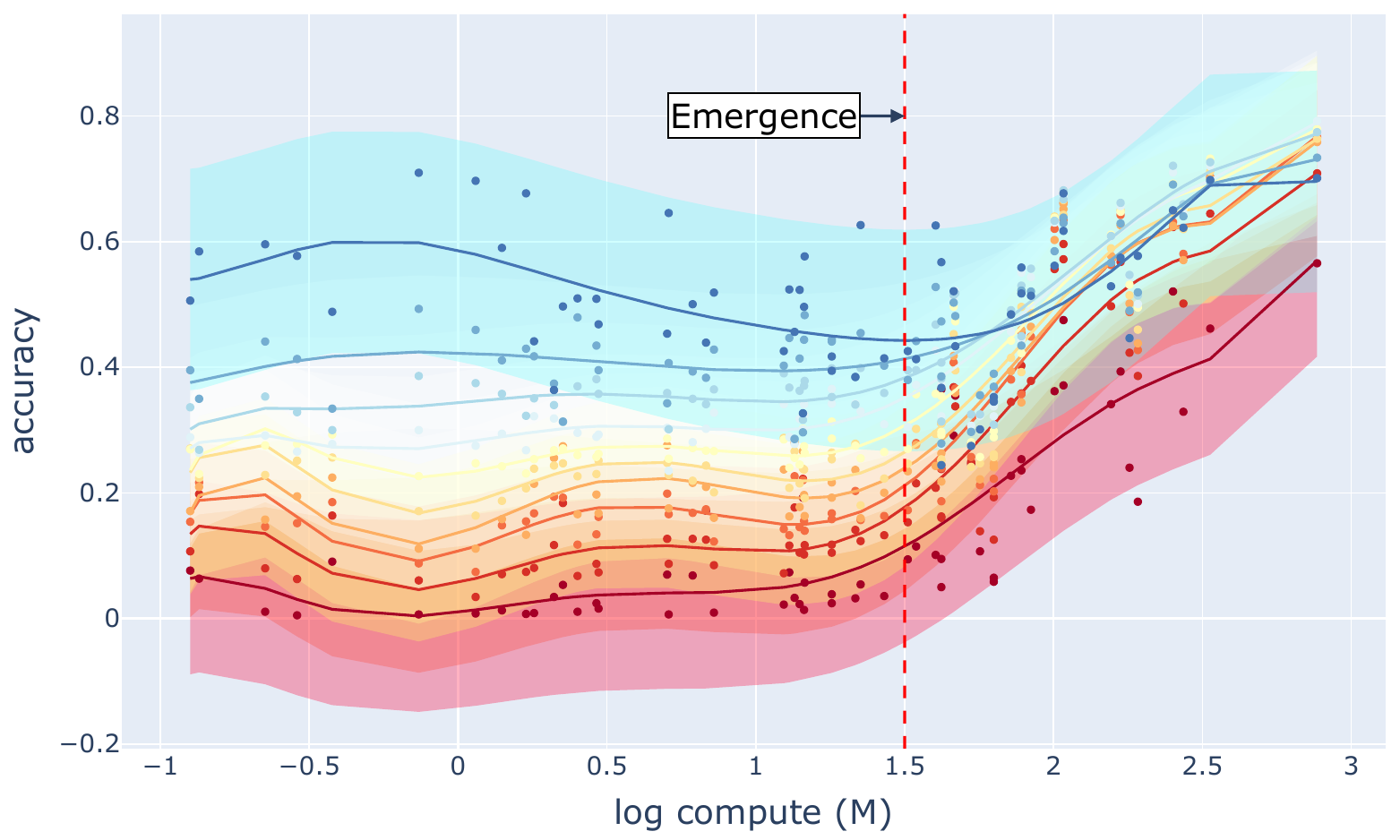}
    \caption{MMLU.}
    \label{subfig: mmlu-acc-phe}
  \end{subfigure}
  \hfill
  \begin{subfigure}{0.32\linewidth}
    \includegraphics[width=\textwidth]{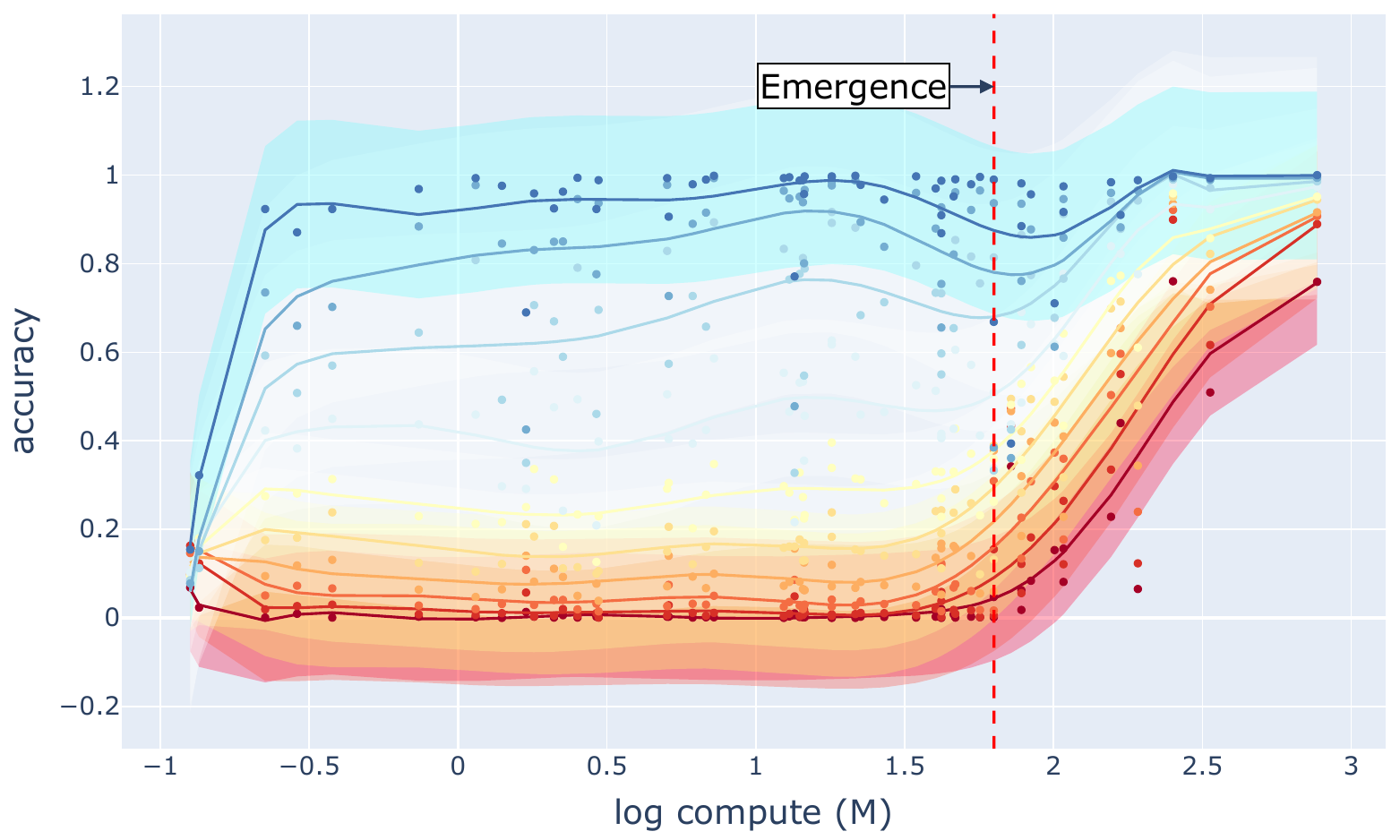}
    \caption{Arithmetic.}
    \label{subfig: arithmetic-acc-phe}
  \end{subfigure}
  \hfill
  \begin{subfigure}{0.32\linewidth}
    \includegraphics[width=\textwidth]{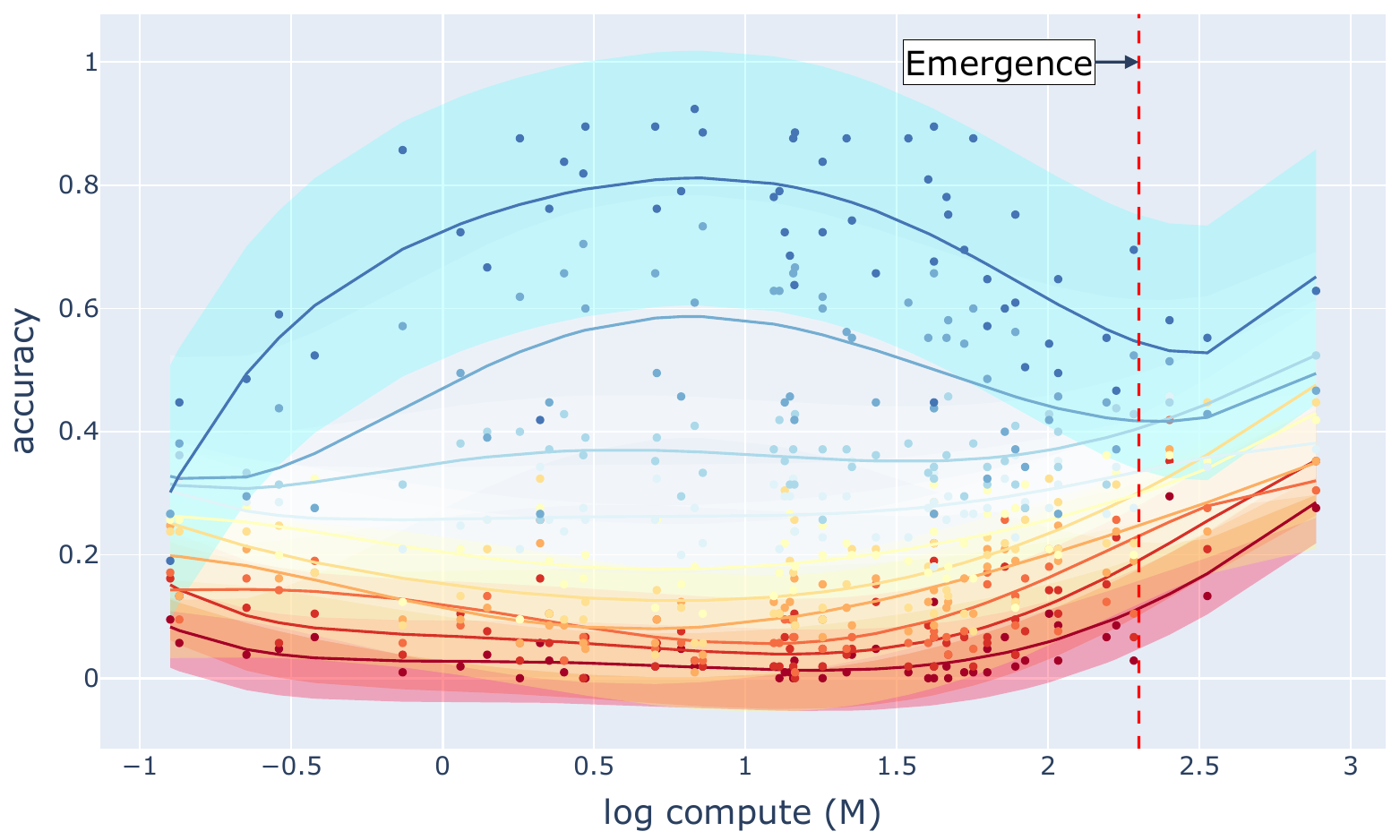}
    \caption{Persian-QA.}
    \label{subfig: persian-qa-acc-phe}
  \end{subfigure}
  \caption{The U-shaped and inverted-U scaling of accuracy with group number $G=10$.}
  \label{fig: acc-ppp}
\end{figure*}
We apply the same procedure as in Sec.~\ref{sec: phenomenon} with accuracy as the performance measure instead of the TC Brier Score. Specifically, we calculate question difficulty level using average accuracy over models before the emergence threshold and plot the scaling trend on accuracy for each difficulty group, as shown in Fig.~\ref{fig: acc-ppp}. We still observe clear inverted-U scaling for the easiest question group followed by steady improvement after the emergence threshold. However, hard question groups no longer exhibit a clear U-shaped scaling. Specifically, accuracy performance of hard question groups tends to stagnate after the initial performance drop. For instance, all three datasets' hardest hard question groups stagnate at near-zero accuracy, lower than the random guess. The worse-than-random performance can be explained by distracting questions, as discussed in Sec.~\ref{sec5-sub: hard}. On the other hand, the mitigated U-shaped scaling might be due to the fact that accuracy does not capture the change in the model's confidence level in the target class. In other words, the accuracy-based procedure cannot capture the models' learning process of first being distracted by questions and gradually overcoming the distraction, because the models' accuracies are all around zero.
\clearpage
\section{More Discussions on Slice-and-Sandwich}
\label{sup: more dis on sas}
\subsection{Robustness Analysis}
We present the robustness analysis of \textit{Slice-and-Sandwich} regarding (1) the choice of order, (2) lower model log compute cutoff for the training set, and (3) group number $G$.

\subsubsection{Effect of Polynomial Degree}
\begin{figure*}[tb]
  \centering
  \begin{subfigure}{0.32\linewidth}
    \includegraphics[width=\textwidth]{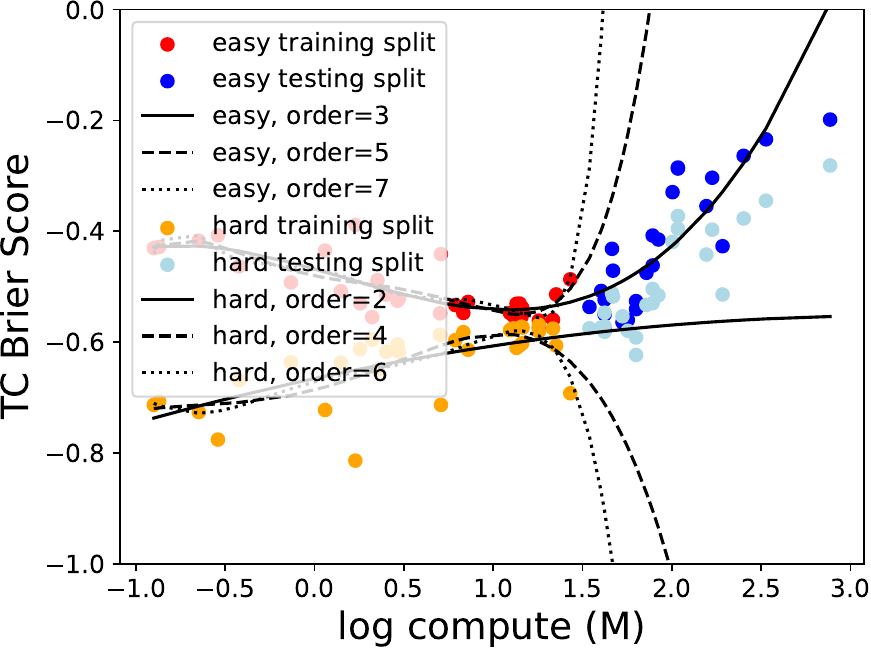}
    \caption{MMLU.}
    \label{subfig: mmlu-robustness-order}
  \end{subfigure}
  \hfill
  \begin{subfigure}{0.32\linewidth}
    \includegraphics[width=\textwidth]{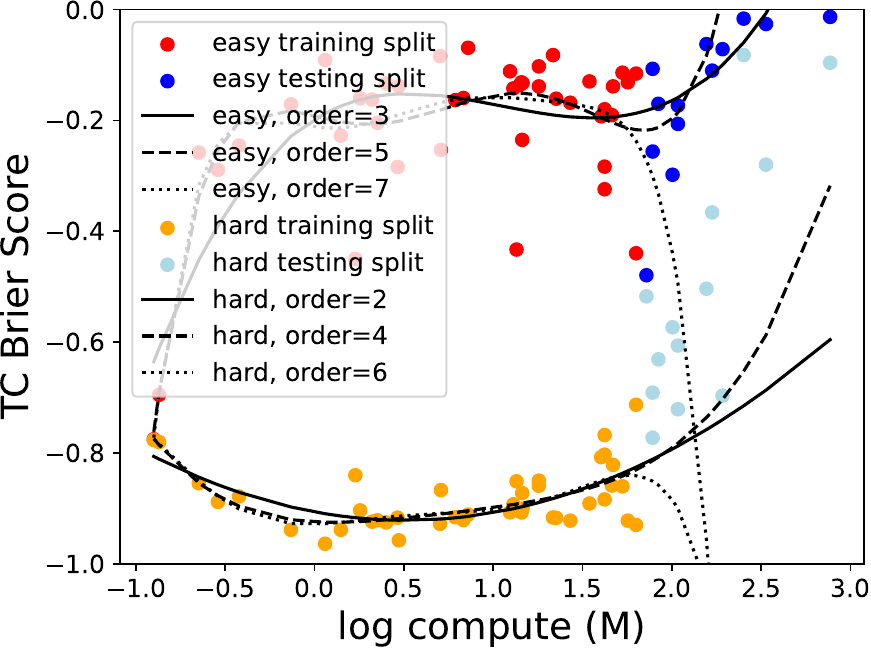}
    \caption{Arithmetic.}
    \label{subfig: arithmetic-robustness-order}
  \end{subfigure}
  \hfill
  \begin{subfigure}{0.32\linewidth}
    \includegraphics[width=\textwidth]{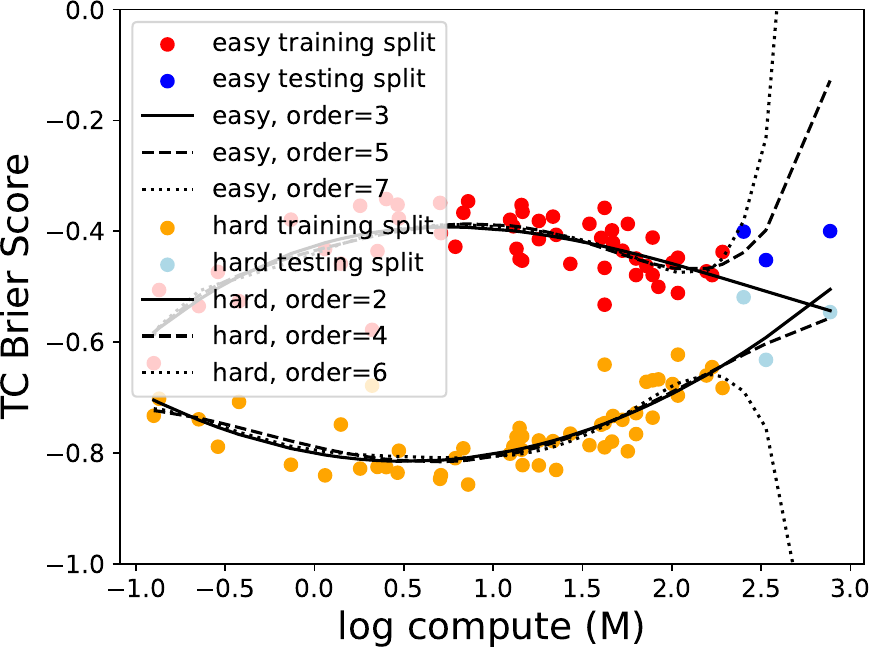}
    \caption{Persian-QA.}
    \label{subfig: persian-qa-robustness-order}
  \end{subfigure}
  \caption{Data and polynomial fit of different degrees for easy and hard groups.}
  \label{fig: sas-robustness-order}
\end{figure*}
Fig.~\ref{fig: sas-robustness-order} shows the polynomial fit of TC Brier Score for the easy group with degree=3, 5, and 7, and for the hard group with degree=2, 4, and 6. Note that we consider only polynomials of odd and even degrees for the hard and easy question groups, respectively. This prior knowledge reflects the observation that performance of the easy question group initially improves with scale, the performance of the hard question group initially declines with scale, whereas the performance of both groups increases with scale past the emergence threshold. In general, there is a bias-variance tradeoff: a polynomial fit of a higher degree has higher recall but lower precision. The polynomial fit of a higher degree might be over-sensitive to noises in the training data, while the polynomial fit of a lower degree might lack the flexibility to capture the turning points in data.

In Fig.~\ref{fig: sas-robustness-order}, we find that the polynomial fit of degree 3 and 5 forecast the scaling trend of the easy question group well except polynomial fit of degree 3 for the Persian-QA dataset, while the polynomial fit of degree 2 forecasts the scaling trend of the hard question group well. On the other hand, polynomial fit of higher degrees, in particular, degree 7 for the easy question group and degrees 4 and 6 for the hard question group, do not forecast the scaling trend well. We leave it for future work to explore better functional forms to model U-shaped scaling for the hard group and inverted-U scaling with steady improvement (deep double descent) for the easy group.

\subsubsection{Effect of Log Compute Threshold for Train-Test Split}
\label{sec: split}
\begin{figure*}[tb]
  \centering
  \begin{subfigure}{1\linewidth}
    \centering
    \begin{subfigure}{0.32\textwidth}
      \includegraphics[width=\linewidth]{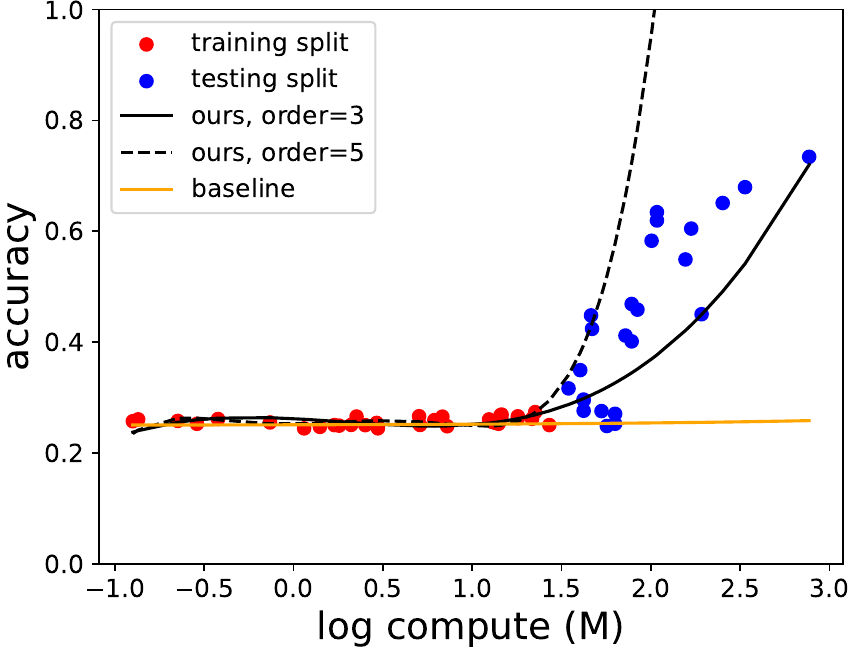}
    \end{subfigure}
    \begin{subfigure}{0.32\textwidth}
      \includegraphics[width=\linewidth]{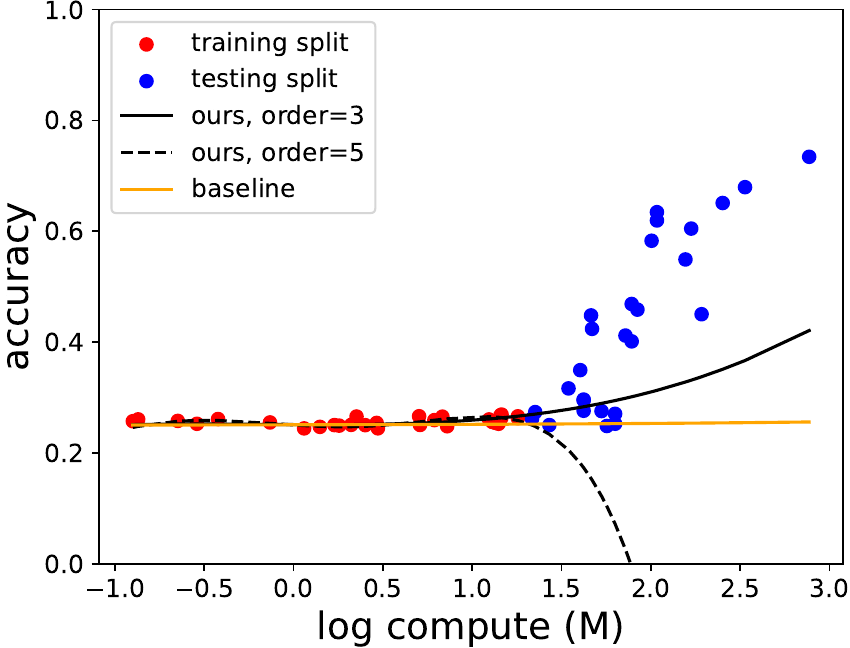}
    \end{subfigure}
    \begin{subfigure}{0.32\textwidth}
      \includegraphics[width=\linewidth]{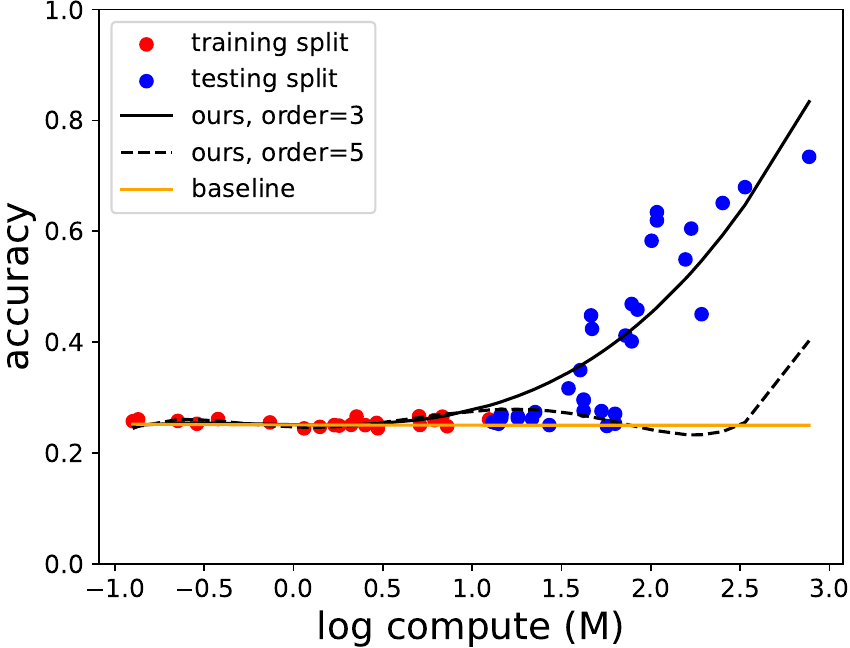}
    \end{subfigure}
    \caption{MMLU. Train-test split threshold being $1.5$, $1.3$, and $1.1$.}
    \label{subfig: mmlu-robustness-threshold}
  \end{subfigure}
  \hfill
  \begin{subfigure}{1\linewidth}
    \centering
    \begin{subfigure}{0.32\textwidth}
      \includegraphics[width=\linewidth]{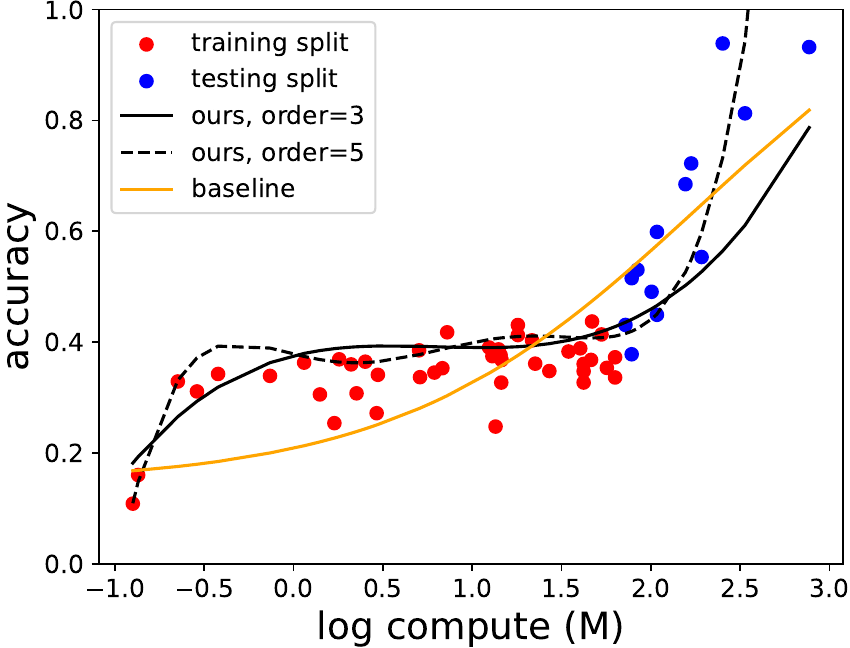}
    \end{subfigure}
    \begin{subfigure}{0.32\textwidth}
      \includegraphics[width=\linewidth]{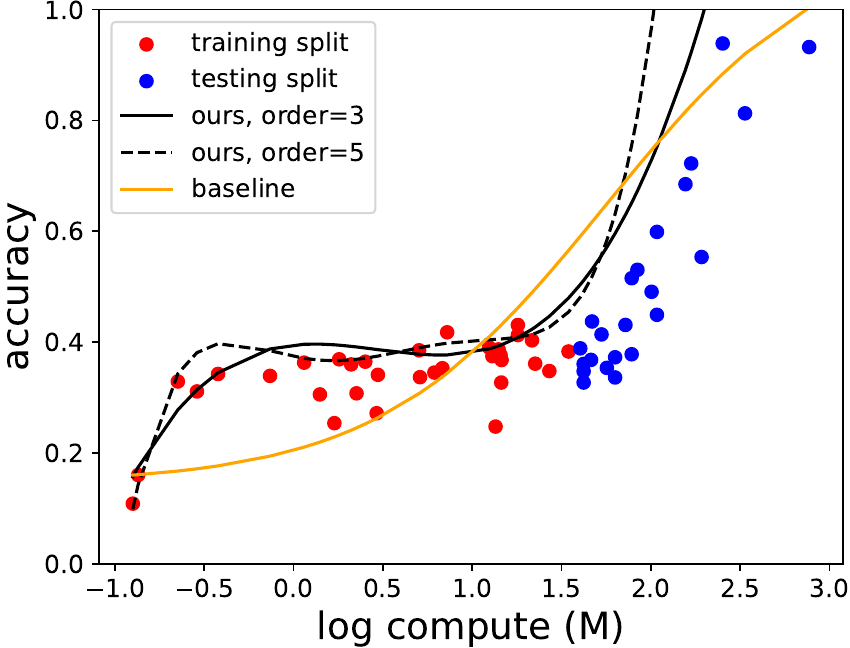}
    \end{subfigure}
    \begin{subfigure}{0.32\textwidth}
      \includegraphics[width=\linewidth]{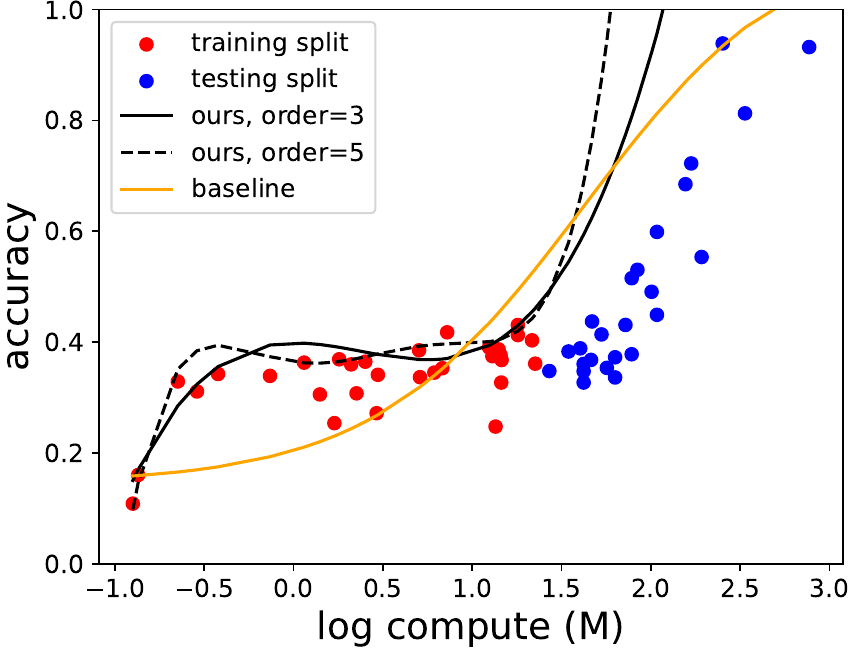}
    \end{subfigure}
    \caption{Arithmetic. Train-test split threshold being $1.8$, $1.6$, and $1.4$.}
    \label{subfig: arithmetic-robustness-threshold}
  \end{subfigure}
  \hfill
  \begin{subfigure}{1\linewidth}
    \centering
    \begin{subfigure}{0.32\textwidth}
      \includegraphics[width=\linewidth]{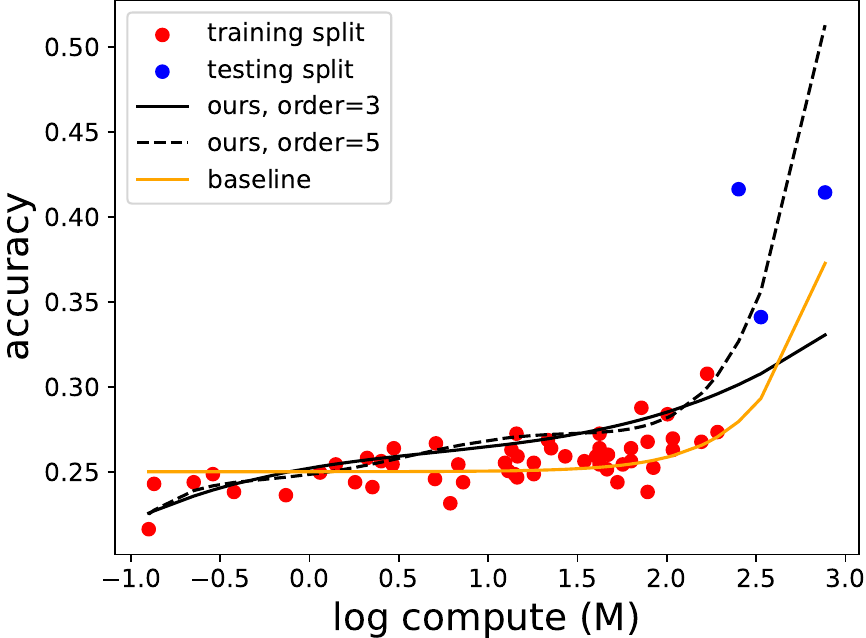}
    \end{subfigure}
    \begin{subfigure}{0.32\textwidth}
      \includegraphics[width=\linewidth]{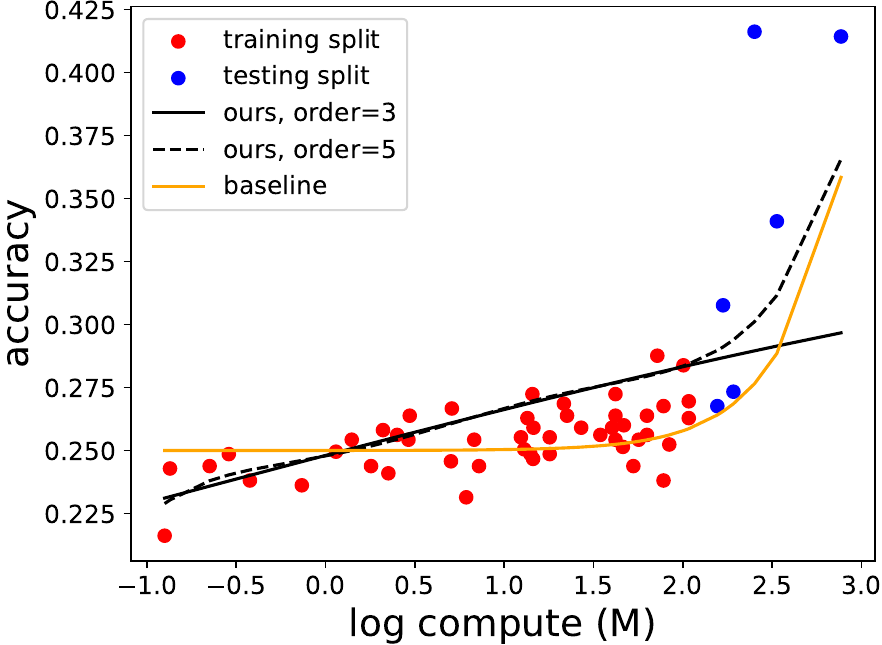}
    \end{subfigure}
    \begin{subfigure}{0.32\textwidth}
      \includegraphics[width=\linewidth]{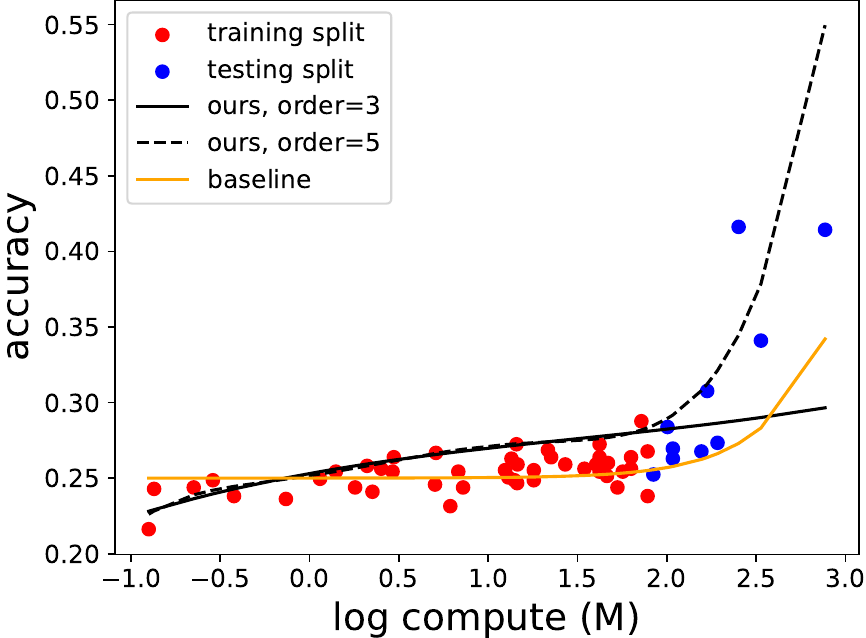}
    \end{subfigure}
    \caption{Persian-QA. Train-test split threshold being $2.3$, $2.1$, and $1.9$.}
    \label{subfig: persian-qa-robustness-threshold}
  \end{subfigure}
  \caption{\textit{Slice-and-Sandwich}'s results of accuracy-based scaling law under different train-test split thresholds. Solid lines are when order=3 is used for easy question fitting, and dashed lines are when order=5 is used.}
  \label{fig: sas-robustness-threshold}
\end{figure*}
Fig.~\ref{fig: sas-robustness-threshold} shows the fitted scaling trend using different train-test splitting thresholds. For the hard question group, we use a polynomial fit of degree 2. For the easy question group, the polynomial fit of degree 3 is represented by a black solid line, and the polynomial fit of degree 5 is represented by a black dashed line. 

The forecast is reasonably robust to the train-test split. All capture the trend and display a similar shape to our original choice of train-test split threshold except for the case where threshold$=1.3$ and degree$=5$ for the MMLU dataset. The polynomial fit of degree 3 for Persian QA is too flat compared to data for all three thresholds and gets flatter as the threshold goes down. We leave it to future work to provide better guidelines as to the least upper bound of training data model log compute that still allows us to confidently predict the onset of emergent abilities.

\subsubsection{Effect of Group Number}
\begin{figure*}[tb]
  \centering
  \begin{subfigure}{1\linewidth}
    \centering
    \begin{subfigure}{0.26\textwidth}
      \includegraphics[width=\linewidth]{figures/images/mmlu_redist_acc_scaling_law.pdf}
    \end{subfigure}
    \begin{subfigure}{0.26\textwidth}
      \includegraphics[width=\linewidth]{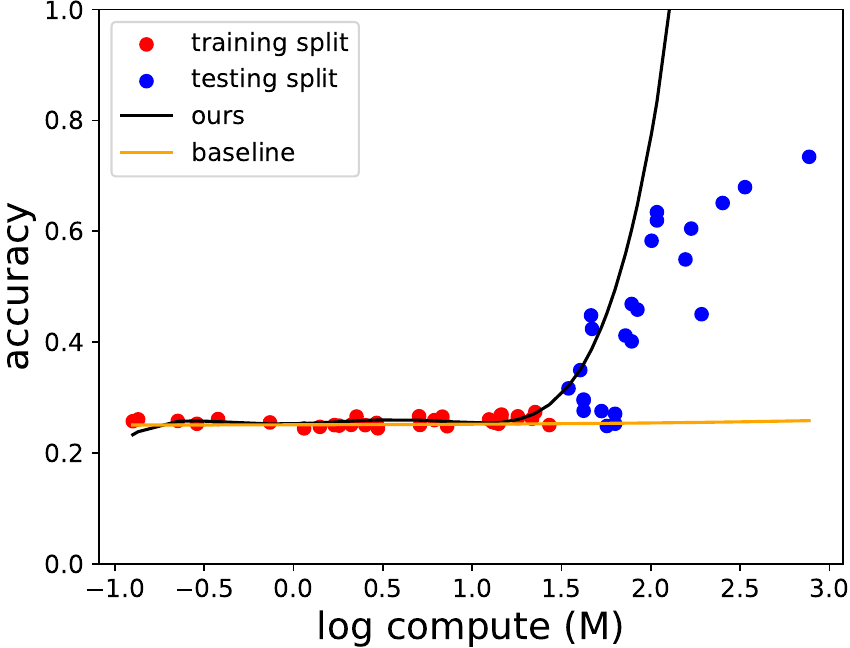}
    \end{subfigure}
    \begin{subfigure}{0.26\textwidth}
      \includegraphics[width=\linewidth]{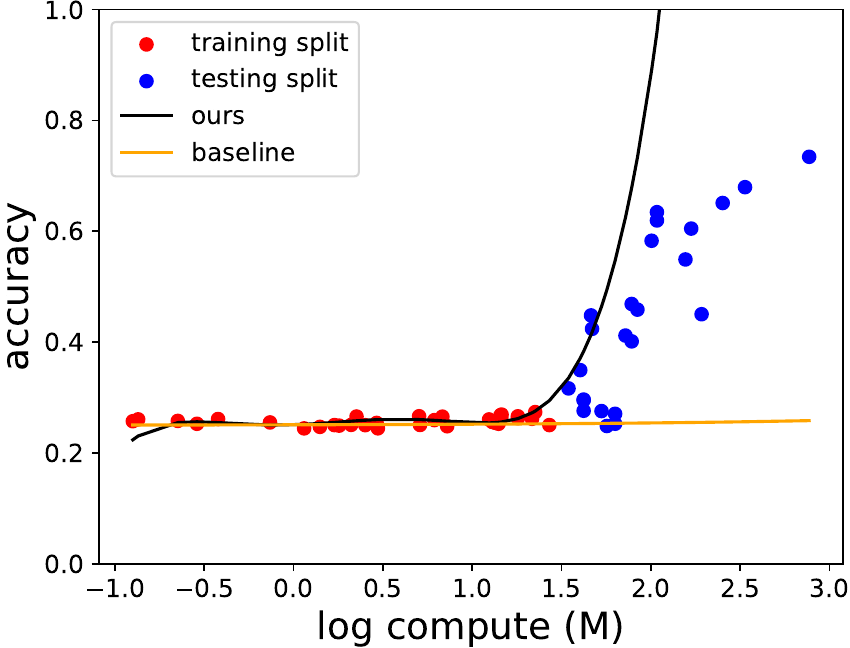}
    \end{subfigure}
    \caption{MMLU with group number $G$ being $3$, $5$, and $7$.}
    \label{subfig: mmlu-robustness-group}
  \end{subfigure}
  \hfill
  \begin{subfigure}{1\linewidth}
    \centering
    \begin{subfigure}{0.26\textwidth}
      \includegraphics[width=\linewidth]{figures/images/arithmetic_acc_redist_acc_scaling_law.pdf}
    \end{subfigure}
    \begin{subfigure}{0.26\textwidth}
      \includegraphics[width=\linewidth]{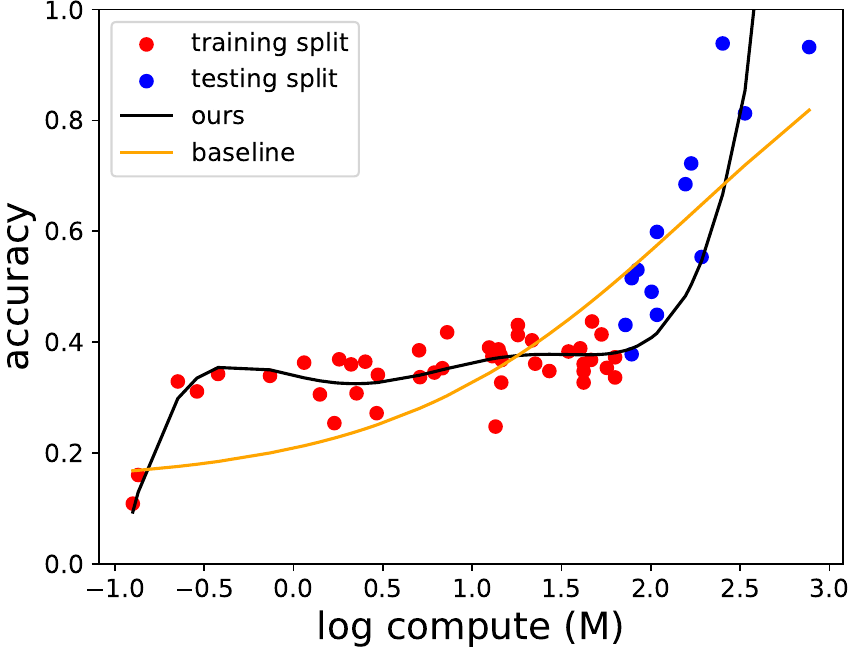}
    \end{subfigure}
    \begin{subfigure}{0.26\textwidth}
      \includegraphics[width=\linewidth]{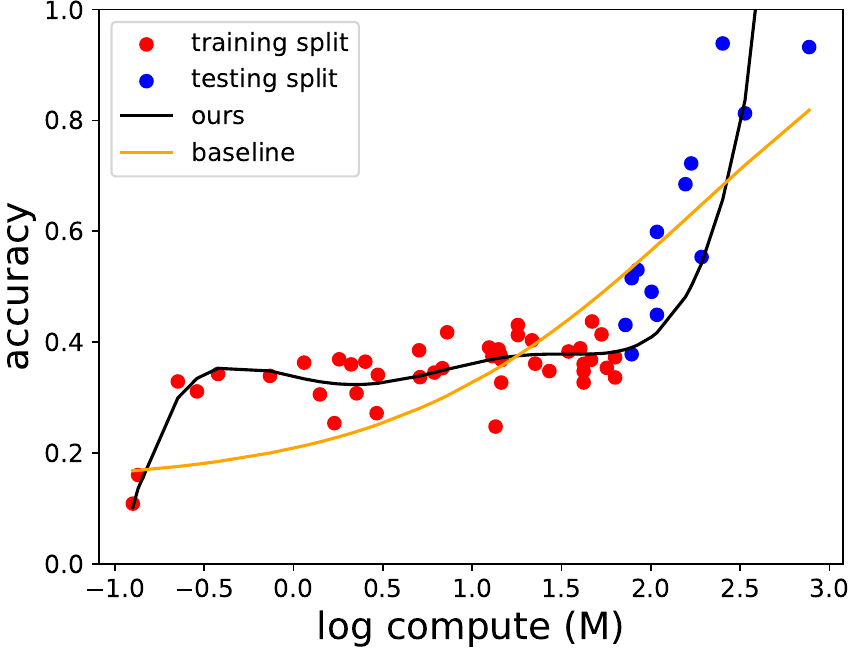}
    \end{subfigure}
    \caption{Arithmetic with group number $G$ being $3$, $5$, and $7$.}
    \label{subfig: arithmetic-robustness-group}
  \end{subfigure}
  \hfill
  \begin{subfigure}{1\linewidth}
    \centering
    \begin{subfigure}{0.26\textwidth}
      \includegraphics[width=\linewidth]{figures/images/persian_qa_acc_redist_acc_scaling_law.pdf}
    \end{subfigure}
    \begin{subfigure}{0.26\textwidth}
      \includegraphics[width=\linewidth]{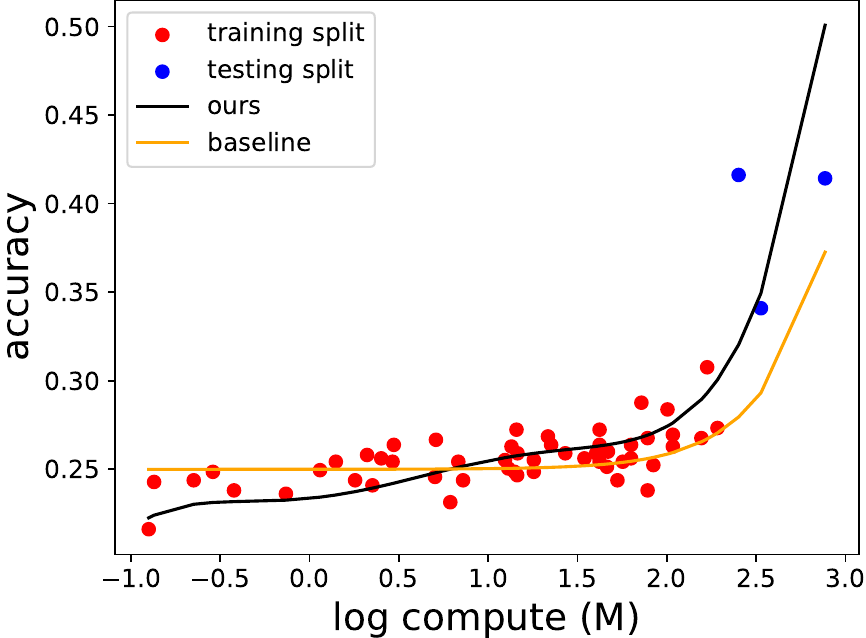}
    \end{subfigure}
    \begin{subfigure}{0.26\textwidth}
      \includegraphics[width=\linewidth]{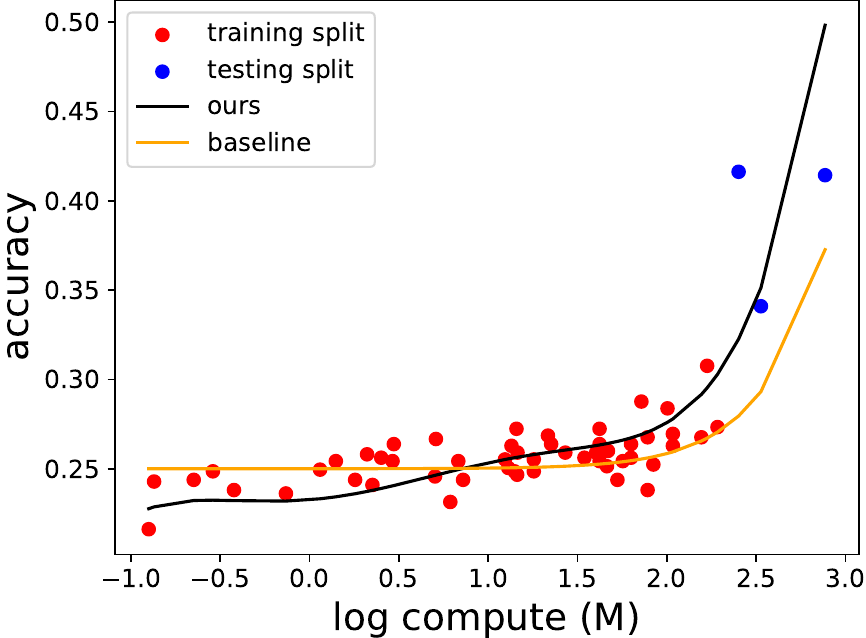}
    \end{subfigure}
    \caption{Persian-QA with group number $G$ being $3$, $5$, and $7$.}
    \label{subfig: persian-qa-robustness-group}
  \end{subfigure}
  \caption{\textit{Slice-and-Sandwich}'s results of accuracy-based scaling law under different group numbers $G$.}
  \label{fig: sas-robustness-group}
\end{figure*}
Fig.~\ref{fig: sas-robustness-group} shows the fitted scaling trends when splitting questions into different numbers of groups. We show group number $G=3, 5,$ and $7$. Following the same procedure and degree parameter in the main paper, easier question groups, such as the groups of difficulty level 1 to 3 for $G=7$, are fitted by polynomial regression of degree=5 due to observed inverted-U scaling; harder question groups are fitted by degree=2 due to U-shaped scaling. Then Eq.~\ref{eq: squeeze} is modified to take the average of Brier-based fitting trends of all but the medium group and project the acquired Brier-based scaling law to the accuracy-based one. \textit{Slice-and-Sandwich} shows its robustness under $G$. The robustness comes from similar fitting results among the same scaling types, such as inverted-U scaling, resulting in a similar final scaling law after taking their averages.

\subsection{Hard Lift - A Simple Alternative Pipeline}
\begin{figure*}[tb]
  \centering
  \begin{subfigure}{1\linewidth}
    \centering
    \begin{subfigure}{0.26\textwidth}
      \includegraphics[width=\linewidth]{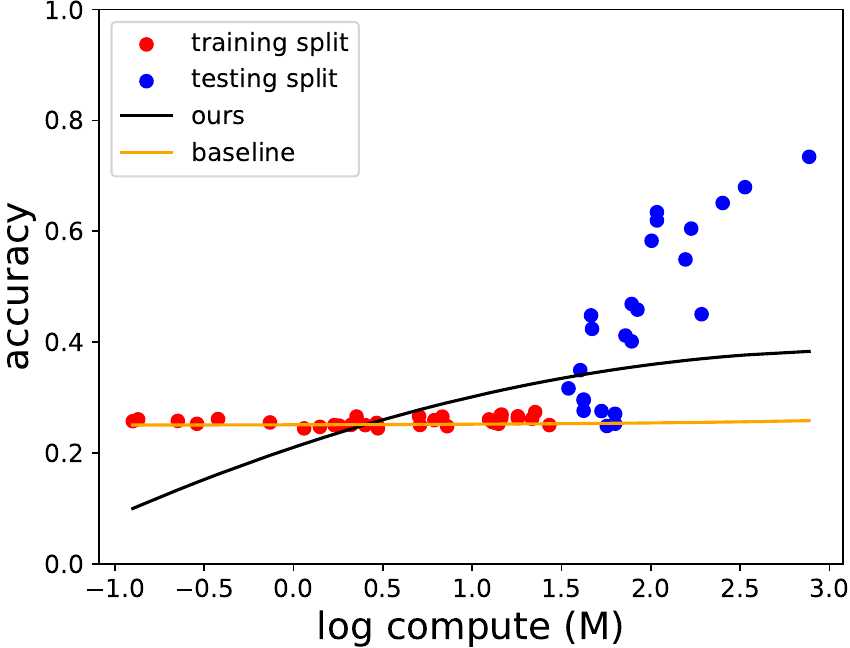}
    \end{subfigure}
    \begin{subfigure}{0.26\textwidth}
      \includegraphics[width=\linewidth]{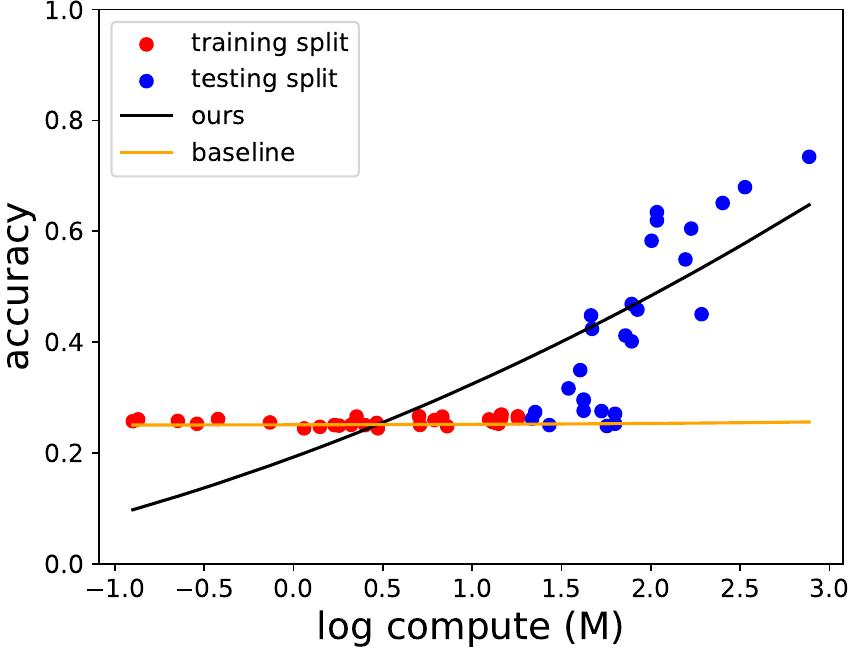}
    \end{subfigure}
    \begin{subfigure}{0.26\textwidth}
      \includegraphics[width=\linewidth]{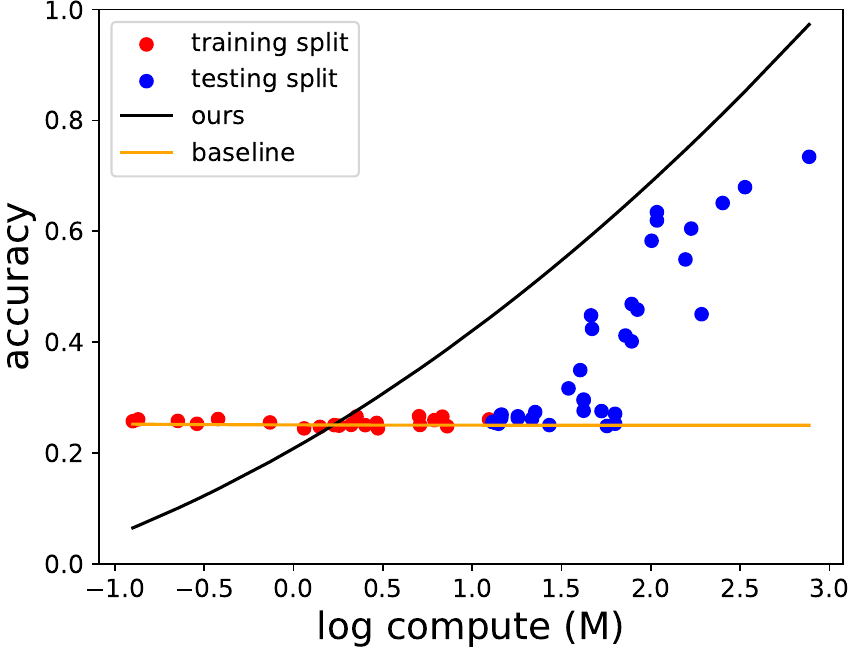}
    \end{subfigure}
    \caption{MMLU. Train-test split threshold being $1.5$, $1.3$, and $1.1$.}
    \label{subfig: mmlu-hard-lift}
  \end{subfigure}
  \hfill
  \begin{subfigure}{1\linewidth}
    \centering
    \begin{subfigure}{0.26\textwidth}
      \includegraphics[width=\linewidth]{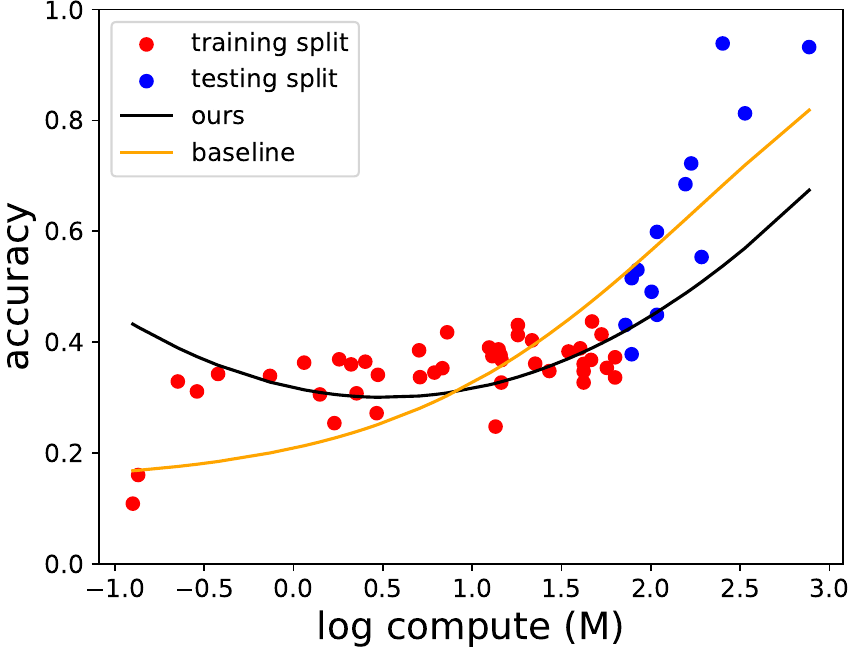}
    \end{subfigure}
    \begin{subfigure}{0.26\textwidth}
      \includegraphics[width=\linewidth]{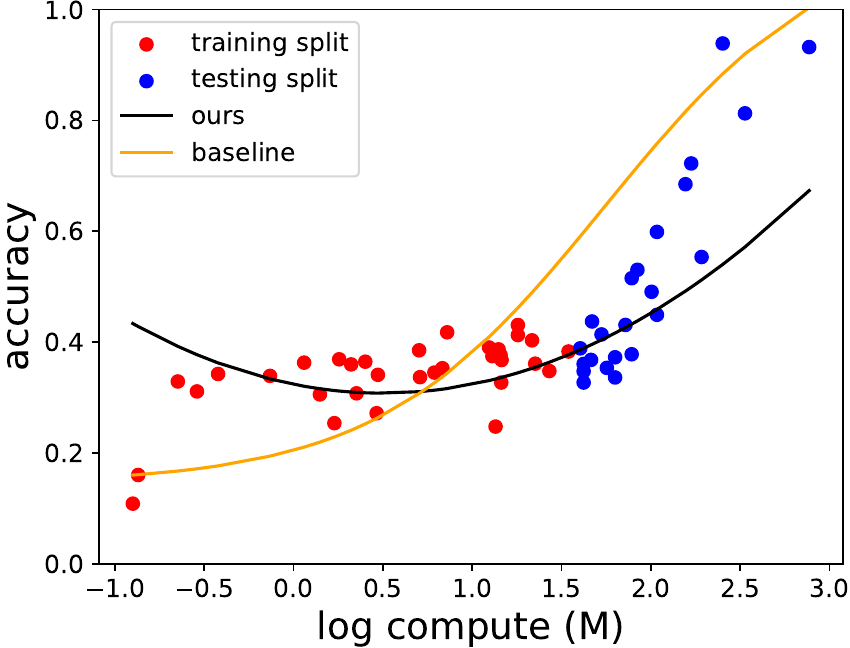}
    \end{subfigure}
    \begin{subfigure}{0.26\textwidth}
      \includegraphics[width=\linewidth]{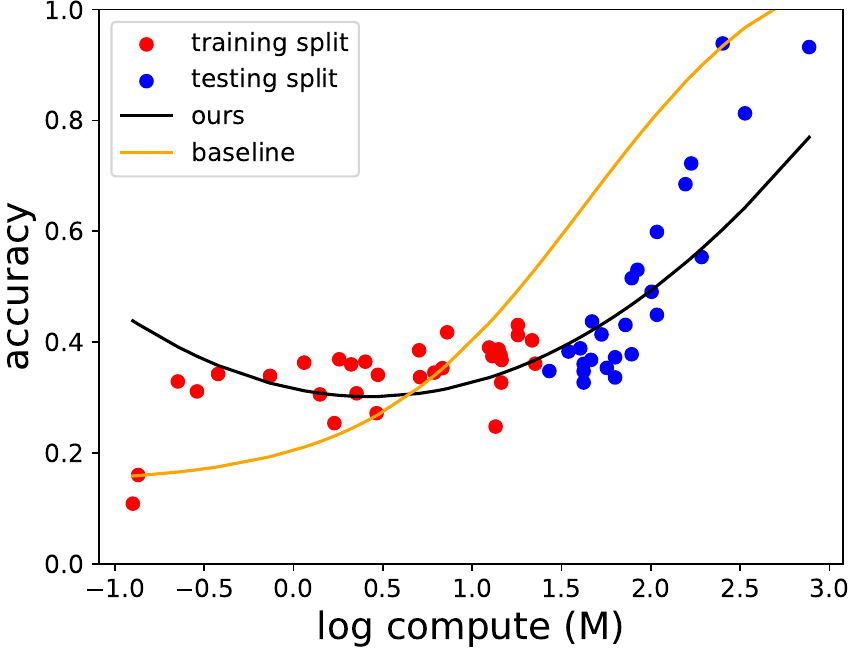}
    \end{subfigure}
    \caption{Arithmetic. Train-test split threshold being $1.8$, $1.6$, and $1.4$.}
    \label{subfig: arithmetic-hard-lift}
  \end{subfigure}
  \hfill
  \begin{subfigure}{1\linewidth}
    \centering
    \begin{subfigure}{0.26\textwidth}
      \includegraphics[width=\linewidth]{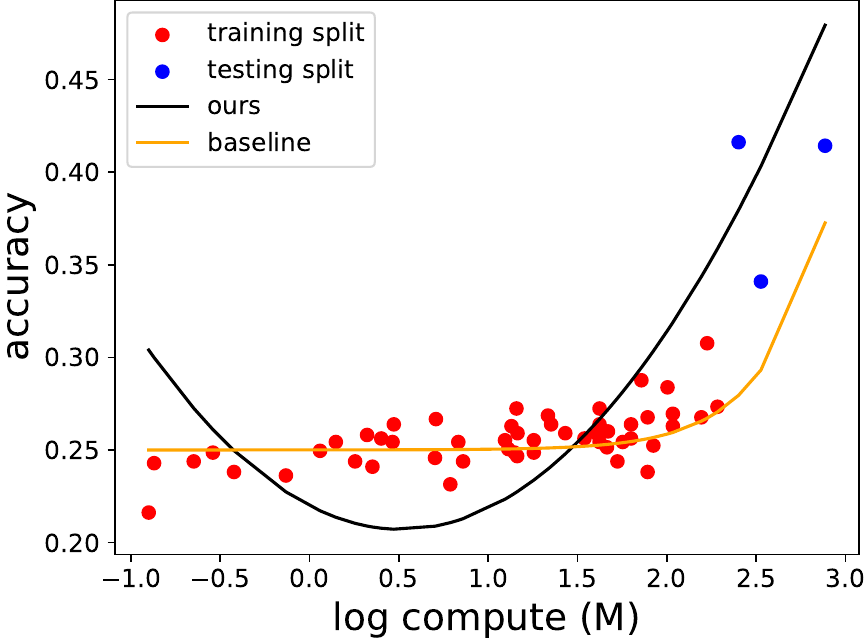}
    \end{subfigure}
    \begin{subfigure}{0.26\textwidth}
      \includegraphics[width=\linewidth]{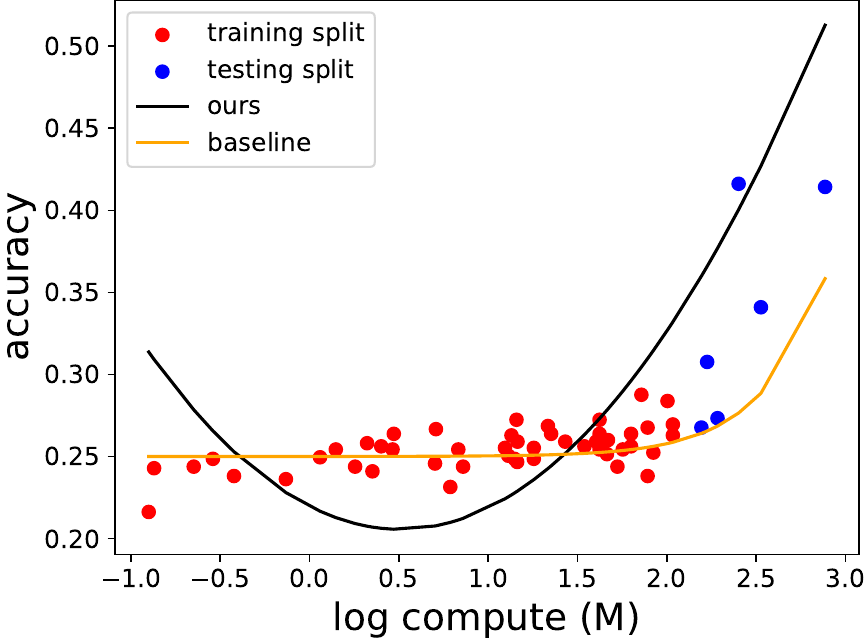}
    \end{subfigure}
    \begin{subfigure}{0.26\textwidth}
      \includegraphics[width=\linewidth]{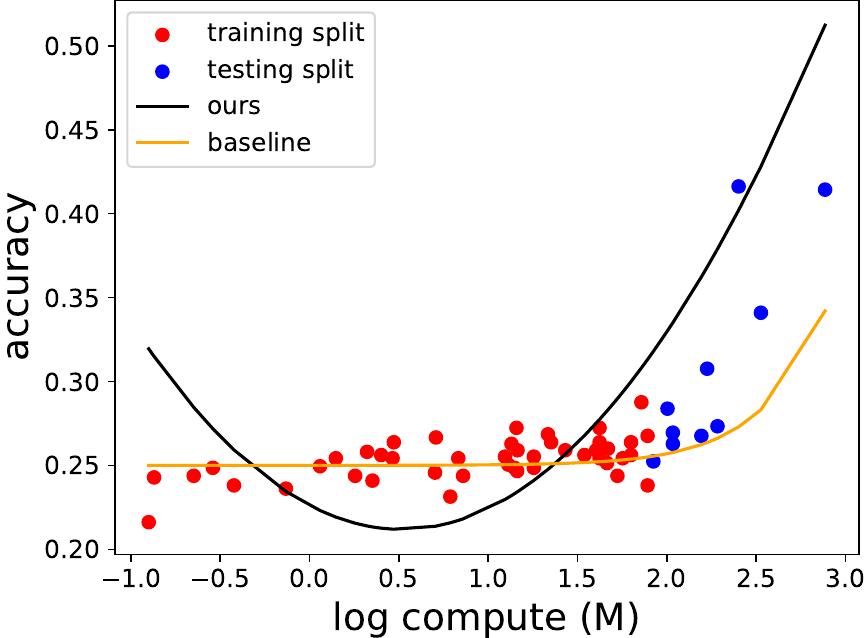}
    \end{subfigure}
    \caption{Persian-QA. Train-test split threshold being $2.3$, $2.1$, and $1.9$.}
    \label{subfig: persian-qa-hard-lift}
  \end{subfigure}
  \caption{\textit{Hard-Lift}'s results of accuracy-based scaling law under different train-test split thresholds.\textit{Hard-Lift} uses order=2 for fitting hard question groups.}
  \label{fig: hard-lift}
\end{figure*}
As an alternative to \textit{Slice-and-Sandwich}, we provide an even simpler pipeline called \textit{Hard-Lift}. Specifically, we take the polynomial fit of degree 2 on TC Brier Score for the hard question group from \textit{Slice-and-Sandwich} and lift it by a constant so the fitted TC Brier Score at the training set model log compute upper bound is equal to the true average. We use this to forecast the TC Brier Score of models past the emergence threshold. We then transform this predicted TC Brier Score back to predicted accuracy via the $G(\cdot)$ function as in \textit{Slice-and-Sandwich}.

Fig.~\ref{fig: hard-lift} shows the results of \textit{Hard-Lift} under different log compute thresholds for train-test split as in Sec. \ref{sec: split}. \textit{Hard-Lift} performs better than the baseline for MMLU (Fig.~\ref{subfig: mmlu-hard-lift}) and Persian-QA (Fig.~\ref{subfig: persian-qa-hard-lift}) datasets, but worse than baseline for the arithmetic dataset. We believe this result reinforces our claim that analyzing difficulty-stratified scaling trends enables more explainable prediction of emergent abilities.
\clearpage
\section{Preliminary Analysis for String-Match Tasks}
\label{sup: ppp on string match}
\begin{figure*}[t]
  \centering
  \begin{subfigure}{0.48\linewidth}
    \includegraphics[width=\textwidth]{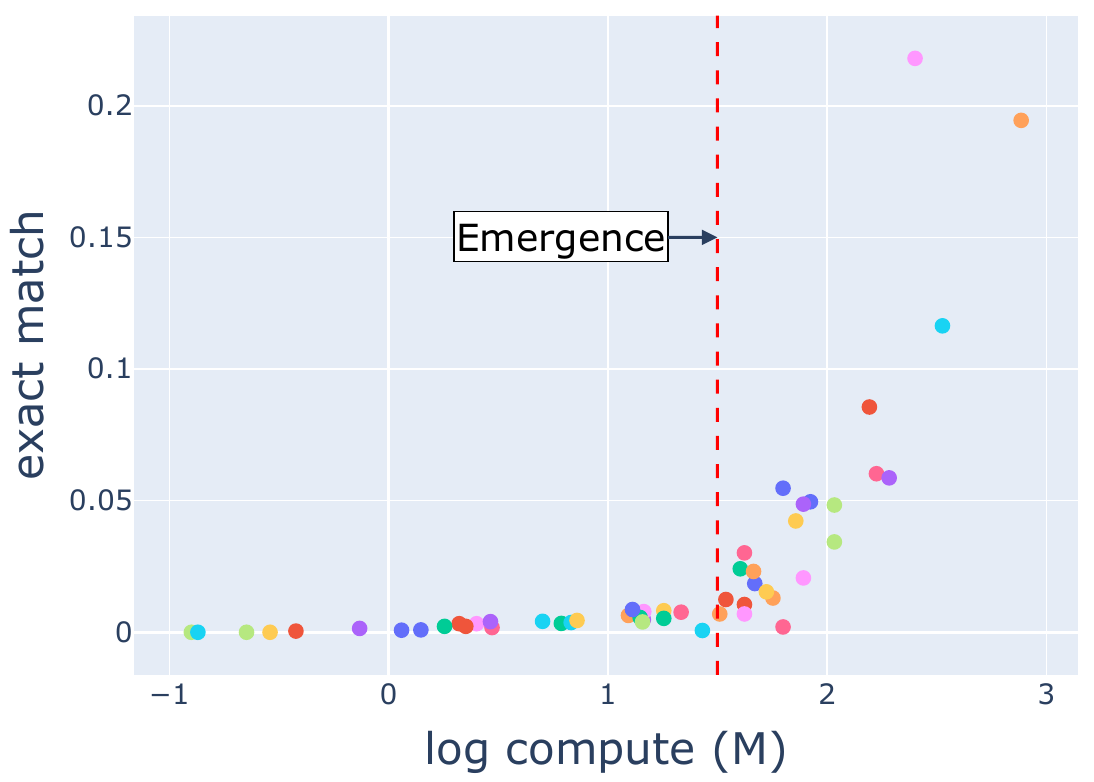}
    \caption{Exact match accuracy vs. log compute (M).}
    \label{subfig: word_unscramble exact match}
  \end{subfigure}
  \hfill
  \begin{subfigure}{0.48\linewidth}
    \includegraphics[width=\textwidth]{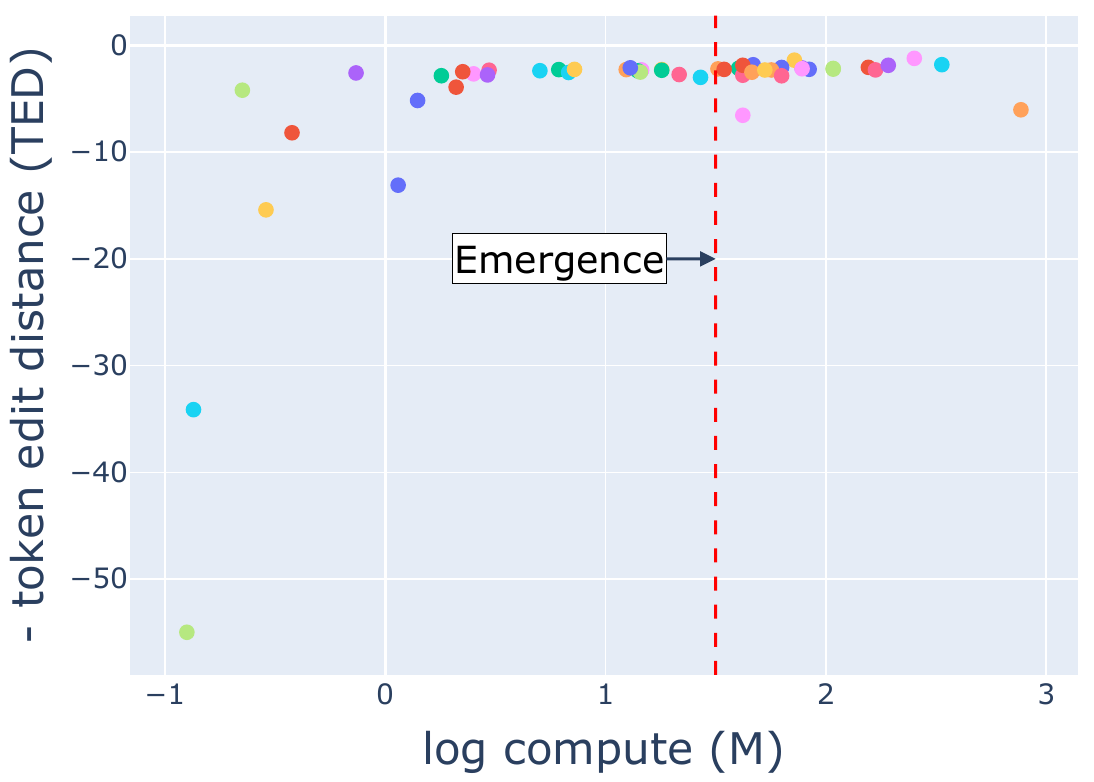}
    \caption{Token edit distance vs. log compute (M).}
    \label{subfig: word_unscramble ted}
  \end{subfigure}
  \hfill
  \begin{subfigure}{0.48\linewidth}
    \includegraphics[width=\textwidth]{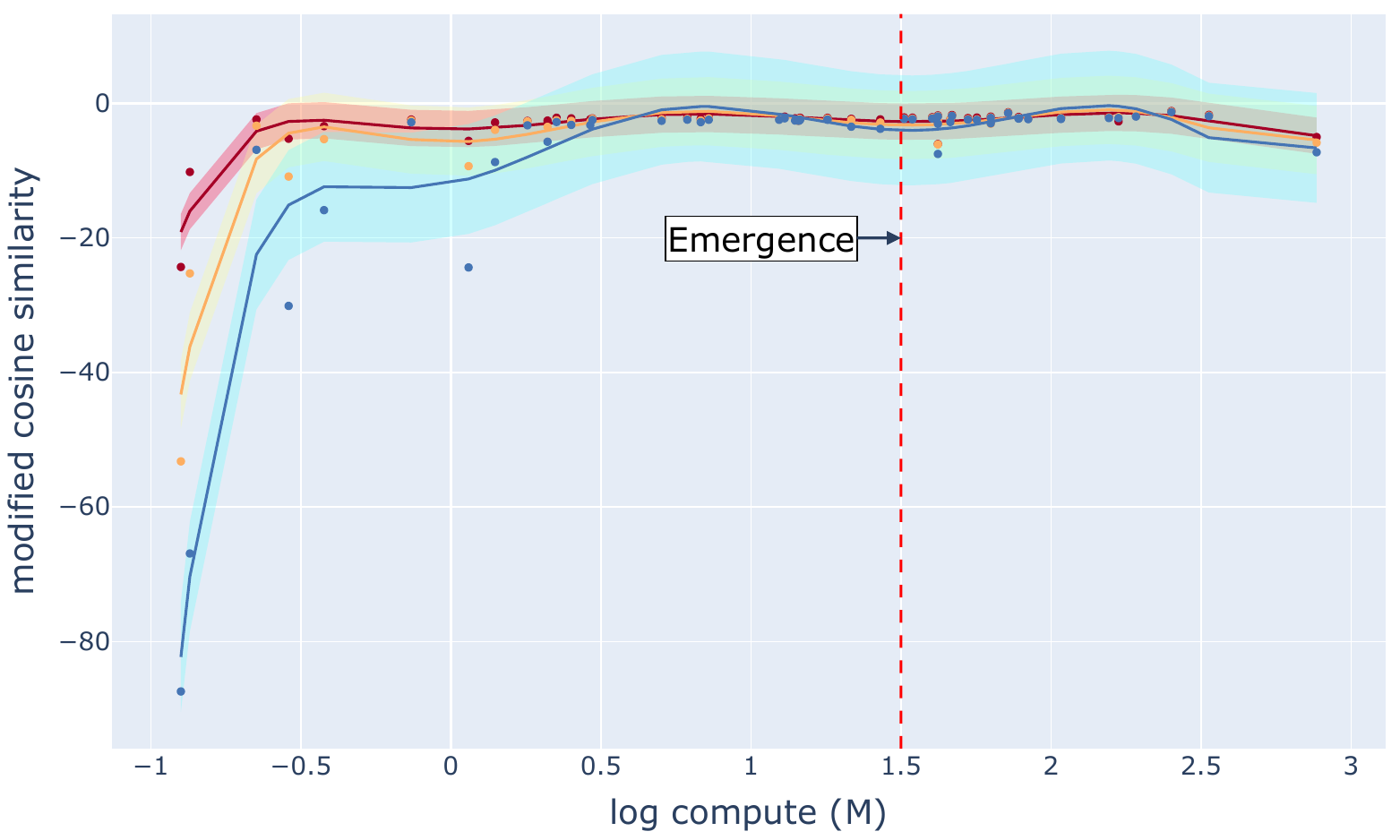}
    \caption{token edit distance vs. log compute (M) with group number $G=3$.}
    \label{subfig: word_unscramble ted group}
  \end{subfigure}
  \hfill
  \begin{subfigure}{0.48\linewidth}
    \includegraphics[width=\textwidth]{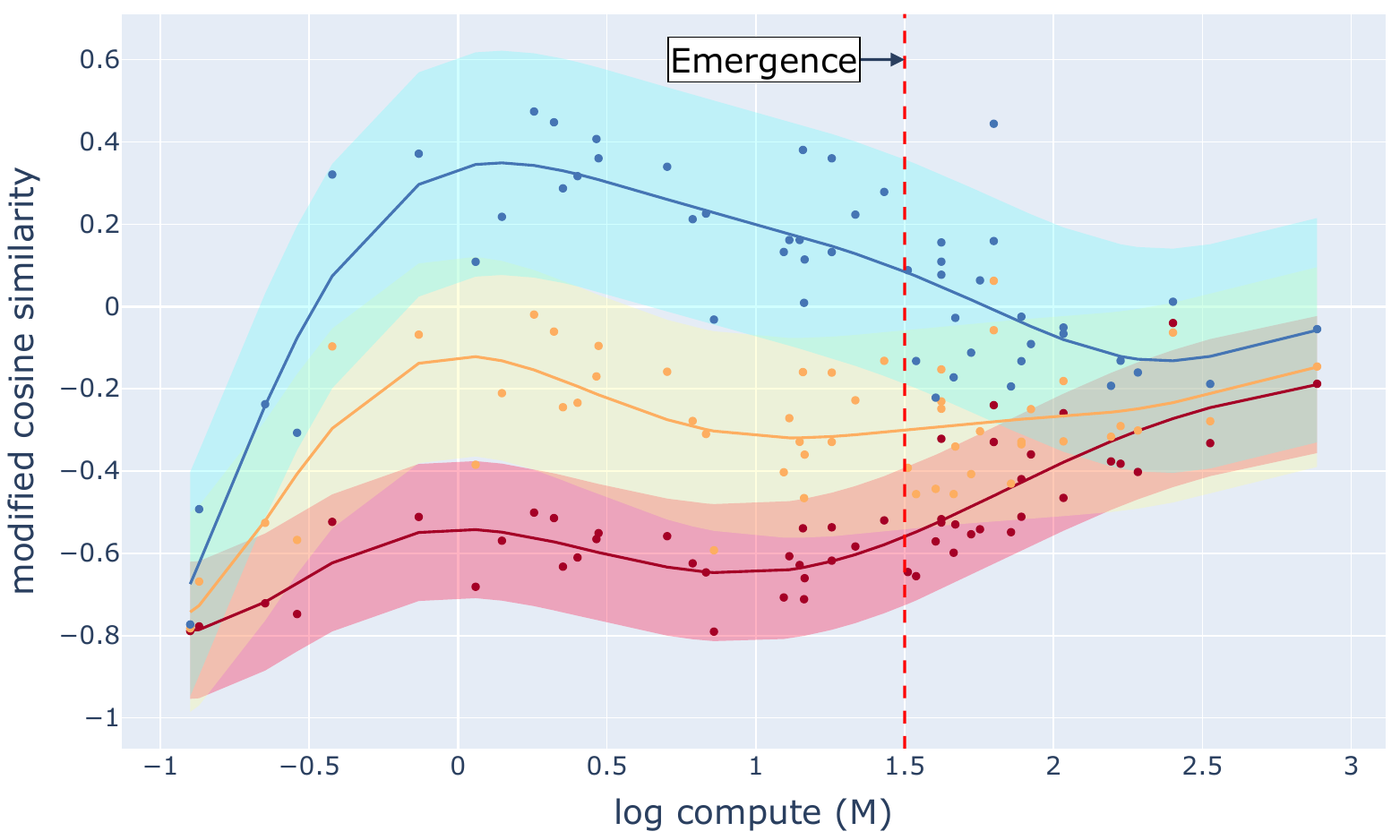}
    \caption{Modified cosine similarity vs. log compute (M) with group number $G=3$.}
    \label{subfig: word_unscramble cos group}
  \end{subfigure}
  \caption{The exact match accuracy, token edit distance (TED), and cosine similarity score vs. log compute (M) on the word unscramble dataset in BIG-bench~\citep{srivastava2023beyond}.}
  \label{fig: word_unscramble phenomenon}
\end{figure*}
\begin{figure*}[t]
  \centering
  \begin{subfigure}{0.48\linewidth}
    \includegraphics[width=\textwidth]{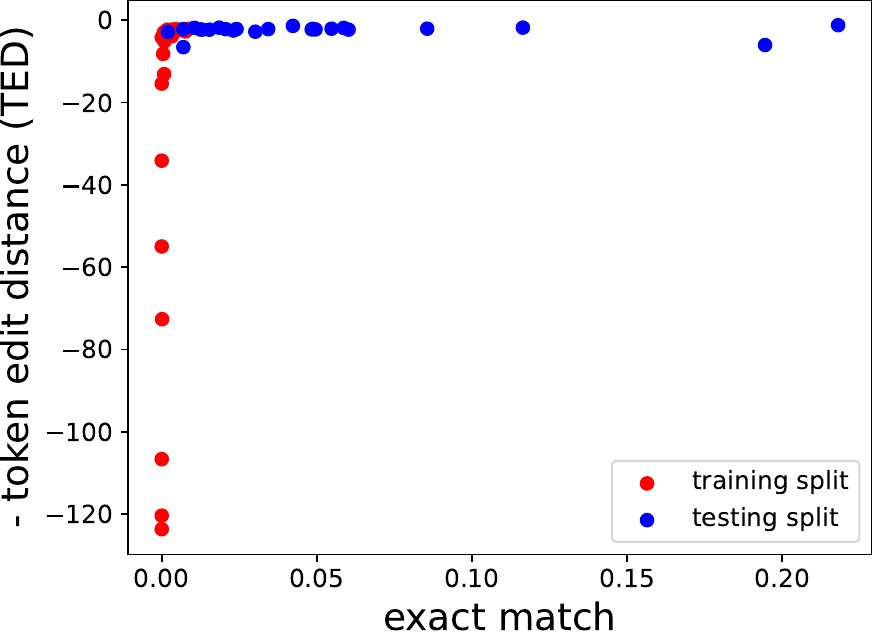}
    \caption{Relation between token edit distance and exact match.}
    \label{subfig: word_unscramble ted vs. exact match}
  \end{subfigure}
  \hfill
  \begin{subfigure}{0.48\linewidth}
    \includegraphics[width=\textwidth]{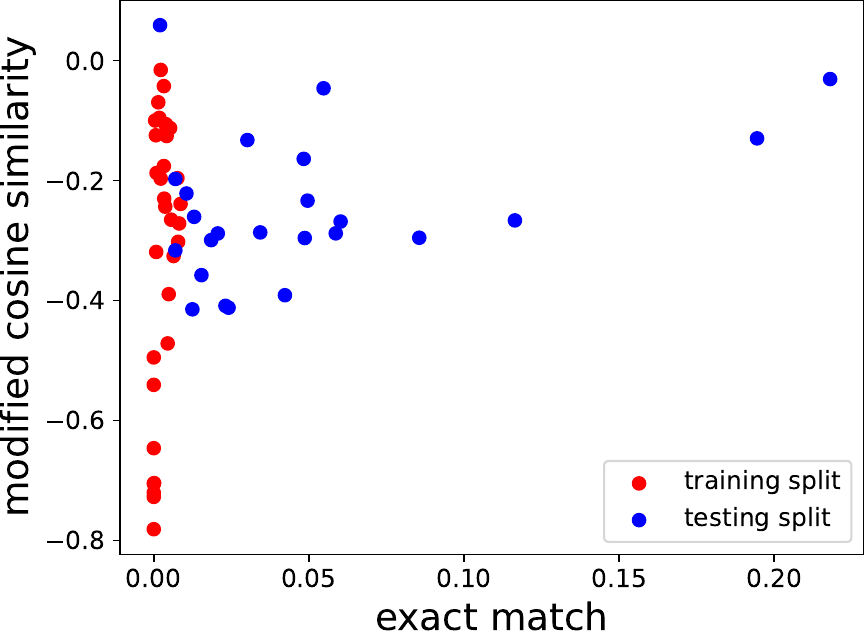}
    \caption{Relation between modified cosine similarity and exact match.}
    \label{subfig: word_unscramble mcs vs. exact match}
  \end{subfigure}
  \caption{Relations between token edit distance/modified cosine similarity and exact match on the word unscramble dataset in BIG-bench~\citep{srivastava2023beyond}.}
  \label{fig: word_unscramble relation}
\end{figure*}
This section provides the preliminary analysis for the exact string match tasks.

Fig.~\ref{subfig: word_unscramble exact match} shows that the word unscramble dataset in BIG-bench~\citep{srivastava2023beyond} exhibits emergent abilities under the traditional metric: exact match accuracy. On the other hand, Fig.~\ref{subfig: word_unscramble ted} shows that model performance measured by token edit distance (TED) as discussed in \citet{schaeffer2024emergent} improves with scale steadily at first and then exhibit flat scaling.

We argue that TED is not a good measure of progress on a string match task. (1) It does not differentiate between easy and hard questions well. Fig. ~\ref{subfig: word_unscramble ted group} shows that performance measured by TED on all three question groups is close for all model log computes above $0.5$. A harder group's TED may be higher or lower than an easier group's. (2) It is not very correlated with exact match, the traditional metric that people are probably ultimately interested in (~\ref{subfig: word_unscramble ted vs. exact match}).

One idea is to measure performance by modified cosine similarity (MCS):
\begin{equation}
  MCS = \frac{F(s1) \cdot F(s2)}{\|F(s1)\| \|F(s2)\|}\cdot\mathbb{I}(s1\subseteq s2),
  \label{eq: modified cosine similarity}
\end{equation}
where $s1$ is the model's output string, $s2$ is the answer string, $F(x)$ is CLIP~\citep{radford2021learning}'s text encoder to project the string to the vector space, and $\mathbb{I}(x)$ is an indicator function having 1 if every single character of $s1$ is contained in $s2$, otherwise 0. MCS takes values in the interval $[-1, 1]$ and is good at differentiating questions by difficulty levels. Fig.~\ref{subfig: word_unscramble cos group} shows that MCS scaling curves of the easy, medium, and hard question groups are clearly ordered.

Interestingly, performance measured by MCS for all question groups exhibits inverted-U scaling followed by steady improvement. The only differences are the model log compute at which scaling reverts from inverse scaling to standard scaling and also how fast performance goes up/down. However, Fig.~\ref{subfig: word_unscramble mcs vs. exact match} shows that MCS is also poorly correlated with the exact match. Even if we can precisely predict the MCS of models above the emergence threshold, conversion back to exact match accuracy will be too noisy to be useful. We hope this section illustrates potential avenues for future work.
\section{Broader Impact}
\label{sup: broader impact}
\subsection{Potential Positive Impacts}
This work identifies U-shaped and inverted-U Scaling of LLM performance once we group questions by difficulty level. We believe this observation can provide the AI community with a deeper understanding of emergent abilities. We also present a forecasting pipeline utilizing the above observation to detect the forthcoming performance soar, the ability of which we believe is crucial in preventing deployment in offensive applications.

\subsection{Potential Negative Impacts}
Given the limitations discussed in Sec. \ref{sec: conclusion},  we do not suggest predicting the forthcoming emergent abilities based on merely one of the methods we discuss. Multiple techniques should be used in parallel to prevent possible false positives or false negatives.

\end{document}